%% file: main.tex
\documentclass[nopreprintline]{elsarticle}

\usepackage{lineno,hyperref}
\usepackage[utf8]{inputenc}
\usepackage{amsmath,amsfonts,bm}
\usepackage{amsthm}

\usepackage[Lenny]{fncychap}

\usepackage{graphicx}
\usepackage{hyperref}       
\usepackage{nicefrac}       
\usepackage{microtype}      
\usepackage{booktabs} 
\usepackage{comment}
\usepackage{mathtools}
\usepackage{fancyhdr,amssymb}
\usepackage[noend]{algorithmic}
\usepackage[linesnumbered]{algorithm2e}
\usepackage{wrapfig}
\usepackage[font=small]{caption}
\usepackage{subcaption}
\usepackage[english]{babel}

\usepackage{bbm}
\usepackage{bm}

\usepackage{minitoc}
\usepackage{url}
\usepackage{empheq}
\usepackage[noend]{algorithmic}
\usepackage{array,booktabs}
\usepackage{nameref}
\usepackage{enumitem}

\usepackage[dvipsnames]{xcolor}
\usepackage{eso-pic} 

\usepackage[nottoc]{tocbibind}
\usepackage{tikz}
\usepackage{totcount}
\newtotcounter{citnum}
\def\oldbibitem{} \let\oldbibitem=\bibitem
\def\bibitem{\stepcounter{citnum}\oldbibitem}

\newtheorem{remark}{Remark}

\usepackage[most]{tcolorbox}

\usepackage[framemethod=tikz,xcolor=true]{mdframed}
\newmdenv[%
    leftmargin=0.2cm,
    backgroundcolor=yellow!10,%
    roundcorner=5pt,%
    tikzsetting={draw=red, line width=2.0pt}%
    ]{SpecialText}%

\def\mut{{\mathbf{M}^{++}}}
\def\Rin{{\mathbb{R}^{d_{\text{in}}}}}
\def\Rout{{\mathbb{R}^{d_{\text{out}}}}}

\def\rhoi{{\rho}}

\def\x{\bold{x}}

\def\nunits{\texttt{n\_units}}
\def\nlayers{\texttt{n\_layers}}

\renewcommand{\eqref}{(~\ref)}
\input{math_commands.tex}

\newcommand{\dif}{\,\mathrm{d}}
\newcommand{\dPx}{\dif \mathcal{P}_X}
\newcommand{\Int}{\displaystyle \int}

\newcommand{\Sum}{\displaystyle \sum}

\modulolinenumbers[5]

\journal{Journal of Computational Physics}

\begin{document}

\begin{frontmatter}

\title{Accelerating hypersonic reentry simulations using deep learning-based hybridization \\
(with guarantees) }

\author[cea,inria,cmap]{Paul Novello\corref{mycorrespondingauthor}}
\cortext[mycorrespondingauthor]{Corresponding author. Now at IRT Saint Exupery and Artificial and Natural Intelligence Toulouse Institute (ANITI), DEEL team}
\ead{paul.novello@outlook.fr}

\author[cea]{Gaël Poëtte}
\ead{gael.poette@cea.fr}

\author[cea]{David Lugato}
\ead{david.lugato@cea.fr}

\author[cea]{Simon Peluchon}
\ead{simon.peluchon@cea.fr}

\author[inria,cmap]{Pietro Marco Congedo}
\ead{pietro.congedo@inria.fr}

\address[cea]{CESTA, CEA, Le Barp, France}
\address[inria]{Inria Saclay, Palaiseau, France}
\address[cmap]{CMAP, Ecole Polytechnique, Palaiseau, France}

\begin{abstract}
In this paper, we are interested in the acceleration of numerical simulations. We focus on a  hypersonic planetary reentry problem whose simulation involves coupling fluid dynamics and chemical reactions. 
Simulating chemical reactions takes most of the computational time but, on the other hand, cannot be avoided to obtain accurate predictions. We face a trade-off between cost-efficiency and accuracy: the simulation code has to be sufficiently efficient to be used in an operational context but accurate enough to predict the phenomenon faithfully.
To tackle this trade-off, we design a hybrid simulation code coupling a traditional fluid dynamic solver with a neural network approximating the chemical reactions.
We rely on their power in terms of accuracy and dimension reduction when applied in a big data context and on their efficiency stemming from their matrix-vector structure to achieve important acceleration factors ($\times 10$ to $\times 18.6$). 
This paper aims to explain how we design such cost-effective hybrid simulation codes in practice. Above all, we describe methodologies to ensure accuracy guarantees, allowing us to go beyond traditional surrogate modeling and to use these codes as references.
\end{abstract}

\begin{keyword}
Reentry \sep Chemical reactions \sep Machine Learning \sep Deep neural networks 
\end{keyword}

\end{frontmatter}

\section{Introduction}
\label{intro}

\begin{figure}[!h]
  \centering
  \includegraphics[width=0.5\linewidth]{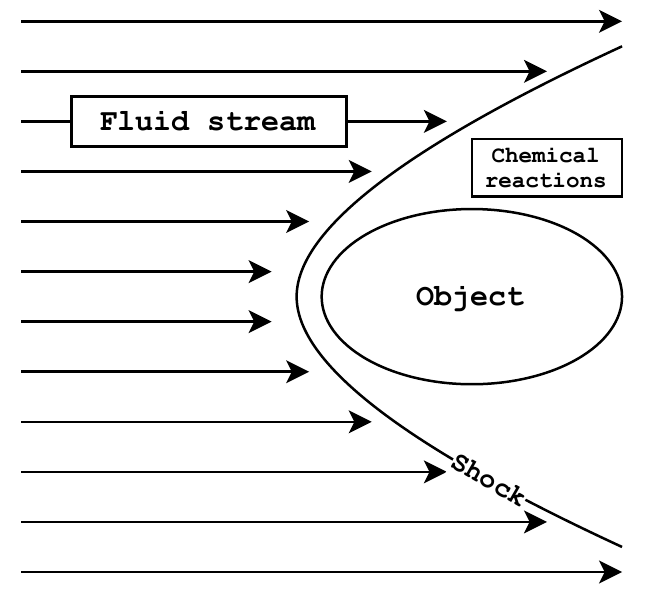}
  \caption{General sketch of a reentry problem: the object entering the atmosphere is subject to a high-speed fluid stream creating a shock ahead of the object. The temperature and
  pressure rise between the shock and the object leading to chemical dissociation reactions. An accurate prediction of the flow field is mandatory in order to design efficient
  protections and ensure the integrity of the object.}
  \label{fig:simulation}
\end{figure}

In this paper, we are interested in the acceleration of numerical simulations. More specifically, we focus on a hypersonic planetary reentry problem \cite{StrongCoupling, Milos, NSHYPABL, peluchon, latige, Yin2009}: during a high-speed planetary atmosphere reentry, a shock wave forms ahead of the entering object leading to 
an increase of temperature and pressure of the fluid across the shock (see figure \ref{fig:simulation} for a general sketch).

\begin{figure}[!h]
  \centering
  \begin{subfigure}{0.45\textwidth}
    \begin{center}
      \includegraphics[trim={0 2cm 0 0},width=1\textwidth]{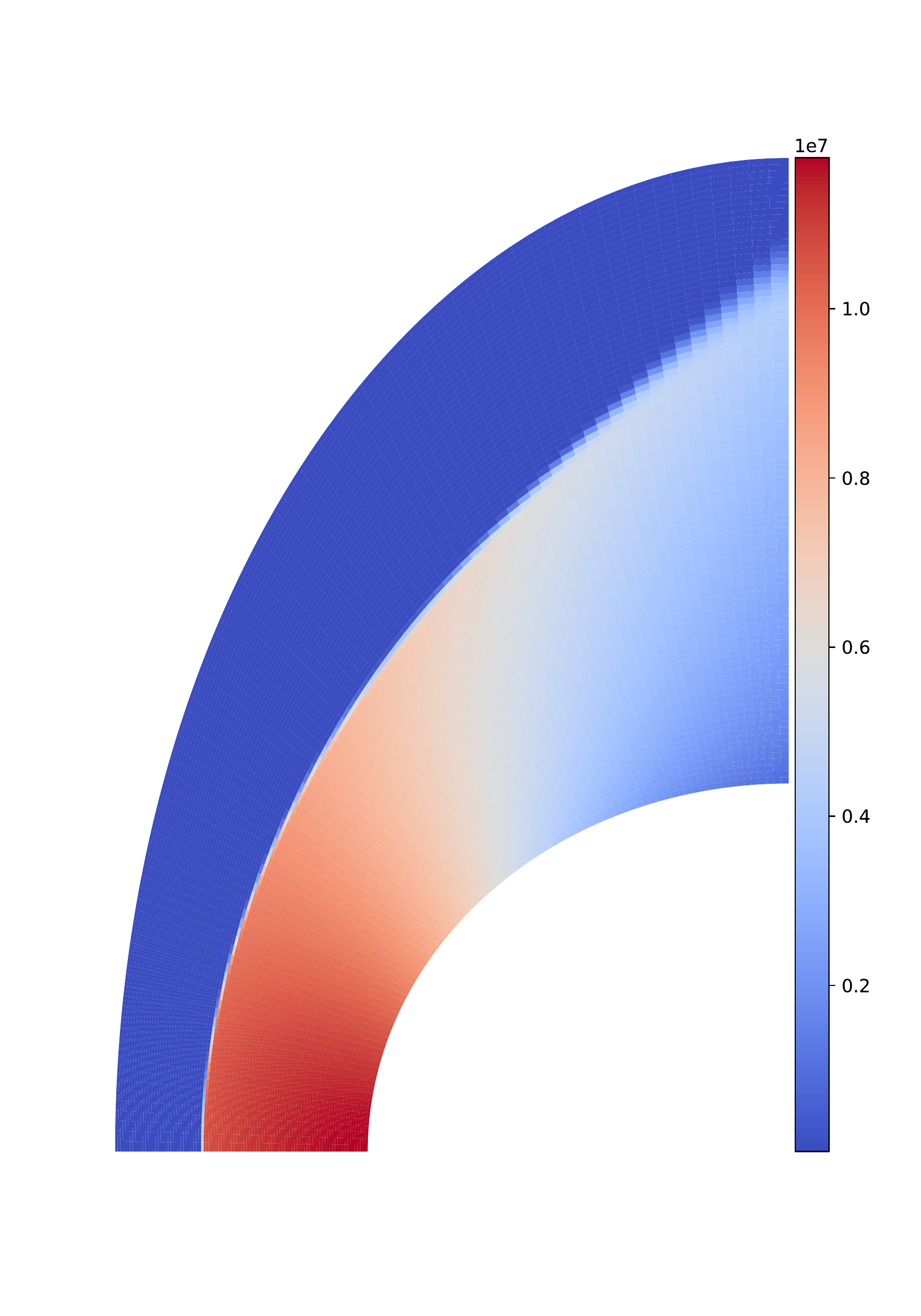}
    \end{center}
    \caption{Without chemical reactions \\ run-time: 81 s.}
  \end{subfigure}
  \begin{subfigure}{0.45\textwidth}
    \begin{center}
      \includegraphics[trim={0 2cm 0 0},width=1\textwidth]{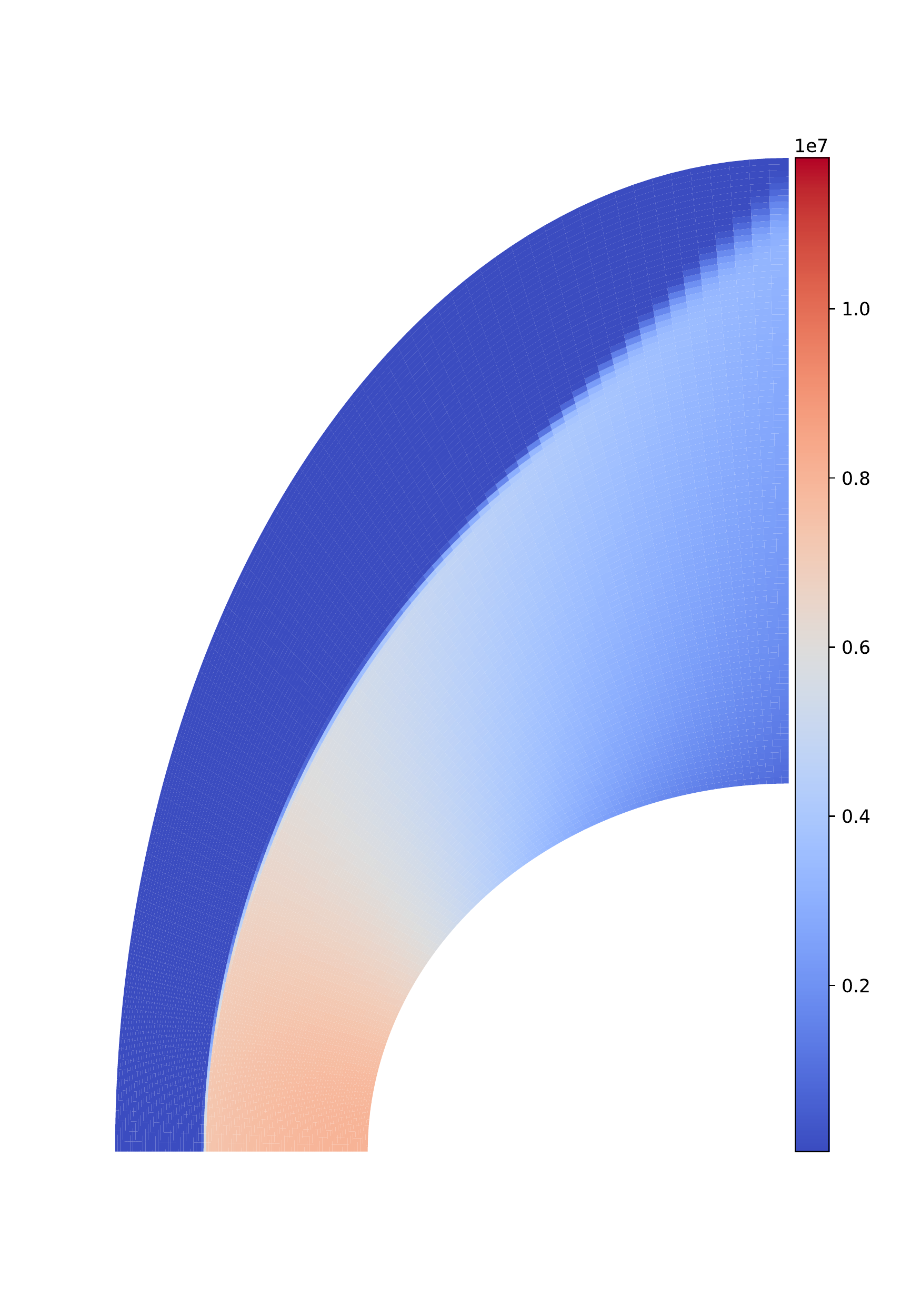}
    \end{center}
  \caption{With chemical reactions \\ run-time: 4090 s.}
  \end{subfigure}
   \caption{Pressure fields (in $Pa$) for a sphere entering earth's atmosphere with and without taking into account chemical reactions (see details in section \ref{numres}).}\label{fig:field1}
\end{figure}

This increase generates chemical dissociation reactions within the shocked fluid and changes its composition. The composition strongly affects the thermodynamic quantities between
the shock and the object \cite{peluchon, latige, mutation}: in other words, we here face a strong coupling problem between gas dynamics and reactions.  
Several other physical phenomenon are certainly also important, such as turbulence \cite{danvin, danvin_ola}, ablation \cite{peluchon, latige}, pyrolysis \cite{peluchon, Yin2009} etc. Although it may be necessary to simulate the phenomenon accurately, the coupling of these different physics can make the computations prohibitively intensive. In the following, we focus on the coupling between compressible gas dynamics and the physics of reacting fluids. Despite being simpler than simulating the full physics, this test case is sufficiently challenging to emphasize the computational problems arising when coupling different physics. 
Let us now give an idea of the difficulty of such simulation by considering a simple motivating example:
figure \ref{fig:field1} presents two pressure fields (the same scale is used on both pictures) of air around a sphere entering (a simplified) earth's atmosphere at a normal velocity of $4930.83$ $m. s.^{-1}$ (Mach $16$). 
The complete details of the simulations are given later on in section \ref{numres}, let us here focus on the results.
The two simulations of figures \ref{fig:field1} (a) and (b) only differ from the fact that on figure \ref{fig:field1} (a), chemical reactions
are neglected whereas on figure \ref{fig:field1} (b), reaction $N+O \leftrightharpoons NO$ is taken into account.  
The pressure on figure \ref{fig:field1} (a) is higher than on figure \ref{fig:field1} (b) and the position of the shock is different. 
By neglecting the chemical reactions, the pressure on the boundary of the object is overestimated in this case, which can lead to sub-optimal object conception. As an example, designing protections for such an object based on the simulations that do not take chemical reactions into account would lead to heavier designs, with propellers consuming more fuel. However, simulating chemical reactions (even in this simplified case where we only consider $N+O\leftrightharpoons NO$) comes with a cost: the simulation of figure \ref{fig:field1} (b) is about $\times 50$ more costly than the one of figure \ref{fig:field1} (a). Simulating chemical equilibrium takes most of the computational time of the code, but on the other hand, Figure \ref{fig:field1} shows that we cannot avoid simulating it to obtain
accurate predictions. There are consequently high stakes in accelerating the computations related to the chemistry of the problem (let us keep in mind that to design a new
reentry object, more than one computation is needed, for example, to propagate uncertainties \cite{PC_LemaitreHaar, Walters,PC_Poette,PC_LucorJFM07} or in order to optimize designs under uncertainties
\cite{PC_congedo_optimization, SA_daveiga_GP}). Some uncertain situations are considered in section \ref{numres}.


The previous example helps to understand our objective:
we want to be able to perform accurate predictions for reentry problems by taking into account chemical reactions (at equilibrium in this paper), together with comprehensive parametric studies. Hence, we need to accelerate our simulations. To achieve this goal, in this paper, we study the design of a neural network-based hybrid code. 

Neural networks have already been intensively used in computational physics for surrogate modeling, such as, for instance, in molecular simulations \cite{al1mol,al2mol,al3mol}, biological simulations \cite{data_assimil,data_assimil1}, fluid dynamic simulations \cite{NNcalib1, CNNCFD, gilles1, danvin, MILAN2021110567}. Even though supervised approaches are often followed, some emerging techniques are becoming increasingly used, such as physics-informed deep learning \cite{pinn,pinngen,pinnreview,deepOnet,}, deep learning aided simulations \cite{LI2022110884,HUANG2020109675,WANG2020108963} or hybridization \cite{hybrid_1,hybrid_2,hybrid_3,hybrid_4,hybrid_5,gilles2}.

In this work, we study the last approach and leverage the coupling structure of the simulation code to replace the chemical reaction solver with a neural network. The advantages of this approach are twofold. First, the definition of neural networks allows us to easily vectorize their calls on array-like data structures such as meshes. It is a solid computational advantage compared to the original simulation code, which has to call the chemical reaction solver in each mesh cell. Second, the chemical reaction solver is not costly when executed as a standalone application. Consequently, it is possible to build a large training database and substantially improve the neural network's accuracy without additional cost during inference. Besides, the neural network only approximates the chemical reactions of the problem, thereby being applicable, once trained, to any simulation which involves the same chemical reactions.

%
Still, these gains are not necessarily easily earned as the neural network built to approximate the solution of the system of chemical reactions must be both accurate and cost-effective in order to accelerate reentry computations. 
The {\em aim of this paper is to explain how we build such cost-effective neural networks in practice and how we progressively test them before embedding them within the simulation code.
Above all, how we use them so that guarantees of accuracy are ensured with the hybrid simulation code}. As an important point, the methodologies introduced in this paper are not specific to the studied reentry test case: they can be applied to any simulation code involving a strong coupling between different physics.

The paper is organized as follows. Section \ref{cfd} describes the physical model we consider in this paper. The model is relatively simple but representative of the difficulties encountered in real-life applications. Furthermore, its simplicity allows for obtaining reference solutions in practical times.
Section \ref{ml} focuses on Machine Learning (ML) and explains why we choose to consider (deep) neural networks in order to accelerate the simulation codes rather than more classical ML models such as polynomial regression or Kriging. We also provide a methodology to anticipate the potential gains in terms of acceleration before plugging any model within the simulation code. Then, we emphasize the trade-off between accuracy and describe how to construct neural networks that are both accurate and cost-effective based on a comprehensive goal-oriented sensitivity analysis of their hyperparameters.
Section \ref{numres} is devoted to numerical results and gives a practical description of the methodology we apply in this paper. We assess the hybrid code and find that the obtained predictions are both quantitatively and qualitatively very promising while being $18.7$ times faster. Above all, we explain how we {\em obtain guarantees with the hybrid code}. 
Finally, section \ref{ccl} is a concluding section: in particular, it summarizes the conditions under which the described methodology can be applied to any other physics/simulation
codes.

\section{Coupling compressible gas dynamics with chemical reactions at equilibrium}
\label{cfd}
%

In this section, we describe the set of partial differential equations (PDEs) solved in order to produce the results of figure \ref{fig:field1} together with its resolution
strategy. 
It corresponds to the coupling of compressible gas dynamics with chemical reactions at equilibrium.
The fluid dynamic is modeled thanks to the Euler equations in 2D spatial dimension. 
It describes the behavior of non-viscous compressible gas. 
In reentry problems, Navier-Stokes' equations are generally considered a finer model, but Euler's system is enough for the purpose of this paper, and the following material can
easily be applied to any other fluid model without more difficulties. 
The Euler system in 2D spatial dimension solved in the spatial domain $\mathcal{D}$ is given by 
\begin{equation}\label{eq:euler}
    \begin{dcases}
  \partial_t U(x,y,t) + \nabla \cdot F(U(x,y,t)) = 0, & \forall (x,y) \in\mathcal{D},\\
      U(x,y,t)=b(x,y,t), & \forall (x,y) \in\partial \mathcal{D}.
    \end{dcases}
\end{equation}
In the above equation, $b$ corresponds to the boundary conditions: typically, it corresponds to an incoming flux boundary condition everywhere on $\partial \mathcal{D}$ except on
the boundary of the object
where it corresponds to a no-slip one (this is what has been used in the results of figure \ref{fig:field1}). 
In \eqref{eq:euler}, the different quantities are defined by (we drop the spatial and time dependences for the sake of conciseness)
\begin{equation*}
    U = 
    \begin{pmatrix}
        \rhoi_1\\
        \dots \\
        \rhoi_{n_e}\\
        \rho v\\
        \rho w\\
        \rho E\\
    \end{pmatrix},
    F_x(U) = 
    \begin{pmatrix}
      \rhoi_1 v\\
      \dots \\
      \rhoi_{n_e}v\\
        \rho v^2 + p\\
        \rho vw\\
        \rho u(E + \frac{p}{\rho})\\
    \end{pmatrix},
    F_y(U) = 
    \begin{pmatrix}
      \rhoi_1 w\\
      \dots \\
      \rhoi_{n_e}w\\
        \rho vw\\
        \rho w^2 + p\\
        \rho w(E + \frac{p}{\rho})\\
    \end{pmatrix}.
\end{equation*}
The first $n_e$ equations stand for the conservation of the mass of the different elements
of the fluid ($N$ and $O$ typically for the two elements of reaction $N+O\leftrightharpoons NO$). 
The partial density of element $k$ can be expressed as
$\rho_k = \sum_{i=1}^{n_s} a_i^k \frac{m^e_k}{m_i} \rho_i$, where $m^e_k$ and $m_i$ are molar masses of element $k$ and species $i$ while $a_i^k$ is the number of the~$k^{th}$ element in species~$i$. Partial mass of the $n_s$ species are denoted by $(\rho_i)_{i\in \{1,\ldots,n_s\}}$.
The density of the fluid can be deduced from the partial densities of the elements as $\rho = \sum_{k=1}^{{n_e}} \rho_k$. 
Besides, $v$ and $w$ are respectively the horizontal and vertical velocities of the fluid so that the equations on $\rho v$ and $\rho w$ in \eqref{eq:euler} ensure the conservation of
momentum. 
Finally, $E = \epsilon + \frac{v^2 + w^2}{2}$ is the total energy of the fluid and $\epsilon$ its internal energy. The last equation of (\ref{eq:euler}), on $\rho E$, ensures the conservation of
the total energy of the fluid. 
The system remains to be closed: we need one last equation to relate the pressure $p$ to the other quantities. 

The first solution for the latter purpose is to make the hypothesis that chemical reactions have a negligible effect. It is the case, for example, when the fluid is considered a perfect gas, in which case $p$ is related to $\epsilon$ and directly to $\rho$ with
\begin{equation*}
    p = (\gamma - 1)\rho \epsilon,
\end{equation*}
where $\gamma$ is the Laplace constant. 
In the case of a diatomic gas, for example, $\gamma=1.4$.
When chemical reactions are neglected, the closure is simple and computationally fast, and the system can be closed without updating the partial densities of the elements
$(\rho_i)_{i\in\{1,...,n_s\}}$. 
However, this hypothesis is coarse (see figure \ref{fig:field1} and the related comments). In a more general case, the pressure $p$ can be accurately computed by simulating the chemical equilibrium between the species produced by the $N_r$ chemical reactions that occur during the dynamic. 
Every elementary chemical reaction $r\in\{1,...,N_r\}$ can be described through the general formula,  
$$
\Sum_{i=1}^{n_s} \nu_{ir}A_i \leftrightharpoons  0, 
$$
where $n_s$ is the number of species (i.e. $n_s=3$ in the case of reaction $N+O\leftrightharpoons NO$ for $N,O$ and $NO$) and 
where $(\nu_{ir})_{i\in\{1,...,n_s\}}$ are the forward minus reverse stoichiometric coefficients for species $(A_i)_{i\in\{1,...,n_s\}}$ in reaction $r$ (i.e. $\nu_{ir}>0$ if $A_i$
disappears in reaction $r$, $\nu_{ir}<0$ if $A_i$ appears in reaction $r$ and $\nu_{ir}=0$ if $A_i$ is not involved in reaction $r$).
In this case, the pressure is given by (see \cite{anderson})
\begin{equation}\label{eq:pT}
    \begin{dcases}
      p(\rho, \epsilon, \x) = \frac{\rho R T(\epsilon,\x)}{\sum_{i=1}^{n_s}x_im_i},\\
        T(\epsilon, \x) = \frac{\epsilon m - \sum_{i=1}^{n_s} x_i m_i h^0_i}{\sum_{i=1}^{n_s} x_i m_i {C_v}_i},\\
    \Delta G (\x, U) = 0. 
    \end{dcases}
\end{equation}
  In the above equations, $R$ is the universal gas constant, $(h^0_i)_{i\in\{1,...,n_s\}}$ and $({C_v}_i)_{i\in\{1,...,n_s\}}$ are respectively the mass enthalpies of formation and the mass heat
capacities at constant volume of the $n_s$ species. 
These expressions also involve their molar fractions and molar masses $\{x_i\}_{i \in \{1,...,n_s\}}$ and $\{m_i\}_{i \in \{1,...,n_s\}}$. 
Finally, the vector of mass fractions $\x$ is obtained by minimizing the Gibbs free energy $G$, which is implicitly recalled by the last equation of the system (\ref{eq:pT}), i.e.
{\em via} the fact that $\x$ cancels $\Delta G$, the differential of $G$.  
Note that the minimization of the Gibbs free energy depends on the vector of unknowns of the Euler system $U$ and on the molar fractions $\x$. In a nutshell, at equilibrium, we
must have 
$$
\Delta G(\x, U) = \Sum_{r=1}^{N_r}\Sum_{i=1}^{n_s} \nu_{ir} G_i(x_i, p, T)  = \Sum_{i=1}^{n_s} \nu_{i} G_i(x_i, p, T) = 0,
$$
where $(G_i)_{i\in\{1,...,n_s\}}$ are the Gibbs free energy of species $i\in\{1,...,n_s\}$ per mole of $i$, see \cite{anderson,mutation,scoggins} for more details.

Finally, the resolution of the whole strongly coupled system of equations can be summed up by $\forall t\in\mathbb{R}^+$ 
\begin{equation}
  \label{eq:sumup}
  \begin{dcases}
    \partial_t U(x,y,t) + \nabla \cdot F(U(x,y,t), \x(x,y,t)) = 0, &\forall x,y\in\mathcal{D},\\
      \Delta G (\x(x,y,t), U(x,y,t)) = 0, &\forall x,y\in\mathcal{D},\\
      U(x,y,t)=b(x,y,t), & \forall x,y\in\partial \mathcal{D},
  \end{dcases}
\end{equation}
in which the dependence on $\x$ is made explicit in the expression of the flux $F$.\\

Now, for our reentry problem, system (\ref{eq:sumup}) must be solved for long times (stationary problems).
In practice, a second order in time splitting is operated so that one iteration of the resolution, which is closely related to a time step $[t^n,t^{n+1}=t^n+\Delta t]$, consists in the resolution of 
\begin{equation}
  \label{eq:sumup_euler}
  \begin{array}{l}
  \begin{dcases}
    \partial_t U(x,y,t) + \nabla \cdot F(U(x,y,t), \x(x,y,t^n)) = 0, &\forall x,y\in\mathcal{D},\\
      U(x,y,t)=b(x,y,t), & \forall x,y\in\partial \mathcal{D}, 
  \end{dcases}
  \end{array}
\end{equation}
during time step 
$\forall t\in[t^n,t^{n+\frac12}]$, 
followed by the resolution of 
\begin{eqnarray}
  \label{eq:sumup_gibbs}
    \Delta G (\x(x,y,t^{n+1}), U(x,y,t^{n+\frac12})) = 0, &\forall x,y\in\mathcal{D},
\end{eqnarray}
during time step $\forall t\in[t^{n+\frac12},t^{n+1}]$.

In our simulation code, the Euler counterpart (\ref{eq:sumup_euler}) of the splitting is solved using the numerical scheme presented in \cite{peluchon, peluchon_jcp, peluchon_proceed}: it is
a Lagrange$+$remap scheme.
The main idea of the splitting is to separate the acoustic and dissipative phenomena
from the transport one.
In Low Mach computations (as we aim at treating liquid ablation in further work \cite{peluchon, latige}), an implicit treatment of the Lagrangian step is done
since the fast acoustic waves would induce very small time steps otherwise. 
The remapping step is explicit and performed with a finite volume scheme.
The overall scheme resulting from this splitting operator strategy is very robust,
conservative, and preserves contact discontinuities.

The Gibbs free energy minimization counterpart (\ref{eq:sumup_gibbs}) is solved thanks to the library Mutation++ \cite{mutation}. This library 
provides accurate and efficient computation of physicochemical properties
associated with partially ionized gases in various degrees of thermal nonequilibrium.
The users can compute thermodynamic and transport properties, multiphase linearly-constrained equilibria,
chemical production rates, energy transfer rates, and gas-surface interactions (i.e. Mutation++ is also a promising tool for other test cases than atmospheric reentry). The framework is based
on an object-oriented design in C++, allowing users to plug-and-play various models, algorithms, and
data as necessary. Mutation++ is available open-source under the GNU Lesser General Public License
v3.0.

Mutation++ allows performing accurate reference solutions for our reentry problem but remains 
costly for our needs. Remember the example of figure \ref{fig:field1}: we only consider the reaction
  $N + O \leftrightharpoons NO.$
In that case, there are only three species, $N$, $O$, and $NO$, and the computations are already computationally intensive. 
Those three species are far from being sufficient to characterize Earth's
atmosphere. A relevant set of species for Earth needs at least $n_s=18$ species  
and is even more computationally intensive (see
section \ref{capabilities}). Besides, it does not even take into
account the species ejected from the ablating surface of the object \cite{peluchon, latige, Yin2009}.\\ 

Let us present a sketch of the simulation code: algorithm \ref{reentry_code} presents the main steps of the resolution. We insist on the fact that algorithm \ref{reentry_code} 
certainly stands for a coarse description of the simulation code. But it is enough in order to present the methodology applied in this paper. It also testifies to the simplicity
of application of the material of this paper. 
\ \\ \ \\
    \begin{algorithm}[H]
    \caption{Core of the reentry code.}
        \label{reentry_code}

      \textit{\# general initialization (mesh, quantities on mesh etc.)}

      initialise\_guess\_vector\_of\_unknowns\_on\_mesh($U^0$,$\x^0$)

      \While {\text{convergence\_criterion\_not\_satisfied}}{

      $U^{n+\frac12}=$solve\_Euler\_equations($U^n$,$\x^{n}$)

      \For{$i\in\{1,...,N_{\mathcal{D}}\}$}{
      
        $\x_i^{n+1}, U^{n+1}=$minimize\_Gibbs\_free\_energy\_with\_mutation++($U^{n+\frac12}_i$,$\x^{n}_i$)
    }
      $U^n \leftarrow U^{n+1}$

      $\x^n \leftarrow \x^{n+1}$

                 }
    \end{algorithm}\ \\

First, in algorithm \ref{reentry_code}, a mesh has to be built and the different quantities must be initialized on this mesh. Of course, the closer to the stationary solution the initialization, the faster the
resolution in terms of iterations. In practice, we rely on uniformly initialized quantities which are certainly far from the solution to the problem. All the information is condensed in function
initialise\_guess\_vector\_of\_unknowns\_on\_mesh in line 2 of algorithm \ref{reentry_code}.
Then comes the {\em while} loop: a convergence criterion must be chosen but it is not central in this paper so we choose not to describe it. 
While convergence is not fulfilled, the code solves the Euler equations (function solve\_Euler\_equations) before feeding the updated field $U^{n+\frac12}$ into the minimization of the Gibbs free energy. 
Note that Navier Stokes' system could be solved instead of the Euler one in this paper, and this would not change the methodology described in the next lines. 
Note also that
the minimization must occur within each cell $i\in\{1,...,N_\mathcal{D}\}$ where $N_\mathcal{D}$ corresponds to the total number of cells.  
The minimization is made in the function minimize\_Gibbs\_free\_energy\_with\_mutation++, which is nothing more than a call of Mutation++. 
It takes as inputs $\rho$, the density, $\epsilon$, the mixture energy, and the mole fractions of the elements initially found in the fluid (i.e. the information contained in $U$). 
It outputs $\{\x_1,...,\x_{n_s}\}$ the mass fractions of the mixture of $n_s$ chemical species but also additional quantities such as $c$, the speed of sound, $C_p$ the heat at
constant pressure, $C_v$, the heat at constant volume, $p$ the pressure, and $T$ the temperature after the equilibrium of the reactions is fulfilled. 
In a sense, the call to Mutation++, denoted by $\mut$, can be summarized as a function of $\mathbb{R}^{n_e+2}$ in $\mathbb{R}^{n_s+5}$
\begin{equation}
\mut 
\;\;\;
: 
\;\;\;
    \begin{pmatrix}
        \x_1 \\
        ...\\
        \x_{n_e}\\
        \rho\\
        \epsilon
    \end{pmatrix}
    \in \mathbb{R}^{n_e+2} \;\;\;
    \longrightarrow
    \;\;\;
    \begin{pmatrix}
        \x_1 \\
        ...\\
        \x_{n_s}\\
        P\\
        T\\
        C_p\\
        C_v\\
        c
    \end{pmatrix}, \in \mathbb{R}^{n_s + 5},
\end{equation}
with the outputs such that $\Delta G=0$.
As highlighted by the example of section \ref{intro}, the minimization of the Gibbs free energy is necessary for model accuracy but very costly.
We would like to build a surrogate model to replace the call to Mutation++ by approximating $\mut$, just as in algorithm \ref{reentry_code_surrogate}, and hopefully accelerate the reentry code without impacting its accuracy. \\ \ \\
    \begin{algorithm}[H]
    \caption{Core of the code with a call to a surrogate model of Mutation++.}
        \label{reentry_code_surrogate}

      \textit{\# general initialization (mesh, quantities on mesh etc.)}

      initialise\_guess\_vector\_of\_unknowns\_on\_mesh($U^0$,$\x^0$)

      \While {\text{convergence\_criterion\_not\_satisfied}}{

      $U^{n+\frac12}=$solve\_Euler\_equations($U^n$,$\x^{n}$)

      $\x^{n+1}, U^{n+1}=$call\_surrogate\_model($U^{n+\frac12}$,$\x^{n}$)

      $U^n \leftarrow U^{n+1}$

      $\x^n \leftarrow \x^{n+1}$

                 }
    \end{algorithm} \ \\
Obviously, for efficiency, the surrogate model has to be well-chosen. 
The classical reflex at this stage would be to build and use some abacuses, as {\em offline} calls to Mutation++ can be made. 
This is classical for tabulated equations of state for example. 
Those abacuses can then be loaded into memory and interpolated during the simulation \citep{artpred1,artpred2}. 
However, this method becomes intractable when the input/output dimensions increases (here we have $d_\text{in}=n_e+2\gg 1$ or $d_{\text{out}}=n_s+5 \gg 1$, see section \ref{capabilities}) because the number of points $N$ needed to obtain a fine interpolation increases exponentially fast with it, together with
a complexity for the search in the database which strongly depends on $N$ too.
For this reason, in the following section, we study the possibility of building different surrogate models of Mutation++ from
gathered data. In particular,  we
explain why we are interested in neural networks. As can be seen with algorithm \ref{reentry_code_surrogate}, our methodology is
{\em intrusive}, we need to modify a few lines of the simulation code as we are going to plug a neural network in it, hence the {\em hybrid} denomination.

\section{Neural networks as approximators for hybridization}
\label{ml}


In the previous section, we formalized the problem of replacing Mutation++ with a surrogate model as an approximation problem.
In section \ref{surrogates}, we investigate the different types of potential surrogate models allowing us to reach our needs and explain why we consider neural networks.
In section \ref{capabilities}, we study the capabilities of acceleration of neural networks on several benchmarks/atmospheres and verify their behaviors in terms of complexity with respect to
the input and output dimensions ($d_{\text{in}}$ and $d_{\text{out}}$). Neural networks are promising (see section \ref{capabilities}), but their design has a strong impact on the final performances of the hybrid code: section \ref{sec:hsic} is devoted to
explaining how we look for {\em accurate and cost-effictive} neural networks.

\subsection{Many possible classical surrogate models}
\label{surrogates}

With the last paragraph of section \ref{cfd}, we formalised our problem as approximating a function of $X\in \Rin \longrightarrow u(X)\in\Rout$ from $N$ available data
$(X_i,u(X_i))_{i\in\{1,...,N\}}$. Of course, for our application, $u$ is nothing more than the call to $\mut$.  
In other words, we face an {\em approximation theory} problem.
In approximation theory (and in ML, which largely intersects with this field), it is classical to look for a parametric function 
$$(X,\theta)\in \Rin \times \mathbb{R}^{d_\theta}
\longrightarrow u(X, \theta) \in\Rout,$$ 
which has to be the closest possible to $u(X)$ in a certain metric $L$. This goal is achieved with the optimization of $\theta$ driven by the minimization of 
$$
J(\theta) = \Int L(u(X), u(X,\theta))\dPx,
$$
where $\dPx$ is the measure of the input space. In practice, we do not have access to $J(\theta)$ because the measure $\dPx$ is unknown. Hence, we approximate $J(\theta)$ with an experimental design $(X_i, w_i)_{i\in\{1,...,N\}}$, which is a discretisation\footnote{In the sense that $\forall f\in L_2$, $\sum_{i=1}^N w_i f(X_i) \overset{L_2}{\underset{N\rightarrow
\infty}{\longrightarrow}} \int f(X)\dPx$.} of $(X,\dPx)$, as: 
\begin{eqnarray}
  \label{J}
  J(\theta) = \Int L(u(X), u(X,\theta))\dPx \approx J_N(\theta)=\Sum_{i=1}^N w_i L(u(X_i),u(X_i,\theta)),  
\end{eqnarray}
where $X$ is a random vector (of potentially correlated components) and $\dPx$ is its probability measure. 

There exist many different types of surrogate models. Amongst the most classical ones in numerical and uncertainty analysis, we can count\footnote{The lists of
    references in the following points are not exhaustive and have been chosen because of the proximity of their application domain.} 

\paragraph{Lagrange interpolation or collocation} (and higher order ones such as Hermite interpolation) \cite{LoevenWitteveenECCOMAS,NobTempWeb,XiuHesthaven,LoevenBijl,
    WitteveenBijl, GanaZaba, runge_ex}: they are based on the choices
    \begin{itemize}
      \item $L(x,y)=(x-y)^2$ in \eqref{J},
      \item $u(X,\theta) = \Sum_{k=0}^{P}\theta_k X^k$, i.e. a polynomial approximation and a linear application $\theta \rightarrow u(X,\theta)$ with respect to variable $\theta$.  
      \item together with $N=P+1=d_\theta$.
    \end{itemize}
    This ML model ensures $u(X_i)=u(X_i,\theta)$ $\forall i\in\{1,...,N\}$. Besides, spectral convergence can be achieved \cite{XiuHesthaven}, hence very good accuracies, but the convergence behavior strongly depends on the choice of the experimental design $(X_i,w_i)_{i\in\{1,...,N\}}$ discretising
    $(X,\dPx)$ (see the divergence of the approximation of Runge's function with uniform points \cite{cheneycourse, runge_ex}).
    Finally, if $N$ is huge (i.e. in a big data context), $P\sim N$ is huge and the run-time of the ML model strongly depends on the size of the database. It may become prohibitive
    for the desired accuracy.  
    \paragraph{Polynomial regression and generalised Polynomial Chaos} \cite{PC_Wiener, chaospol,sparsepce,PC_LemaitreHaar, Walters,PC_LucorPred,PC_Poette,
    PC_LucorEnaux,PC_congedo_optimization,sudret_global,PC_LucorJFM07, WanKar04} are very popular, especially in problems of Uncertainty Quantification (UQ). Since the seminal work of
    \cite{PC_LemaitreHaar}, it is extensively used in (non-intrusive) uncertainty propagation \citep{Walters,PC_LucorPred,PC_Poette, PC_LucorEnaux,PC_congedo_optimization, Poette_igpc, PoetteCicp, WanKar04}.
    These ML models are based on the choices:
    \begin{itemize}
      \item $L(x,y)=(x-y)^2$ in \eqref{J},
      \item $u(X,\theta) = \Sum_{k=0}^{P}\theta_k \phi_k(X)$, is linear with respect to $\theta$ and $(\phi_k(X))_{k\in\{0,...,P\}}$ are orthonormal polynomials with respect to the scalar product defined by
        $\dPx$.
      \item The number of parameters $d_{\theta} = P+1$ is not constrained by $N$. 
    \end{itemize}
    Spectral convergence with respect to $P$ is ensured \cite{ErnstUllman}. 
    The orthonormality of the basis helps with round-off errors and conditioning \cite{poette_revuq} while having a model for which $N$ and $P$ are not correlated anymore (i.e. we
    can take $N\gg P$, big datasets, with a model having a run-time depending on $P \ll N$).

    \paragraph{Gaussian Process regression or Kriging} Popularized for ML by \cite{gprasmussen} and intensively used in Uncertainty Quantification (UQ) \cite{BayesianSA,
    SA_daveiga_GP, GPMarrel, KOH, gramacy2020surrogates, BachocPhD, PapierBachoc, poette_gpc_kriging, BO_blackbox, BO_design}, this technique has established to a leading position in surrogate modeling. It can be summed-up as taking 
    \begin{itemize}
      \item $L(x,y)=(x-y)^2$ in \eqref{J},
      \item $u(X,\theta) = \Sum_{k=0}^{P}\theta_k \phi_k(X) + Z(\theta_{P+1},...,\theta_{d_\theta})$, 
        \begin{itemize}
          \item where $(\phi_k(X))_{k\in\{0,...,P\}}$ can be orthonormal polynomials with respect to the scalar product defined by
        $\dPx$ as in \cite{SudretPCK, SudretPCK2, SudretPCK3, poette_revuq} or classical polynomials \cite{BachocPhD},
      \item and where $Z$ is a gaussian process conditionned to satisfy
        $u(X_i)=u(X_i,\theta)$ $\forall i\in\{1,...,N\}$. Some particular shapes of $Z$ are {\em a priori} determined by choosing particular covariance functions for the process
        \cite{BachocPhD,PapierBachoc,PapierBachoc2}. 
      \item The ML model is linear with respect to $(\theta_0,...,\theta_P)$ and nonlinear with respect to $(\theta_{P+1},
        ...,\theta_{d_\theta})$.  
        \end{itemize}
      \item The number of parameters $d_{\theta}$ is not constrained by $N$. 
    \end{itemize}
    This type of ML model can be understood as a way to make the best of the two previous approaches as we have $u(X_i)=u(X_i,\theta)$ $\forall i\in\{1,...,N\}$ together with convergence
    properties for the mean \cite{SudretPCK, SudretPCK2, SudretPCK3, poette_revuq} and with having $N> d_{\theta}$. But the run-time of such an ML model still strongly depends on $N$.
    For huge databases, these models can be very accurate but far from being cost-effective.

\subsection{The advantages of neural networks}

This brings us to the last type of approximator we consider in this paper: (deep or shallow) neural networks. They are nonlinear approximators leading to a non-convex loss function $J$ \cite{goodfellow, math_of_deep}. They are based on the choices:
        \begin{itemize}
          \item $L(x,y)$ can be general ($L_2$-norm, $L_1$-norm, cross-entropy etc., see \cite{goodfellow}).
          \item For a shallow neural network\footnote{Bias are taken into account with this notation, $u(X,\theta) = \Sum_{k=0}^{P}\theta_k^1 \sigma(\theta^2 \cdot X)$, see \cite{math_of_deep}.}, which is
            nonlinear with respect to $\theta=(\theta^1,\theta^2)$. The function $\sigma : \mathbb{R} \rightarrow \mathbb{R}$ is called the {\em activation function}. In practice, it only has to be unbounded and non-constant \cite{Hornik, Barron}. 
          \item Deep neural networks with $L$ layers correspond to $L$ compositions of the above expression: an $L-$layer feed forward neural network is defined recursively as  
            $d_\theta = 2(P+1)\times L$ parameters\footnote{Or
            $d_\theta = 3(P+1)\times L$ parameters if bias a considered \cite{math_of_deep}.} and rely on the recursive formula: 
            $$
            \begin{array}{l}
              u^{0}(X,\theta) = X,\\
              u^{l}(X,\theta) = \Sum_{k=0}^P \theta_k^{l} \sigma \left(\theta^{l-1}u^{l-1}(X,\theta)\right), \forall l\in\{1,...,L-1\}, \\
              u(X,\theta) = \theta^{L-1} u^{L-1}(X,\theta), \\
            \end{array}
            $$
            where $\sigma$ is applied element-wise. Note that in this formula, the number of neurons per layers $P=\nunits$ is considered constant but may change with $l\in\{1,...,L=\nlayers\}$.
          \item The number of parameters $d_{\theta}$ is not constrained by $N$. 
    \end{itemize}
The convergence of the approximation is guaranteed under the hypothesis of Hornik's theorem \cite{Hornik} (or Barron's one \cite{Barron}) as the number of neurons grows; 
or under the hypothesis of \cite{LuPuWangRelu} for deep neural networks as the number of layers $L$ grows. In each of the previous theoretical results, the existence of a set of parameters ensuring convergence is guaranteed. Yet, a difference remains with the other approaches in that the loss function $J$ (and $J_N$) may have a lot of local minima \cite{plein_de_minima}, and we consequently have to find the set of parameters ensuring convergence. 

Neural networks share many similarities with classical surrogate models. However, they differ on several points.
\begin{itemize}
 \item For {\em classical surrogate modeling}, some pre-processing is needed in order to take into account correlated input variables \cite{Lebrun, Lebrun1, Dutfoy}. It can be a problem since training data may come from previous simulations and uncontrolled (possibly correlated) distributions.
 \item In {\em classical surrogate modeling}, when the space of output is of size $d_{\text{out}}$, $d_{\text{out}}$ surrogate models must be built. This is problematic when the training and inference time of the model increase. For {\em neural networks}, only the last linear layer depends on $d_{\text{out}}$.
  \item {\em Classical surrogate models} are very sensitive to the curse of dimensionality as the number of parameters $d_\theta$ may grow exponentially fast with both $P$ and $d_\text{in}$. As a consequence, these models are not well suited to high-dimensional problems. {\em Neural networks} complexity scales linearly with $d_\text{in}$, and is famous for its recent breakthrough on high dimensional test cases like image or text processing. In addition, the complexity of one prediction is independent of $N$, so we can leverage very large databases.
  \item The implementation of the inference of {\em neural networks} boils down to a succession of matrix-vector products, which can be easily vectorized. As a result, they can process array-like data structures very efficiently, which makes them perfectly suited to computations on meshes. {\em Classical surrogate models} do not offer such implementation properties.
\end{itemize}

In the next section, we introduce a profiling experiment that gives an idea about the potential computational gains of neural network-based hybridization. 

\subsection{Assessing the capabilities  of neural networks in terms of accelerations}
\label{capabilities}

To test the potential computational gains of hybridization, we construct a simple benchmarking code that compares the run-times of neural networks and Mutation++ for a given number of input points. In our case, this number, $N_{\mathcal{D}}$, is the size of the simulation mesh.
This code (C++) only corresponds to the extraction of the {\em for} loop of algorithm \ref{reentry_code} for Mutation++ and of the call\_surrogate\_model of algorithm
\ref{reentry_code_surrogate}. The neural networks are implemented within this code {\em via}  the Tensorflow C API and a wrapper,
CppFlow\footnote{\url{https://github.com/serizba/cppflow}}.

In the next studies, the neural networks have $\nlayers=5$ hidden layers, and the number of neurons in each layer $\nunits$ is constant per layer and chosen as a parameter of the code. 
In other words, we only study the influence of hyperparameter $\nunits$, the number of neurons per layer, even if many others exist (dropout rate, the different architectures, and their
parameters, the
different optimizers and their parameters, etc. see \cite{goodfellow}) together with the influence of operational conditions $N_{\mathcal{D}}\in\{10, 10^2, 10^3, 10^4, 10^5,
10^6\}$, the numbers of cells of the grid and $n_s\in\{3,18, 38, 
64\}$, the number of
species. In order to study the influence of $n_s$, we consider $4$ different atmospheres (arbitrarily constructed out of the elements found in each atmosphere to obtain increasingly complex test cases): 
\begin{itemize}
  \item Toy problem: the toy problem corresponds to the conditions mentioned in section \ref{intro} with $2$ elements and $3$ species $N,O,NO$.
  \item Earth: it corresponds to the case where the fluid is air with $2$ elements but where $18$ species are considered: 
   $N$, $NO$, $O$, $N_2$, $O_2$, $e^-$, $N^+$, $O^+$, $N_2^+$, $O_2^+$, $NO^+$, $NO_3^-$, $NO_3$, $NO_2$, $O_3$, $NO_2^-$, $O^-$, $O_2^-$, i.e. $n_s=18$ species. 
  \item Cloudy Earth: it corresponds to the same test case as above but where another element is considered, $H$, coming from the clouds that the object can meet. Additional species are therefore considered, for a total of $38$ species: 
$N$, $NO$, $O$, $N_2$, $O_2$, $e^-$, $N^+$, $O^+$, $N_2^+$, $O_2^+$, $NO^+$, $H$, $OH$, $NH$, $H^+$, $OH^+$, $NH^+$, $H_2O^+$, $H_2O$, $H_2$, $H_2^+$, $NH_3$, $NO_3^-$, $NO_3$,
    $NH_4^+$, $H_3O^+$, $NO_2$, $N_2H_2$, $H^-$, $HNO$, $O_3$, $HNO_2$, $HNO_3$, $NO_2^-$, $O^-$, $OH^-$, $NH_2$, $O_2^-$. 
  \item Cloudy Jupiter: on Jupiter, the clouds are made of water but also of ammonium hydrosulfide and ammonia. Hence, $N$ and $S$ are added as input elements, and $64$ species are considered: 
$O$, $O_2$, $C$, $e^-$, $C^+$, $O^+$, $O_2^+$, $CO^+$, $C_2$, $CO$, $CO_2$, $H$, $CH$, $OH$, $H^+$, $CH^+$, $OH^+$, $H_2O^+$, $H_2O$, $H_2$, $H_2^+$, $CH_4$, $He$, $He^+$, $CH_2$, $H^-$, $HCO^+$, $CH_3$,
    $C_2H$, $HCO$, $C^-$, $O^-$, $OH^-$, $C_2^-$, $O_2^-$, $C_2O$, $N$, $NO$, $O$, $N_2$, $O_2$, $e^-$, $N^+$, $O^+$, $N_2^+$, $O_2^+$, $NO^+$, $NH_3$, $NO_3^-$, $NO_3$, $NH_4^+$, $NO_2$, $N_2H_2$, $H^-$,
    $HNO$, $O_3$, $HNO_2$, $HNO_3$, $NO_2^-$, $NH_2$, $S$, $S^+$, $S^-$, $CS$, $CS_2$, $COS$, $CNCOCN$, $CN$, $CN^+$, $CN^-$.
\end{itemize}
The cloudy Jupiter scenario may appear far fetched but considering it allows progressively increasing the number of species $n_s$ from $3$ (toy), $18$ (Earth), $38$ (cloudy
Earth) 
 to $64$ (cloudy Jupiter) and study the impact of the number of outputs of the neural networks.
 Moreover, with $64$ species, the {\em cloudy Jupiter} scenario is close to some operational conditions in which the species of the atmosphere are mixed with some from the ablating surface of the entering object.

\begin{figure}[!t]
  \centering
  \begin{subfigure}{0.49\textwidth}
    \begin{center}
      \includegraphics[width=1\textwidth]{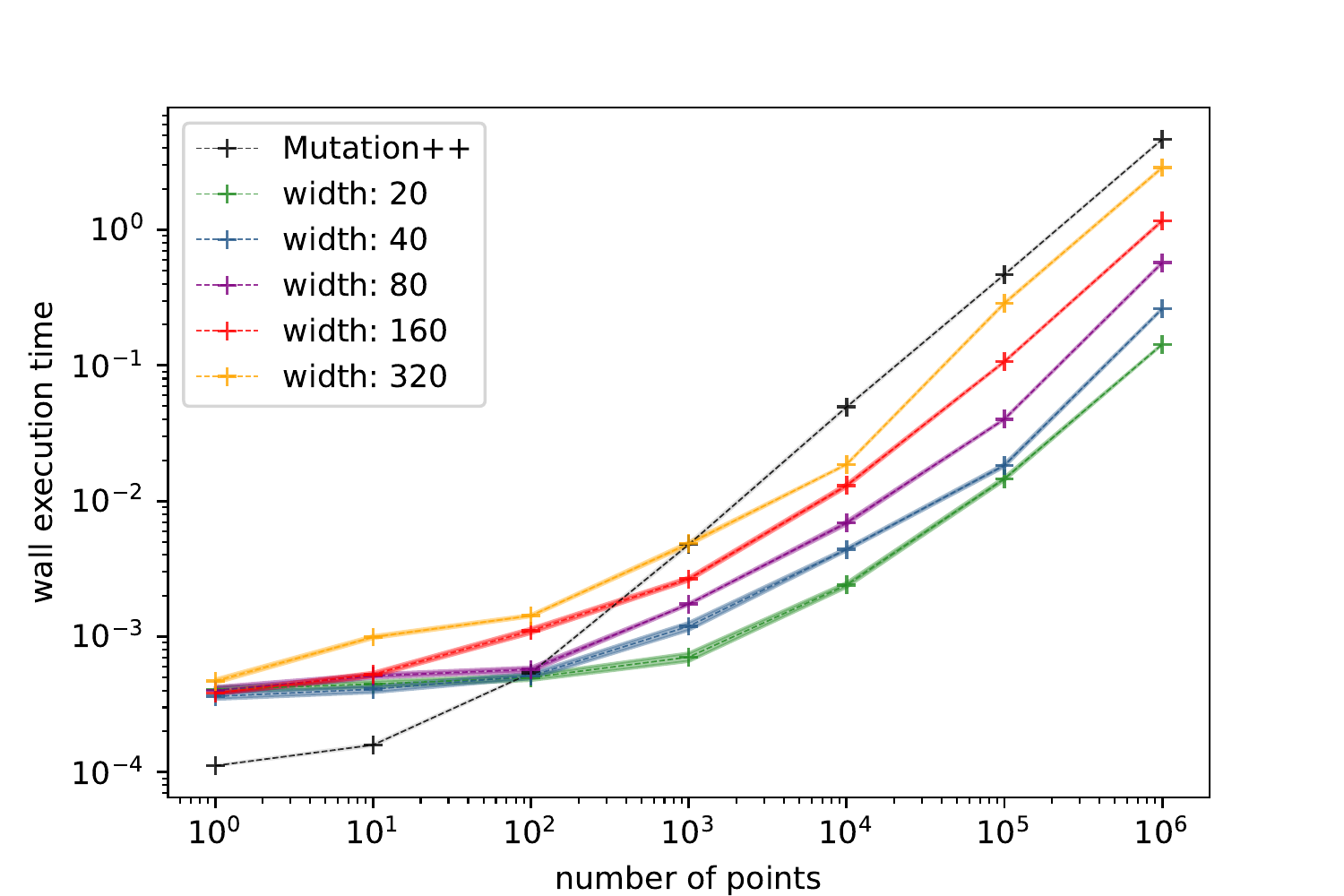}
    \end{center}    
\caption{Toy test case}
  \end{subfigure}
  \centering
  \begin{subfigure}{0.49\textwidth}
    \begin{center}
      \includegraphics[width=1\textwidth]{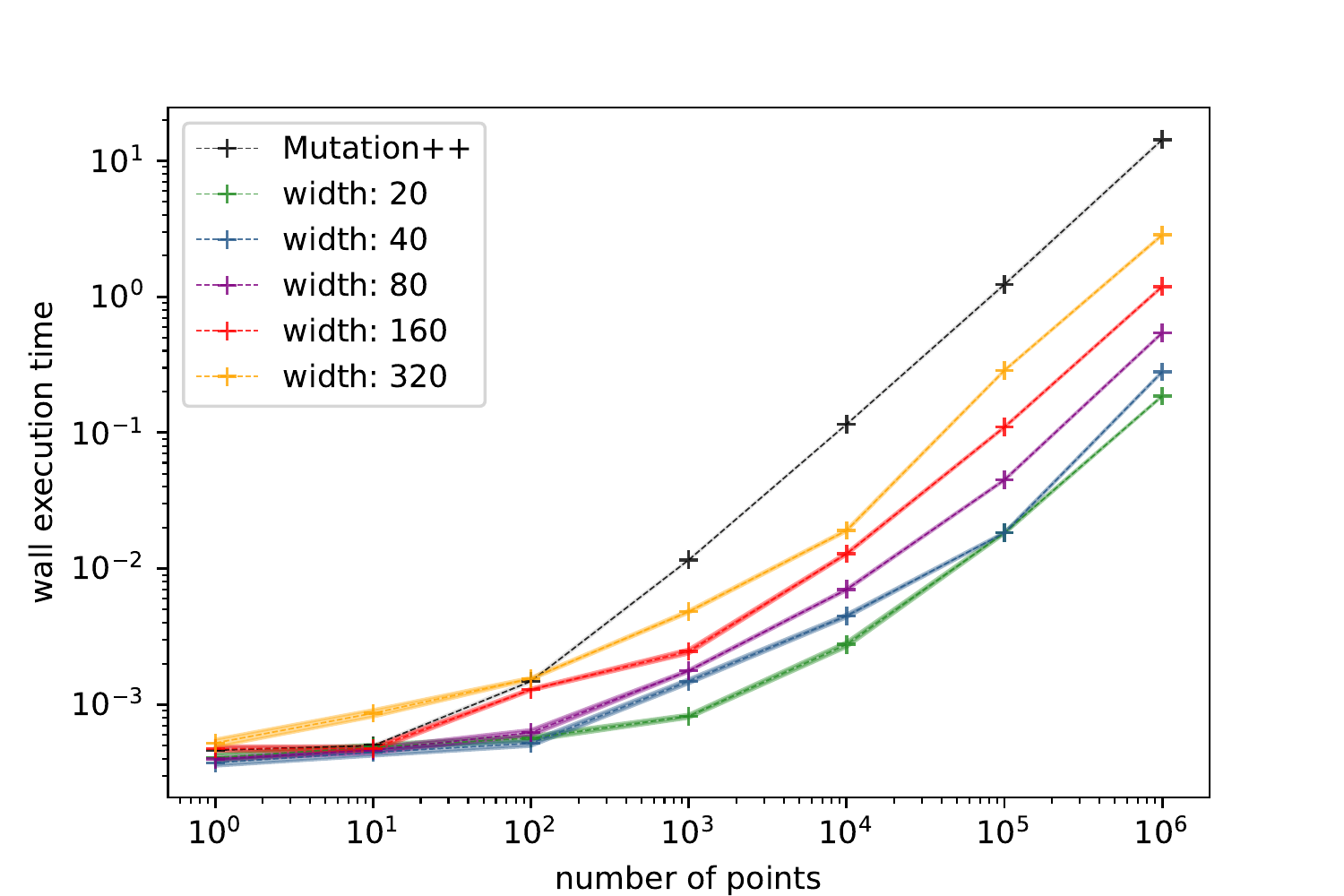}
    \end{center}    
\caption{Earth test case}
  \end{subfigure}
  \centering
  \begin{subfigure}{0.49\textwidth}
    \begin{center}
      \includegraphics[width=1\textwidth]{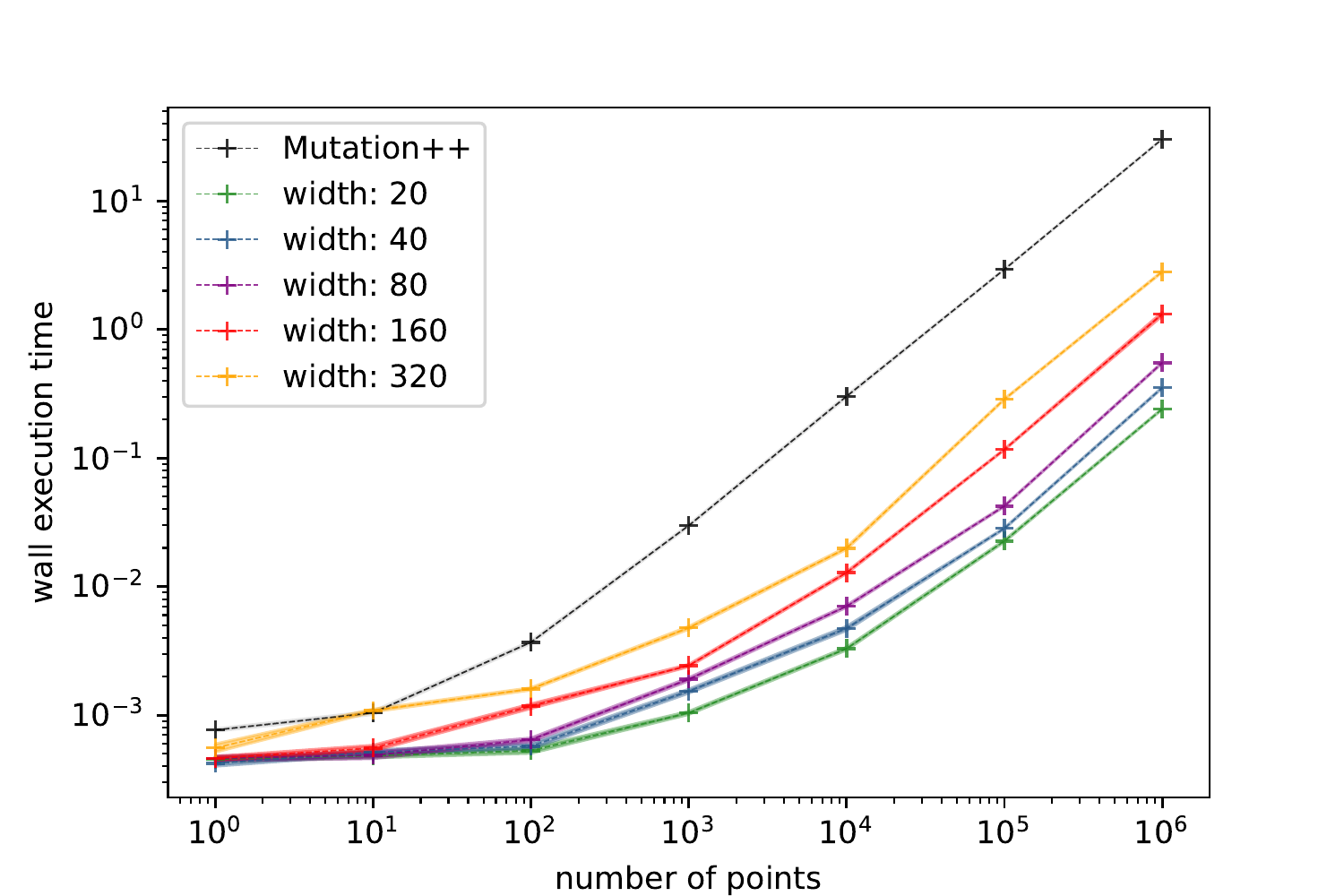}
    \end{center}    
\caption{Cloudy Earth test case}
  \end{subfigure}
  \centering
  \begin{subfigure}{0.49\textwidth}
    \begin{center}
      \includegraphics[width=1\textwidth]{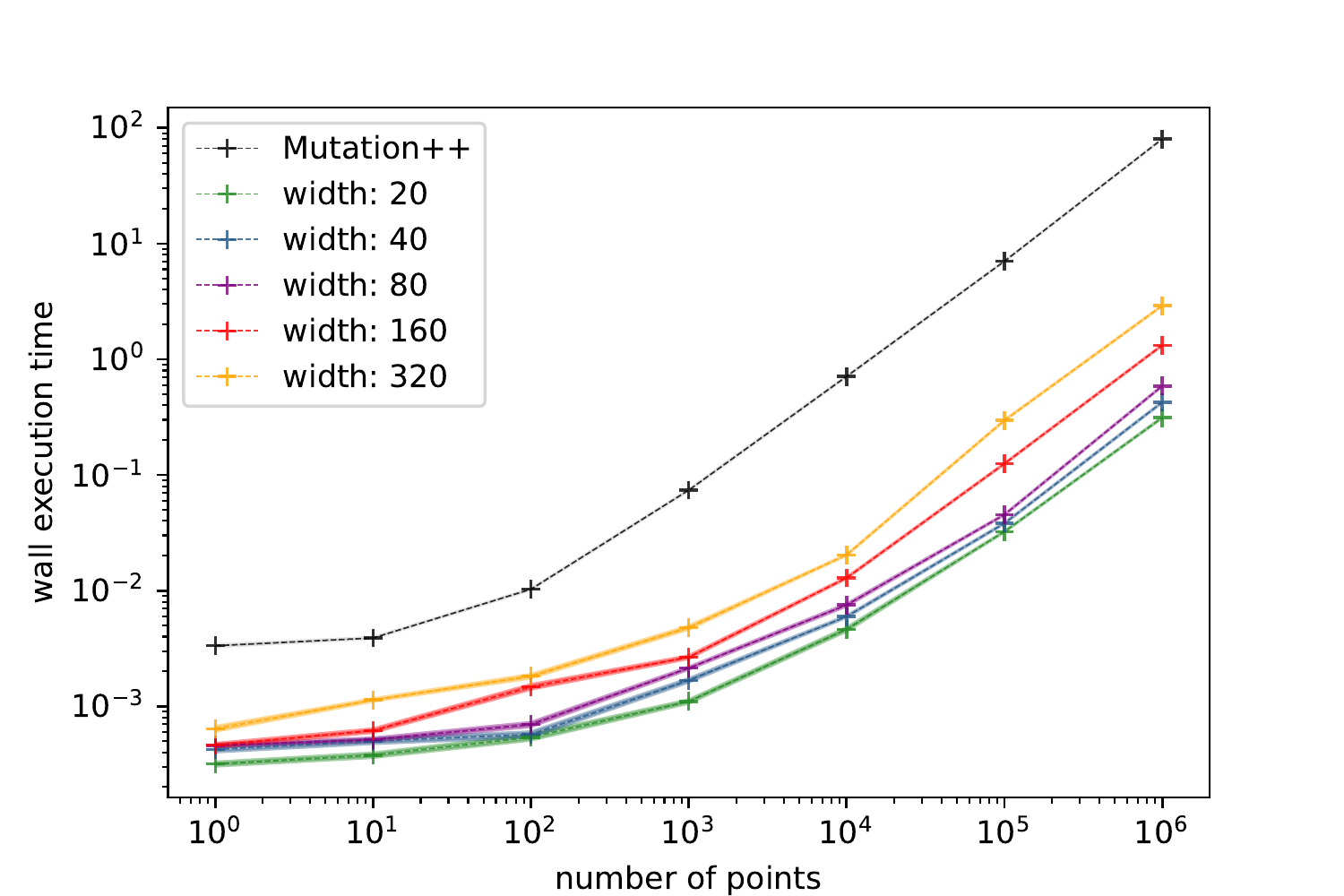}
    \end{center}
\caption{Cloudy Jupiter test case}

  \end{subfigure}
  \caption{Execution time of Mutation++ vs neural networks of different widths for each test case with respect to the number of input points (with log axes).}\label{fig:testcasevsmpp}
\end{figure}

On Figure \ref{fig:testcasevsmpp}, we plot the run-times of Mutation++ and neural networks of different widths $\nunits\in\{20, 40, 80, 160, 320\}$ for each
atmospheres with respect to the number of grid points $N_{\mathcal{D}}$. 
Note that $10$ repetitions are carried out for each curve to check for their stability (standard deviations are plotted, but they are barely visible due to the low value of the variance).
First, for $N_{\mathcal{D}}=10^0=1$, we can compare the {\em sequential} run-times of Mutation++ and of the neural networks for the different atmospheres: the neural networks are not
always faster than the calls to Mutation++.  
However, within the code, Mutation++ is called sequentially, while the neural networks are executed on the whole array of cells in a batch fashion. Figure \ref{fig:testcasevsmpp} illustrates how neural networks exploit vectorial acceleration: as the number of cells $N_{\mathcal{D}}$ increases, their run-times
become competitive with respect to Mutation++. In (a) it happens at $N_{\mathcal{D}}>10^3$. Of course, for some test cases, neural
networks are competitive even sequentially (in (c) and (d)).
In a general manner, for $N_{\mathcal{D}}$ between $10^4$ and $10^5$ - which turns out to be the orders of cells numbers per mesh block ($4000$ and $12000$ in our experiment of section \ref{numres}), neural networks performs much better than Mutation++ in terms of computational time. 

\begin{figure}[!h]
  \centering
  \begin{subfigure}{0.49\textwidth}
    \begin{center}
      \includegraphics[width=1\textwidth]{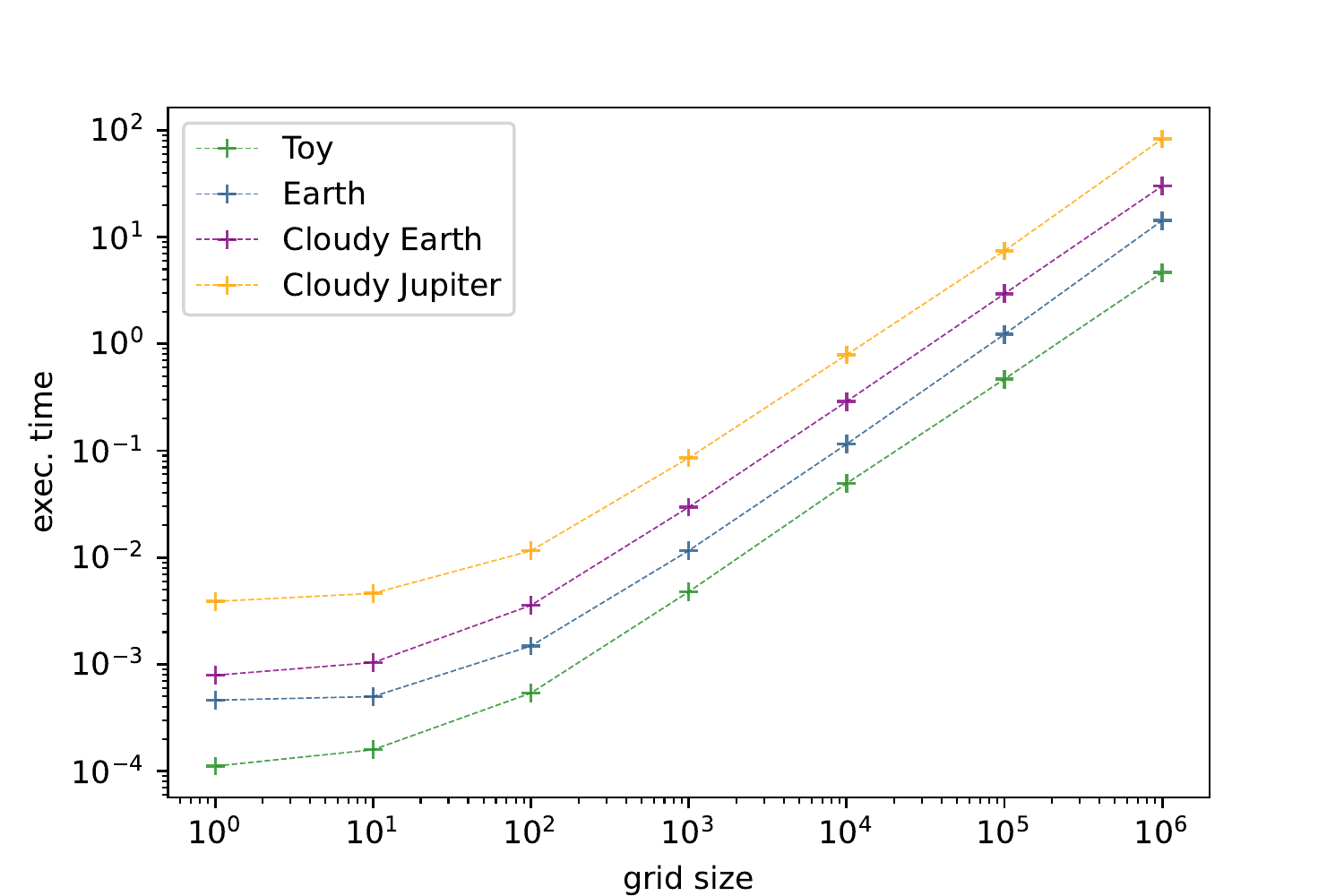}
    \end{center}  
\caption{Test cases with Mutation++}
  \end{subfigure}
  \begin{subfigure}{0.49\textwidth}
    \begin{center}
      \includegraphics[width=1\textwidth]{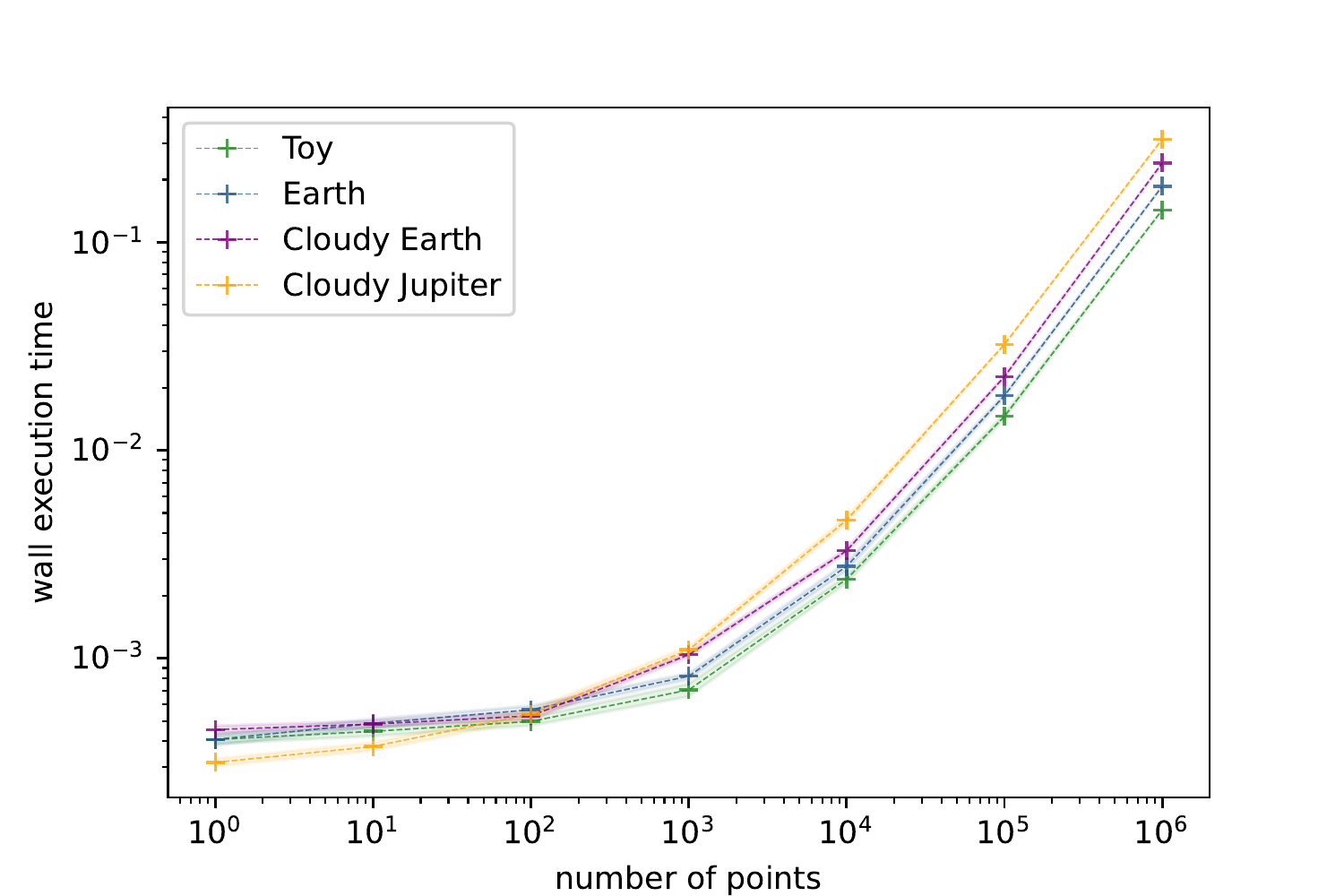}
    \end{center}  
\caption{Test cases with a network of width 20}
  \end{subfigure}
  \centering
  \begin{subfigure}{0.49\textwidth}
    \begin{center}
      \includegraphics[width=1\textwidth]{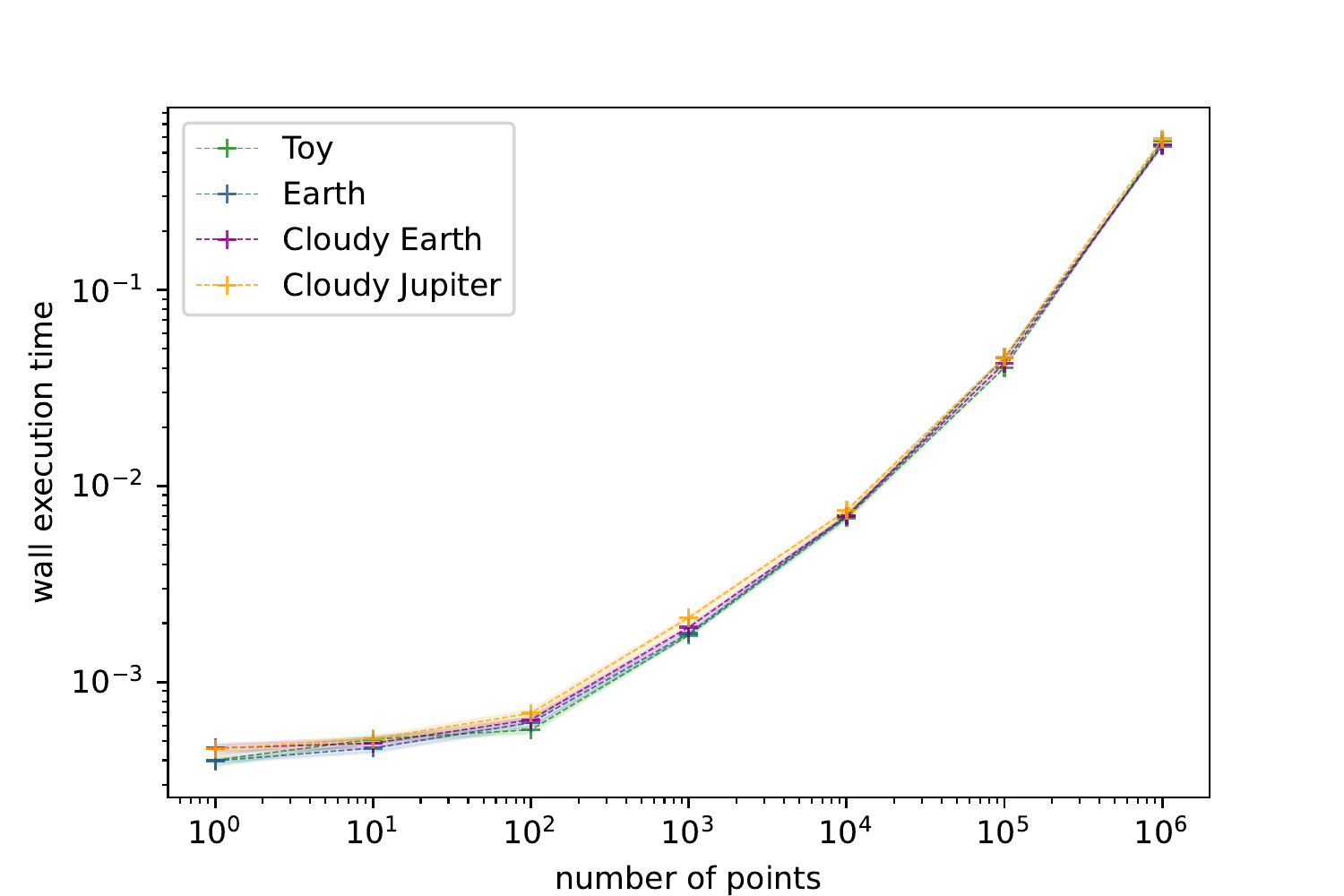}
    \end{center}  
\caption{Test cases with a network of width 80}
  \end{subfigure}
  \caption{Execution time of Mutation++ and a neural network of a given width for the different test cases with respect to the number of input points (with log axes).}\label{fig:testcaseswidth}
\end{figure}

In Figure \ref{fig:testcaseswidth} we visualize the run-times of neural networks and Mutation++ run-times with respect to $n_s$. We can see that the higher the number of species, the higher the run-time of Mutation++. This effect is less marked for the neural networks. For conciseness, we only present the plots for $\nunits$ of $20$ and $80$, which show that even for low width, the effect of $n_s$ on the run-time is limited, and above a width of $80$, it can be hardly distinguished. We recover experimentally the fact that the computational complexity of the operations between the hidden layers is insensitive to the dimensions of the problem. The input and output dimensions can be increased with a limited impact on the run-time, which illustrates how neural networks mitigate the curse of dimensionality.

\begin{figure}[!h]
  \centering
  \includegraphics[width=0.6\linewidth]{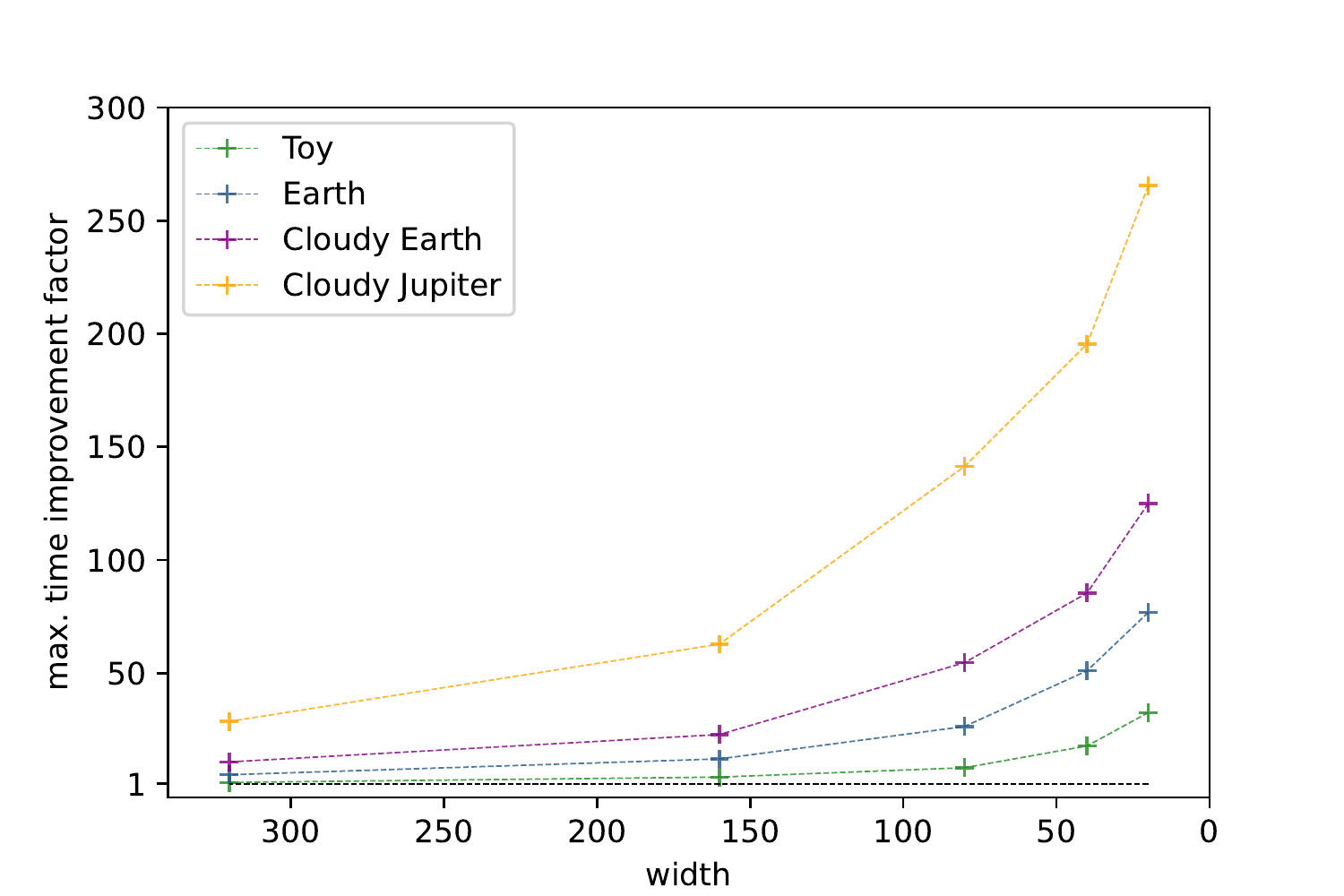}
  \caption{Gains in terms of run-times of neural networks compared to the ones of Mutation++, with respect to the width $\nunits$.}
  \label{fig:factors}
\end{figure}

Figure \ref{fig:factors} summarizes the gain factor that we can hope for each test case, with respect to $N_{\mathcal{D}}$ and $n_s$. It also emphasizes that the highest the number of species $n_s$, the more important the gain with factors going up to $\times 275$ for $\nunits=20$ for {\em cloudy Jupiter}'s atmosphere. Of course, this factor of
gain will be relevant only if a good accuracy can be reached for moderate width. It motivates the next section, which deals with the optimization of neural networks' hyperparameters that have a strong influence on both their accuracy and cost-efficiency.

The following section describes how we design neural networks to maximize this computational gain factor while maintaining a good approximation accuracy.

\section{Design of accurate and cost-effective neural networks with goal-oriented sensitivity analysis of hyperparameters}
\label{sec:hsic}

In this section, we emphasize the impact of neural networks' hyperparameters on their cost-efficiency and accuracy. To that end, we consider the approximation of Mutation++ in
the conditions of the atmosphere of the toy problem described in section \ref{capabilities}.

The hyperparameters involved in the training are given in \textbf{\hyperref[appB]{Appendix B}}. 
We also use the work of \cite{vbsw}, which studies the link between the neural network error and the variance of the output to learn. They define a sampling scheme and its weighting counterparts to account for this variance and improve the error. 
In the previous section, we explained we want to learn as much as possible from operational conditions of ($\rho, \epsilon$). But for this toy problem, relying on the previous independent probability measures gives accurate enough models (as will be seen later on) and considerably eases the reproducibility of the results of this paper.  
The neural networks are trained using Tensorflow in python on a training dataset of $170000$ points, and the hyperparameters are selected using a test set of $20000$ points. The training and test sets are constructed by sampling $\rho$ and $\epsilon$, respectively, uniformly and log-uniformly within the intervals $[0.1,
3.8] \times [2.07503 \times 10^7, 3 \times 10^8]$, defined based on the execution of the simulation code without chemical reactions. 

\begin{figure}[!h]
  \centering
  \includegraphics[width=0.6\linewidth]{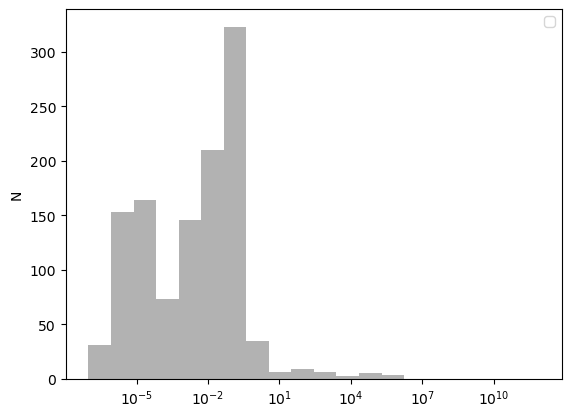}
  \caption{Histogram of the normalized $L_2$ validation error of the neural networks for approximating $\mut$ in the toy problem configuration when performing a random search (with
  $N_{MC}=3\times 10^3$ samples of the hyperparameters).}
  \label{fig:mutationRS}
\end{figure}

Figure \ref{fig:mutationRS} displays the histogram of the $L_2$ errors obtained after a random search \cite{random-search}, i.e. a uniform Monte Carlo sampling
of $N_{MC}=3\times 10^3$ samples on the hyperparameter space described. It shows several important properties: 
\begin{itemize}
  \item first, there is a non-zero probability of having poor performance in terms of accuracies: the L$_2$ errors go up to $\approx 10^6$. For this reason, it is important
    performing an optimization of the hyperparameters and not only rely on one test.
  \item The errors within the range $[10^{-3}, 10^{-1}]$ are more probable than others, but they are certainly not enough for our reentry application. 
  \item Finally, there is a non-zero probability of having very good accuracies, with errors going down to $10^{-8}$. 
\end{itemize}
In a nutshell, neural networks allow a wide range of errors (order $10^{10}$ between the lowest and the highest) depending on the hyperparameters use depending on the hyperparameters used. 
For this problem, the best error (normalized $L_2$) is $9.37 \times 10^{-8}$, which is lower than the (double) round-off error and is consequently promising. The question now is: is
such an accurate neural network cost-effective and competitive with respect to a call of Mutation++? 
The neural network yielding the previous performances has depth $\nlayers = 9$ and width $\nunits = 191$ units, which is close to the upper boundary of the search space for these hyperparameters.
Using such depth and width could significantly affect the expected cost efficiency improvement. Indeed, in figure \ref{fig:factors}, with $\nlayers=5$ and $\nunits=191$, only a gain of approximately $7$ is achievable.

Now, in histogram \ref{fig:mutationRS}, there are other values of errors that are probably acceptable (for example, errors in the range $[10^{-8}, 10^{-6}]$). The question is: are there cost-effective neural networks allowing us to reach such errors? 

In order to answer this question, we rely on the work of \cite{hsic:novello}, based on papers \cite{hsicgretton,
SA_daveiga_GP}: 
  In \cite{hsic:novello}, the authors study the use of goal-oriented sensitivity analysis, based on the Hilbert-Schmidt Independence Criterion (HSIC), for hyperparameter analysis
  and optimization. They design a robust analysis index that is able to quantify hyperparameters’ relative impact on an NN’s final error. This tool allows a better understanding of
  the hyperparameters' effects on both the error and the run-time. 
It is able to identify which hyperparameter(s) is (are) responsible for explaining the lowest errors. 
Once this/these are identified, it allows focusing the optimization on the hyperparameters having a significant impact on the error while considering the constraints of having viable run-times. 
The application of the methodology only consists of quick post-processing of the random search results. 
It can give fast and accurate insights from the previously performed random search. 
We do not give more details on the matter of paper \cite{hsic:novello}; we only present the results and the interpretations.   
\begin{figure}[!h]
  \centering
  \includegraphics[width=0.6\linewidth]{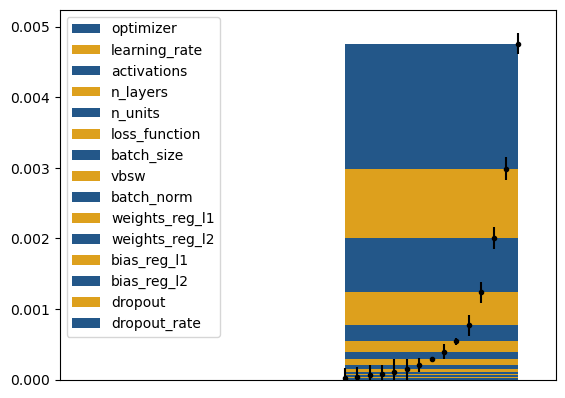}
  \caption{Goal-oriented sensitivity indices of the hyperparameters as in \cite{hsic:novello} obtained by post-processing the random search for Mutation++ for the toy atmosphere. }
  \label{hsic}
\end{figure}

Figure \ref{hsic} presents the aforementioned goal-oriented sensitivity indices for each hyperparameter. The indices are stacked and sorted by decreasing importance from top to bottom. For instance, the choice of the $\texttt{optimizer}$ is the most important hyperparameter to reach the best $10\%$ error percentile, and $\texttt{dropout\_rate}$ is the least important. Besides, estimation error bars for these indices are provided, attesting to relatively converged results with respect to the number of points in the random search $N_{MC}=3\times 10^3$. 
Note that the error bars of the last hyperparameters intersect, so we will not allow ourselves to interpret them. Nonetheless, according to the error bars, we can focus on the primary hyperparameters, at least the $5$ most important. 
In other words, hyperparameters \texttt{optimizer}, \texttt{learning\_rate} and
\texttt{activation} are by far the three most influential hyperparameters in order to reach the $10\%$ best errors. This information is of great value: these hyperparameters do not
impact the run-time once the parameters $\theta$ are tuned. This means that within the $10\%$ best results, there are probably cheap (i.e. shallow and tight) neural networks.
The number of layers $\nlayers$ and of units $\nunits$ only come at the $4^{th}$ and $5^{th}$ position of relative importance. By using the methodologies of \cite{hsic:novello}, it is possible to:
\begin{itemize}
  \item Select cost effective values for $\nlayers$ and $\nunits$, as well as other low-impactful hyperparameters, with a limited impact on the error,
  \item Focus subsequent hyperparameter optimization on the three most impactful hyperparameters, $\texttt{optimizer}$, $\texttt{learning\_rate}$ and $\texttt{activations}$.
\end{itemize}

We perform a Gaussian Process-based bayesian optimization in low dimension -  see the TS-GPBO methodology of \cite{hsic:novello}. The obtained neural network reaches an $L_2$ error of $8.48 \times 10^{-8}$, which is even lower than with the previous random search, 
with only $\nlayers = 5$ layers and $\nunits = 20$ neurons. 
This network has a competitive error with far fewer parameters and much shorter run times. We suggest now plugging this neural network into the reentry code and revisiting the problem of section \ref{intro} with a hybrid simulation code. 

\section{Deep Learning-based hybridization with guarantees}
\label{numres}


In this section, we revisit the motivating example of section \ref{intro} with a hybrid reentry simulation code. The sketch of the code is recalled below in the algorithm
\ref{reentry_code_nn}.\\ \ \\ 

    \begin{algorithm}[H]
    \caption{Core of the code with a call to a neural network surrogate model of Mutation++.}
        \label{reentry_code_nn}

      \textit{\# general initialization (mesh, quantities on mesh etc.)}

      initialise\_guess\_vector\_of\_unknowns\_on\_mesh($U^0$,$\x^0$)

      \While {\text{convergence\_criterion\_not\_satisfied}}{

      $U^{n+\frac12}=$solve\_Euler\_equations($U^n$,$\x^{n}$)

      $\x^{n+1},  U^{n+1}=$call\_neural\_network($U^{n+\frac12}$,$\x^{n}$)

      $U^n \leftarrow U^{n+1}$

      $\x^n \leftarrow \x^{n+1}$

                 }
    \end{algorithm} \ \\

In function call\_neural\_network, we plug the neural network approximating Mutation++ described at the end of section \ref{ml} using the Tensorflow C API and a wrapper,
CppFlow\footnote{\url{https://github.com/serizba/cppflow}}. Note that Algorithm \ref{reentry_code_nn} is slightly different from Algorithm \ref{reentry_code} because, in the former, the neural network is called in a batch fashion to take advantage of the vectorial optimizations, while in the latter, Mutation++ has to be called sequentially.
We qualitatively verify that the results obtained thanks to the neural network are in agreement with the ones of Mutation++ by displaying the maps $(\rho,\epsilon)\rightarrow \alpha(\rho,
\epsilon)$ for $\alpha\in\{\x_O, \x_N,\x_{NO}, P, T, C_p, C_v, c\}$ in Appendix \textbf{\hyperref[appA]{Appendix A}}.

\begin{figure}[!h]
  \raggedleft
  \begin{subfigure}{0.03\linewidth}
    \rotatebox{90}{$\mut$ \hspace{1.8cm} net.} 
  \end{subfigure}
  \begin{subfigure}{0.03\linewidth}
    \rotatebox{90}{$\epsilon$ \hspace{2.3cm} $\epsilon$} 
  \end{subfigure}
  \hspace{-0.35cm}
  \begin{subfigure}{0.22\linewidth}
    \begin{center}
      \includegraphics[trim={0.6cm 0.6cm 0 0}, clip,width=1\linewidth]{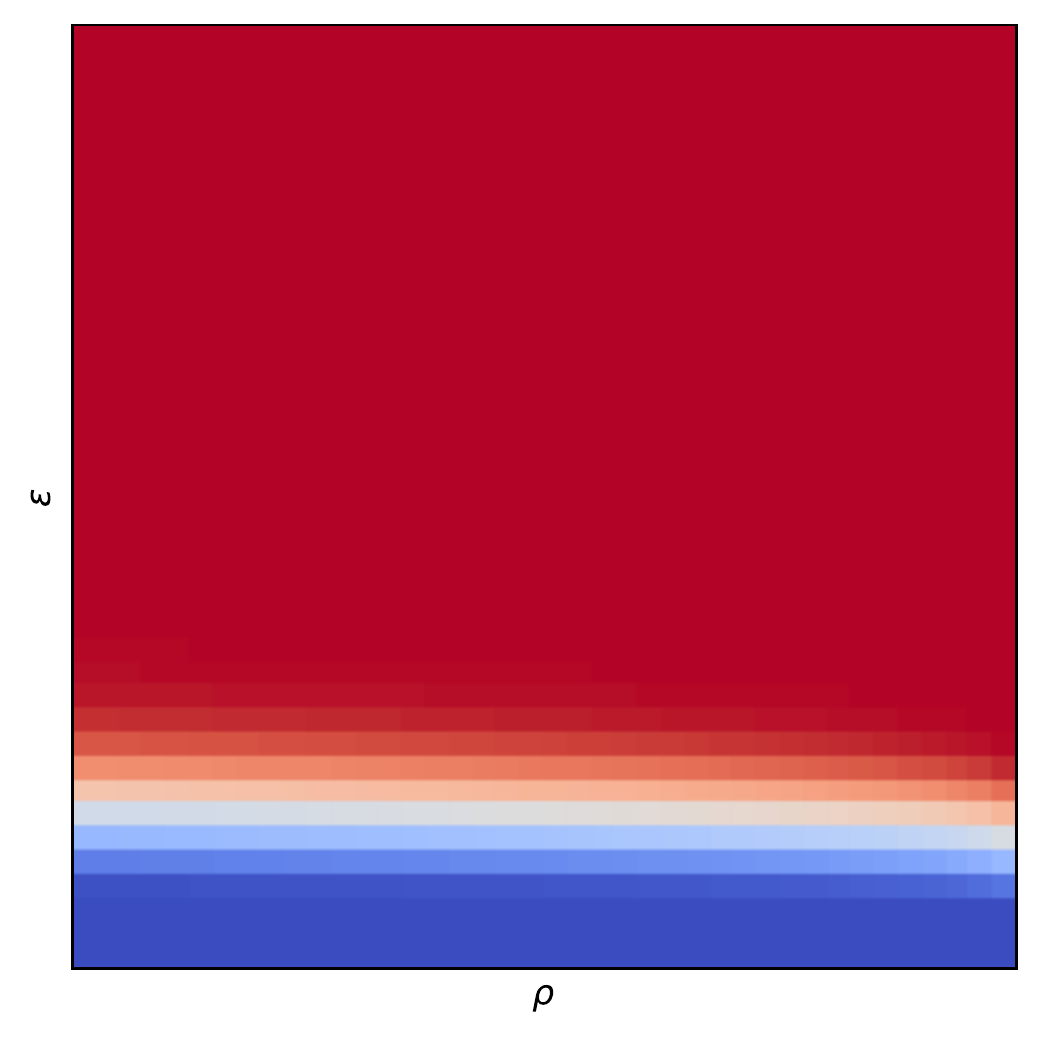}
      \includegraphics[trim={0.6cm 0.6cm 0 0}, clip,width=1\linewidth]{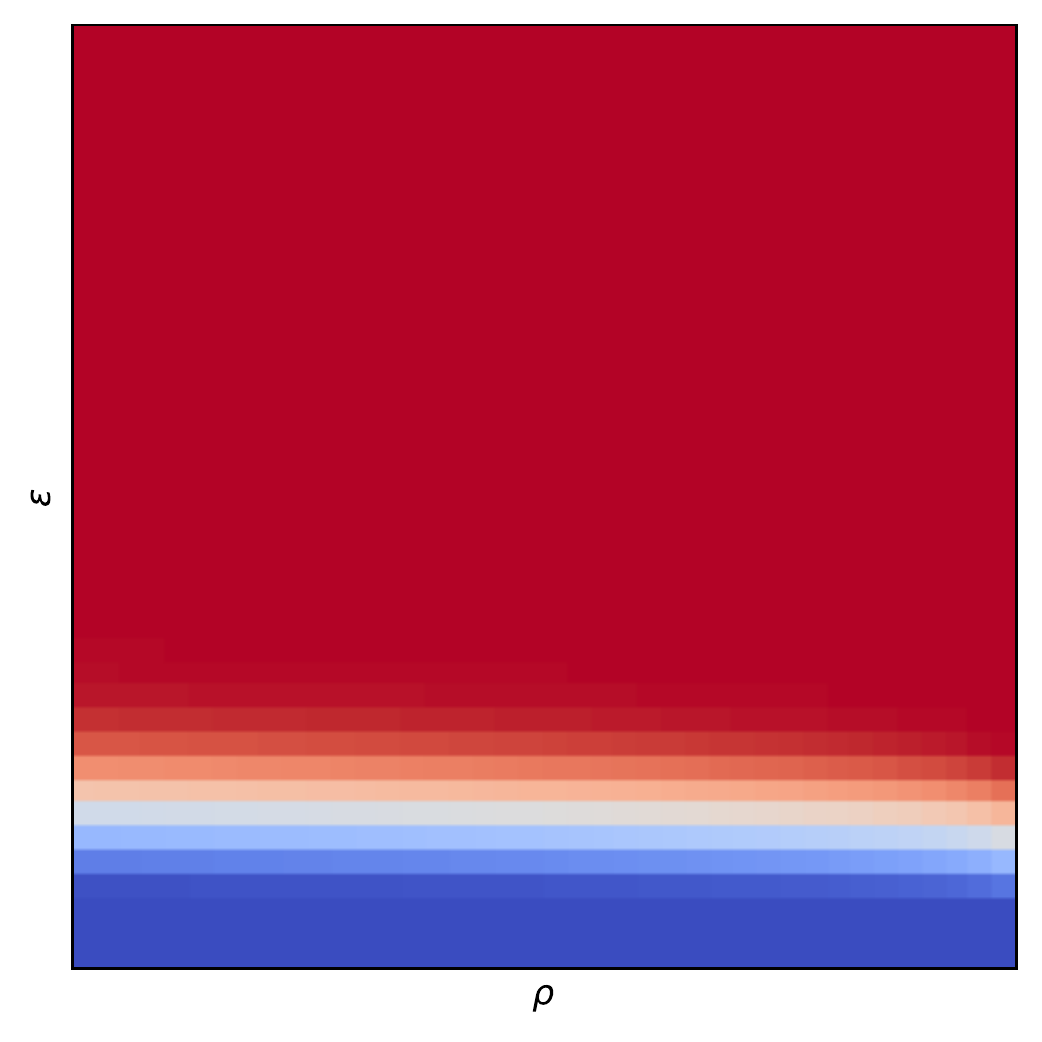}
    \end{center}
  \end{subfigure}
  \begin{subfigure}{0.22\linewidth}
    \begin{center}
      \includegraphics[trim={0.6cm 0.6cm 0 0}, clip,width=1\linewidth]{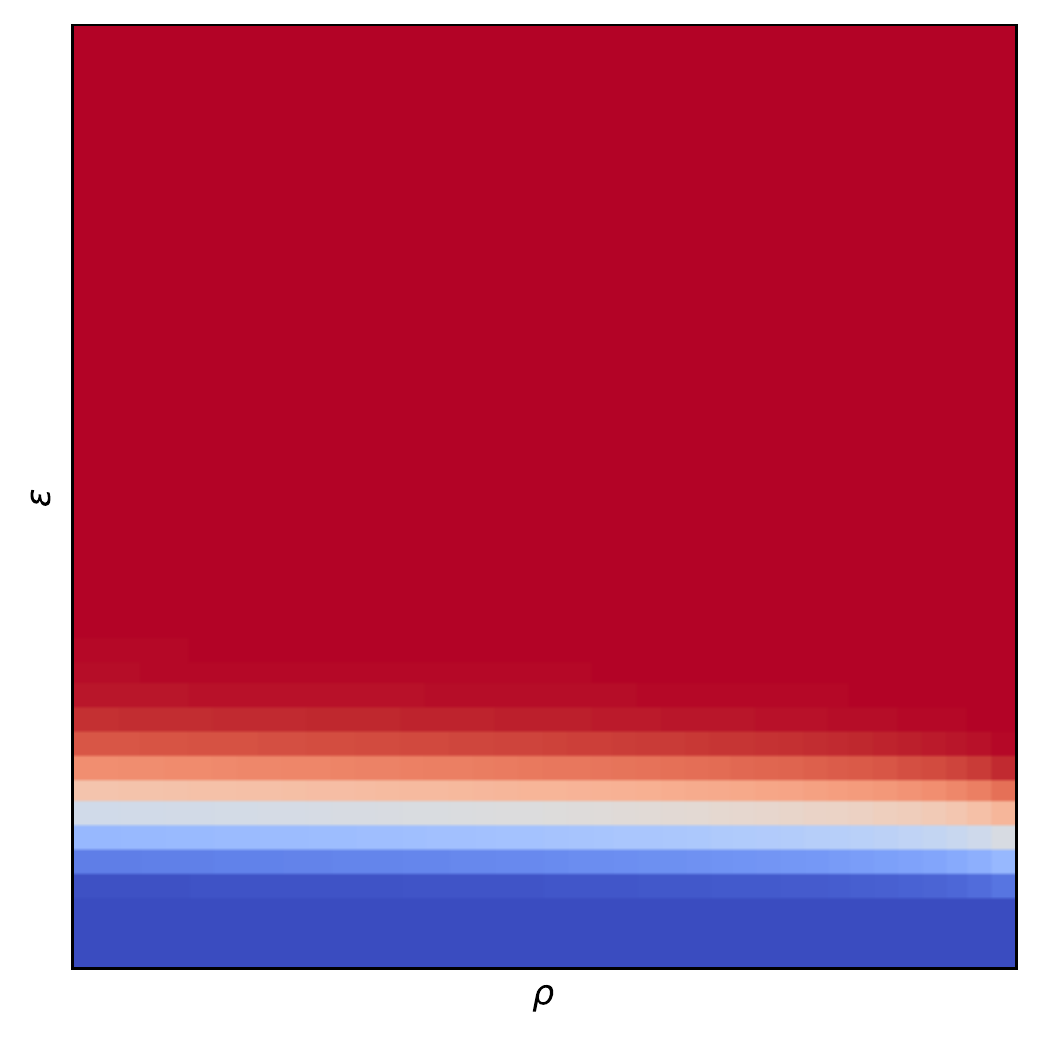}
      \includegraphics[trim={0.6cm 0.6cm 0 0}, clip,width=1\linewidth]{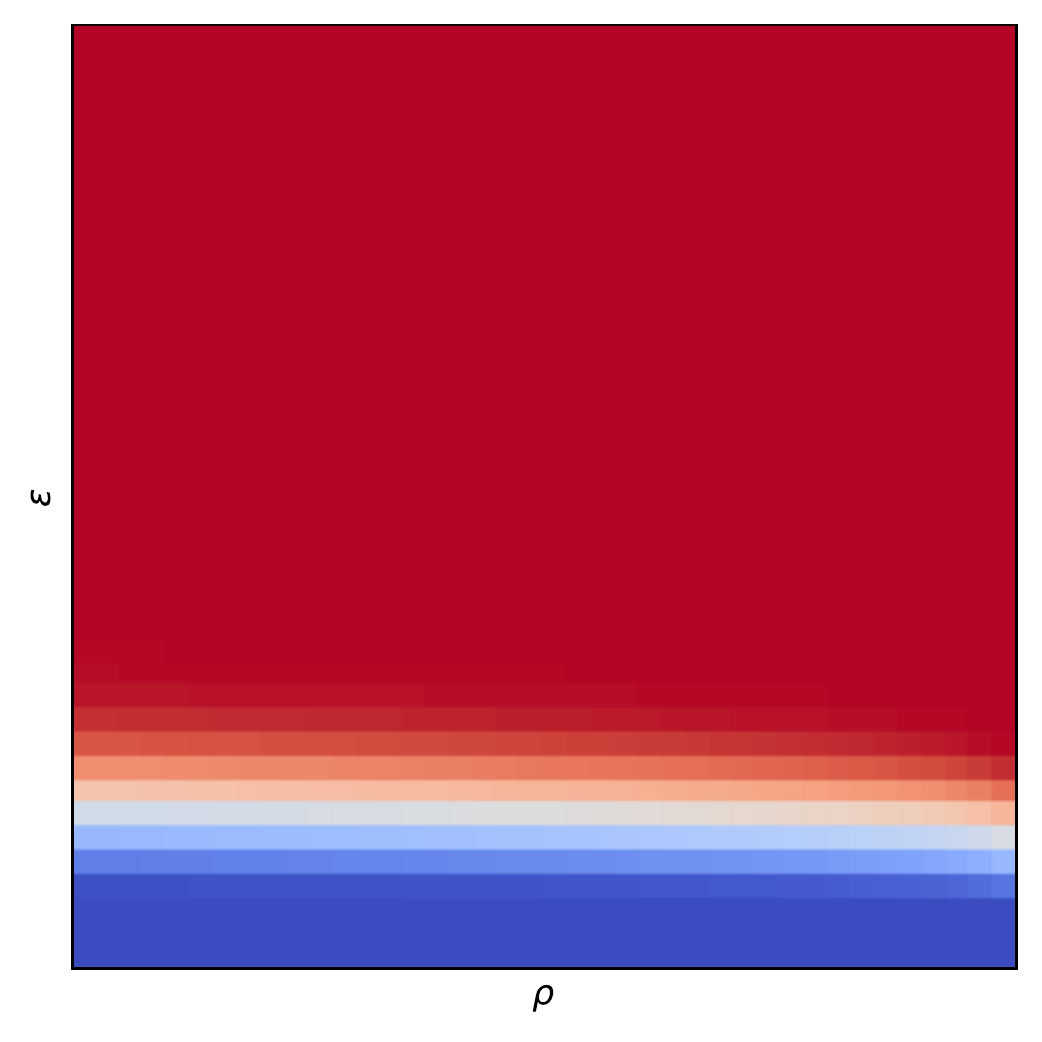}
    \end{center}
  \end{subfigure}
  \begin{subfigure}{0.22\linewidth}
    \begin{center}
      \includegraphics[trim={0.6cm 0.6cm 0 0}, clip,width=1\linewidth]{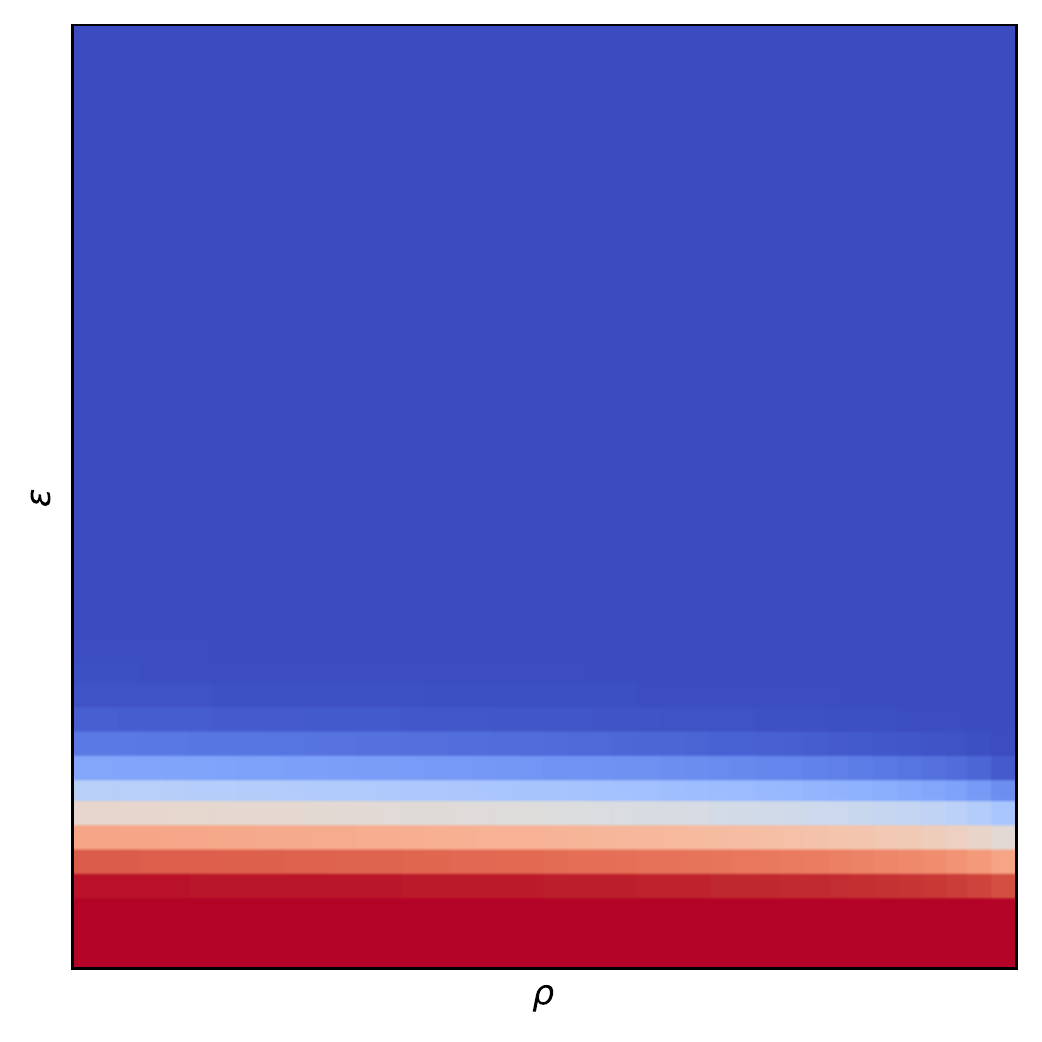}
      \includegraphics[trim={0.6cm 0.6cm 0 0}, clip,width=1\linewidth]{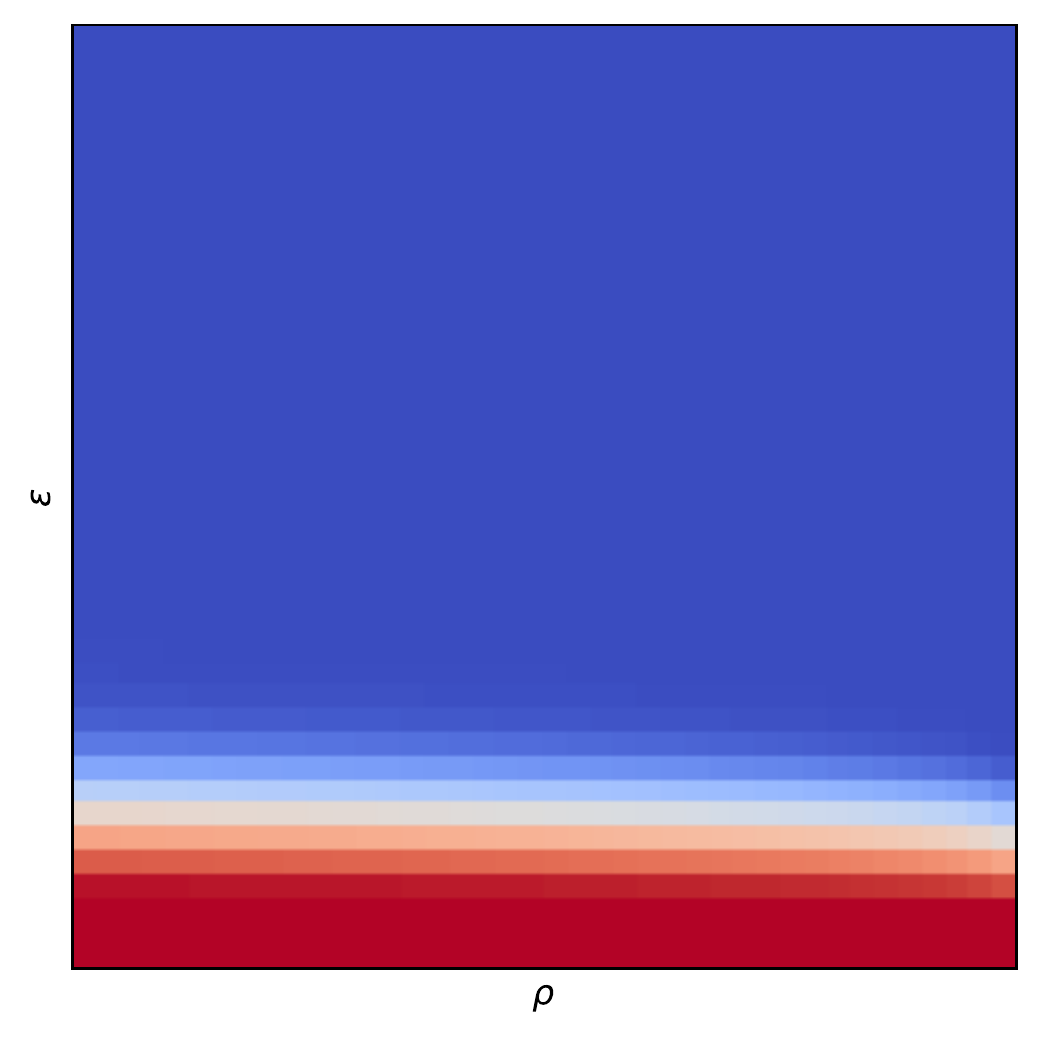}
    \end{center}
  \end{subfigure}
  \begin{subfigure}{0.22\linewidth}
    \begin{center}
      \includegraphics[trim={0.6cm 0.6cm 0 0}, clip,width=1\linewidth]{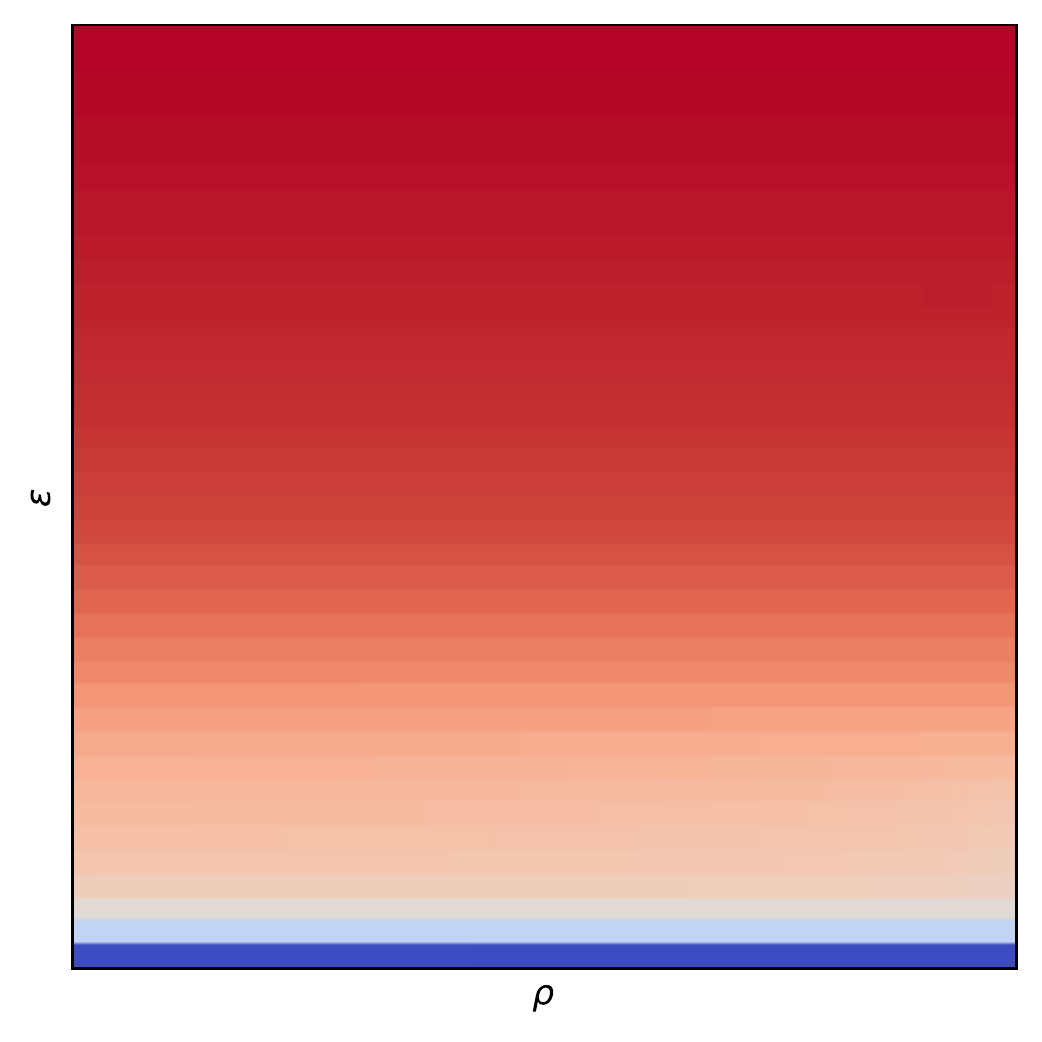}
      \includegraphics[trim={0.6cm 0.6cm 0 0}, clip,width=1\linewidth]{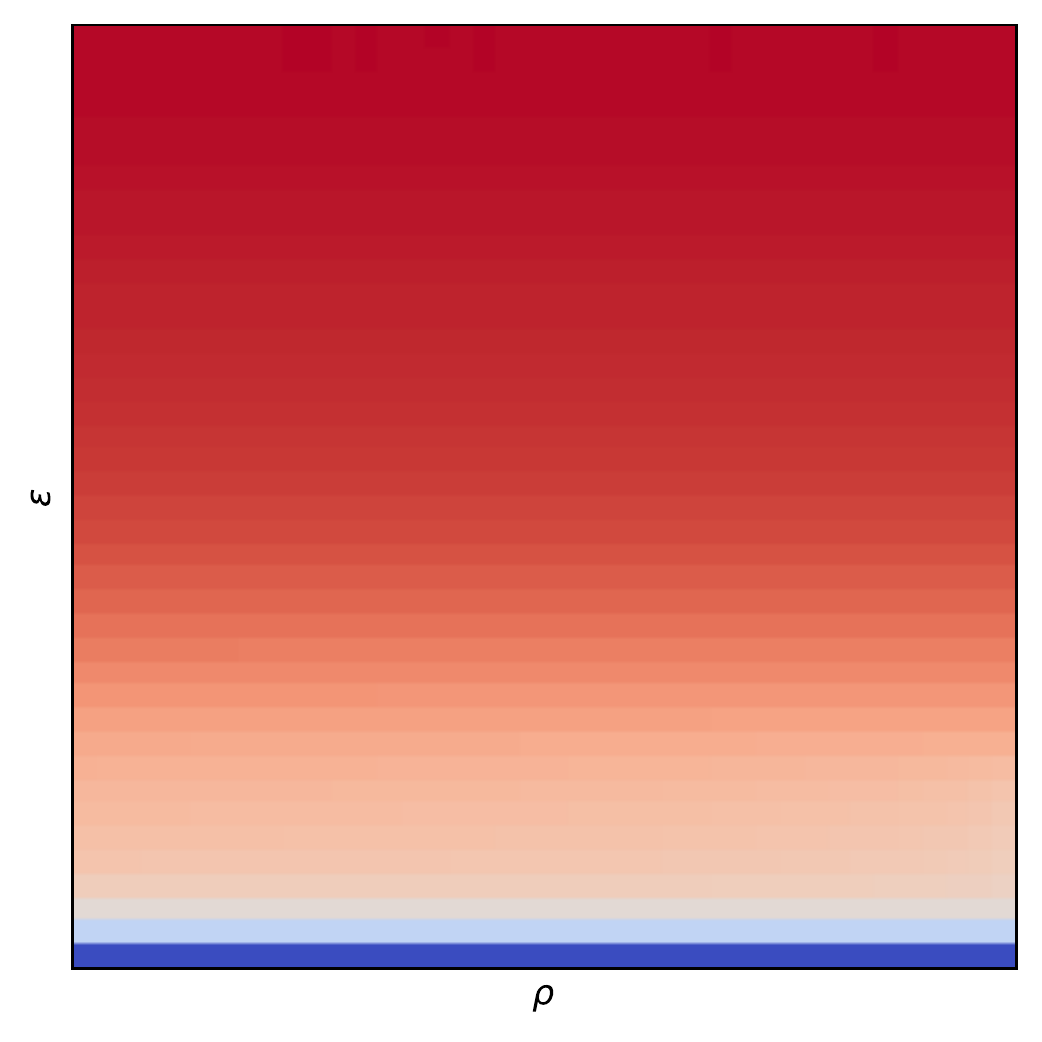}
    \end{center}
  \end{subfigure}\\
  \raggedleft
  \begin{subfigure}{0.08\linewidth}
  \end{subfigure}
  \begin{subfigure}{0.22\linewidth}
    \begin{center}
    $\rho$
    \end{center}
  \vspace{-0.2cm}
  \caption{$\x_O$}
  \vspace{0.7cm}
  \end{subfigure}
  \begin{subfigure}{0.22\linewidth}
    \begin{center}
      $\rho$
    \end{center}
    \vspace{-0.2cm}
    \caption{$\x_N$}
    \vspace{0.7cm}
  \end{subfigure}
  \begin{subfigure}{0.22\linewidth}
    \begin{center}
      $\rho$
    \end{center}
    \vspace{-0.2cm}
    \caption{$\x_{NO}$}
  \vspace{0.7cm}
  \end{subfigure}
  \begin{subfigure}{0.22\linewidth}
    \begin{center}
      $\rho$
    \end{center}
    \vspace{-0.2cm}
    \caption{$p$}
    \vspace{0.7cm}
  \end{subfigure}\\

  \raggedleft
  \begin{subfigure}{0.03\linewidth}
    \rotatebox{90}{$\mut$ \hspace{1.8cm} net.} 
  \end{subfigure}
  \begin{subfigure}{0.03\linewidth}
    \rotatebox{90}{$\epsilon$ \hspace{2.3cm} $\epsilon$} 
  \end{subfigure}
  \hspace{-0.35cm}
  \begin{subfigure}{0.22\linewidth}
    \begin{center}
      \includegraphics[trim={0.6cm 0.6cm 0 0}, clip,width=1\linewidth]{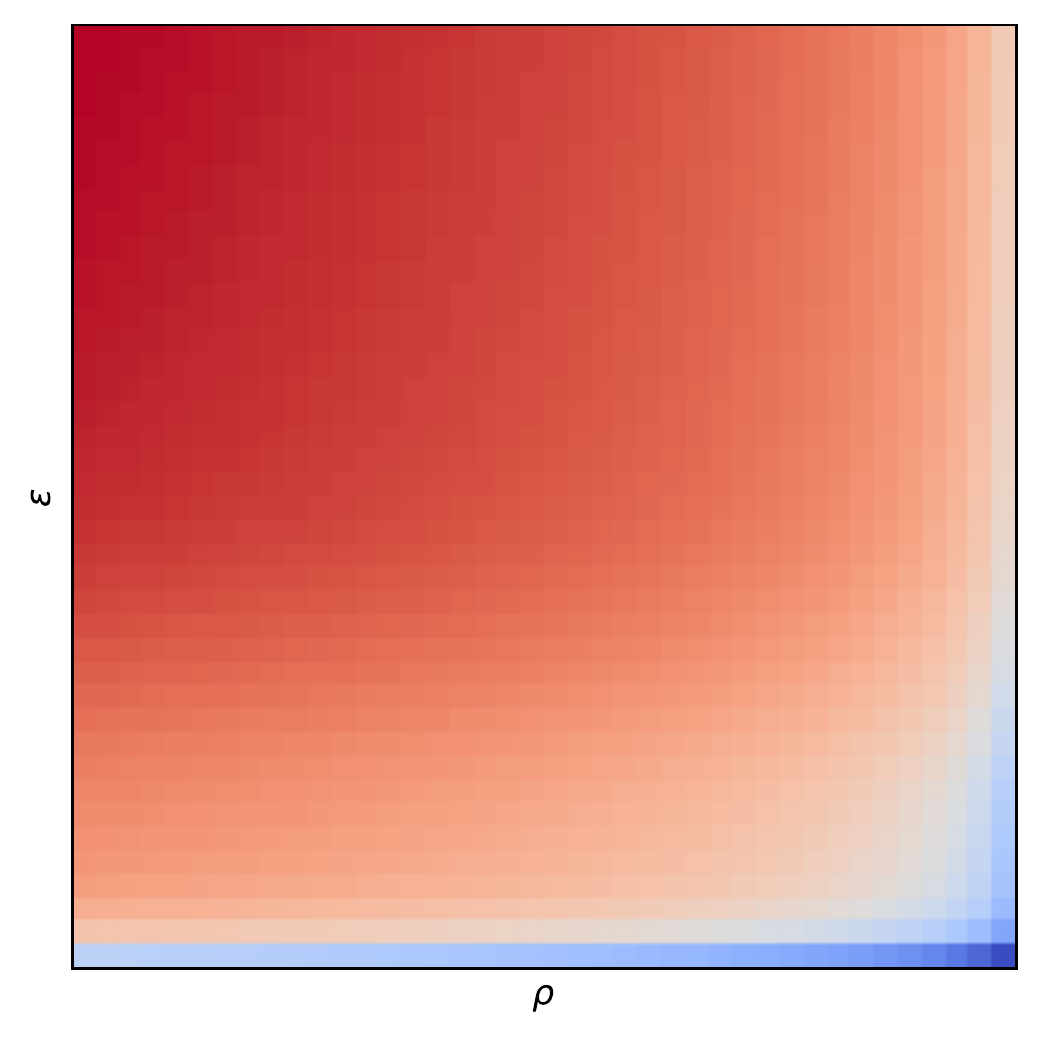}
      \includegraphics[trim={0.6cm 0.6cm 0 0}, clip,width=1\linewidth]{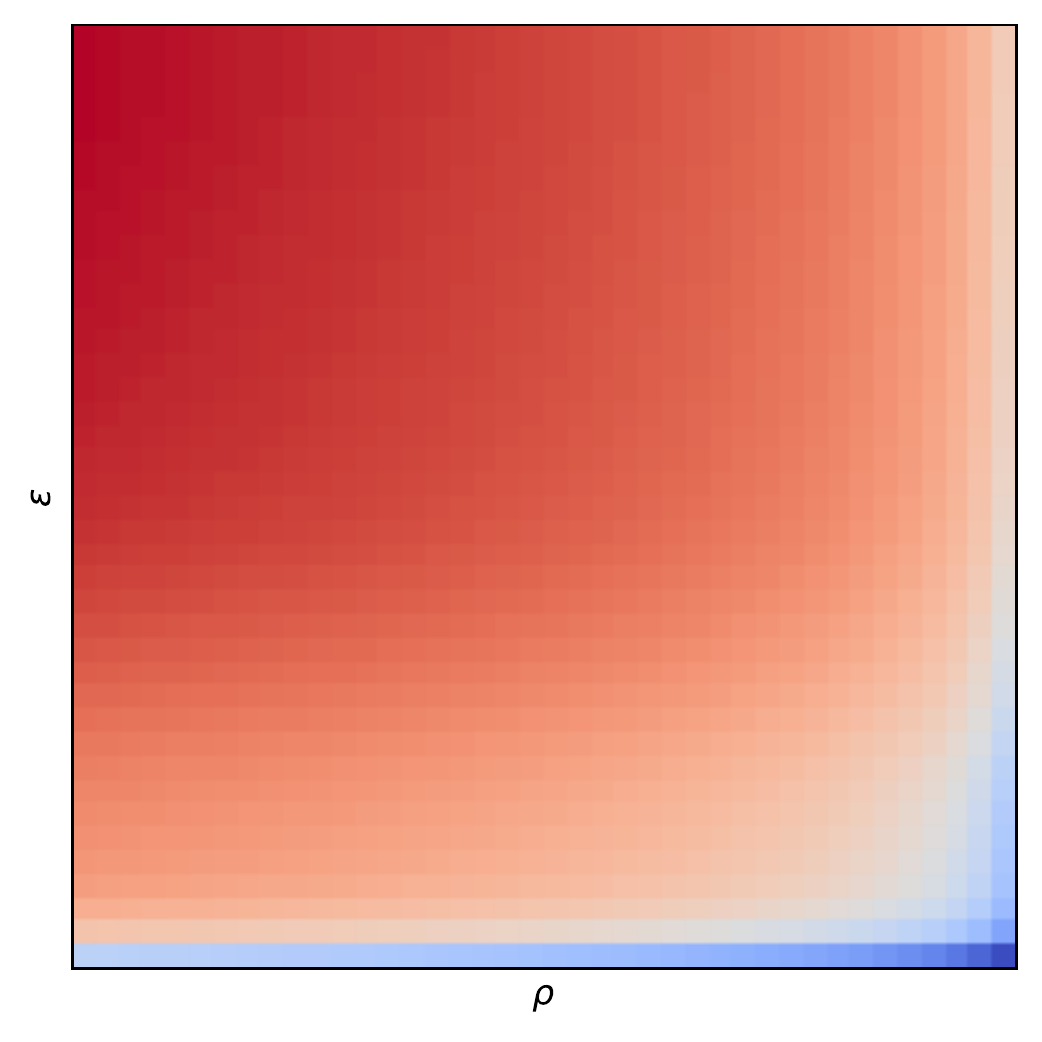}
    \end{center}
  \end{subfigure}
  \begin{subfigure}{0.22\linewidth}
    \begin{center}
      \includegraphics[trim={0.6cm 0.6cm 0 0}, clip,width=1\linewidth]{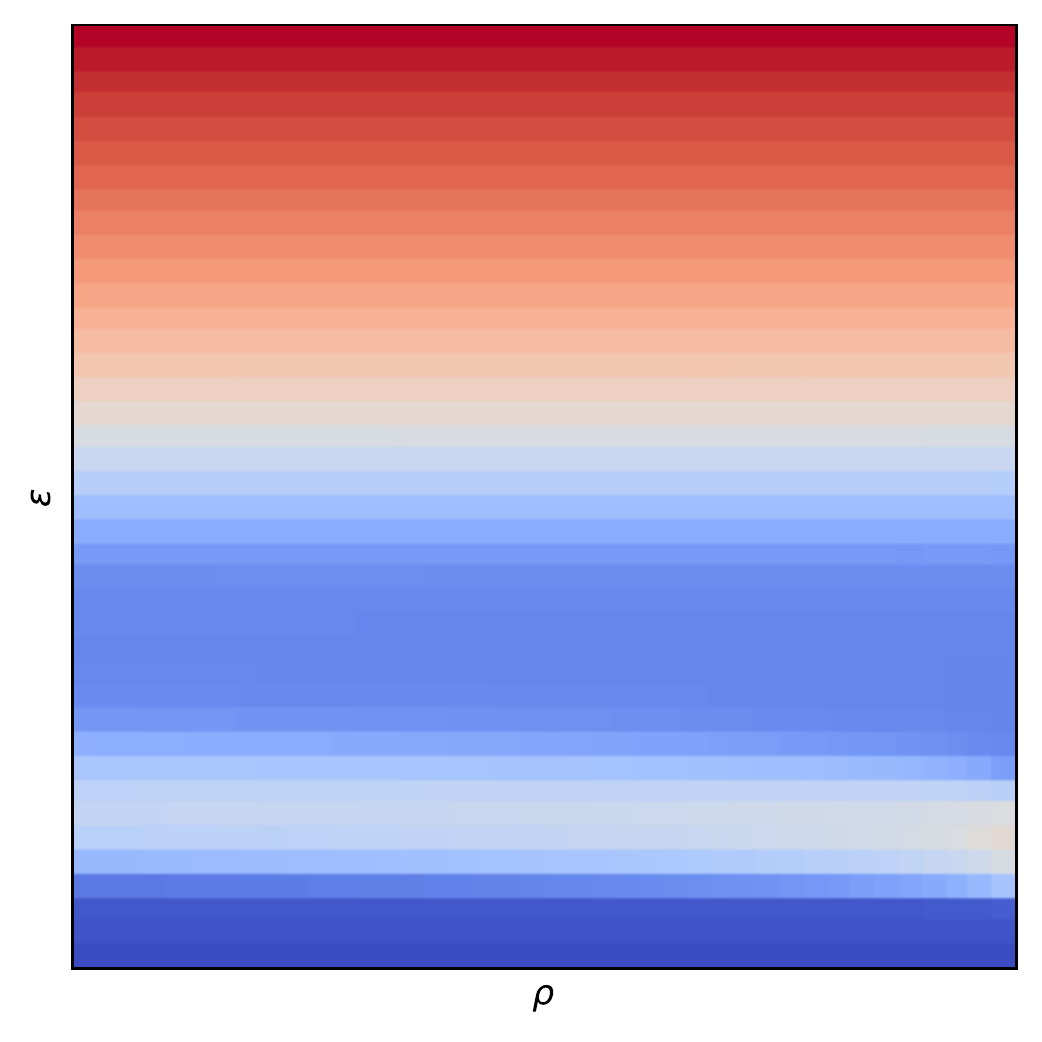}
      \includegraphics[trim={0.6cm 0.6cm 0 0}, clip,width=1\linewidth]{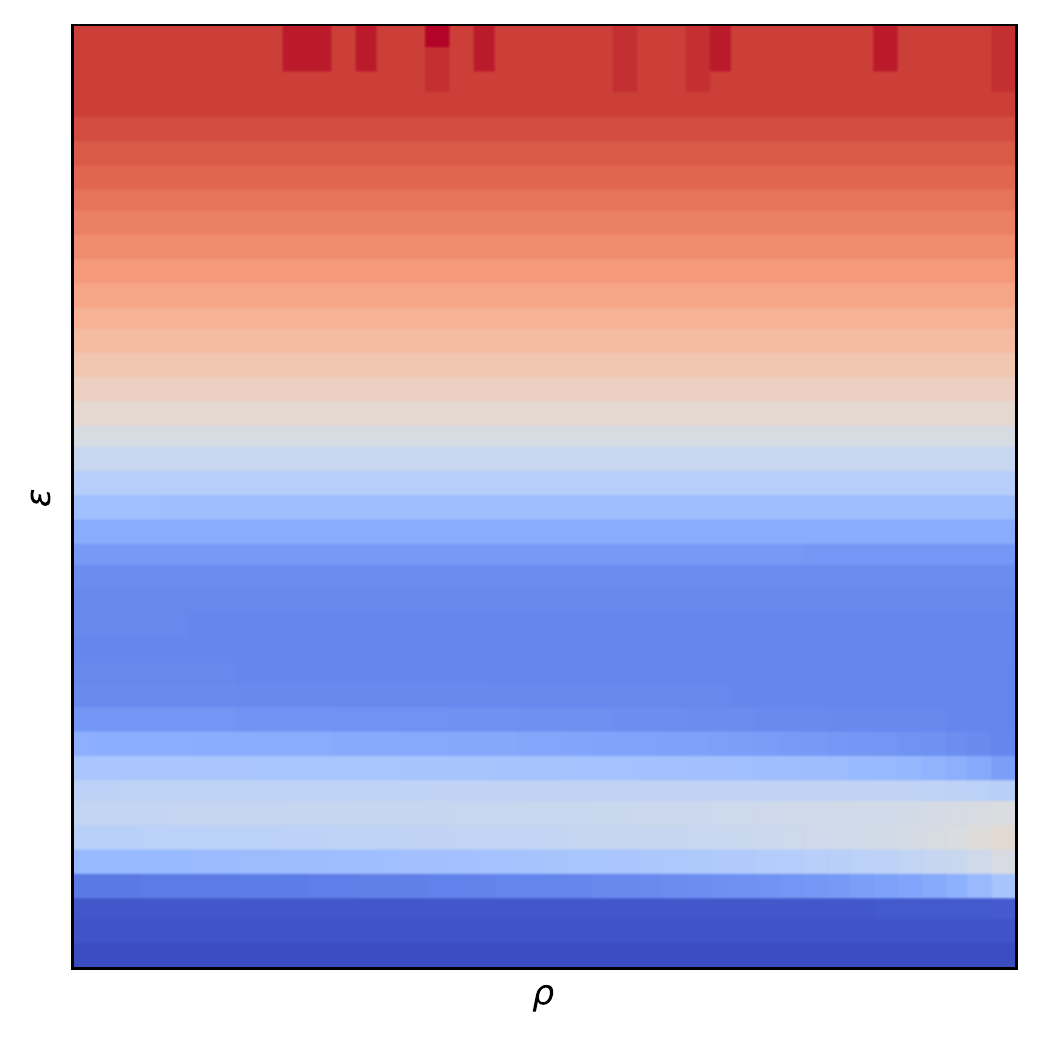}
    \end{center}
  \end{subfigure}
  \begin{subfigure}{0.22\linewidth}
    \begin{center}
      \includegraphics[trim={0.6cm 0.6cm 0 0}, clip,width=1\linewidth]{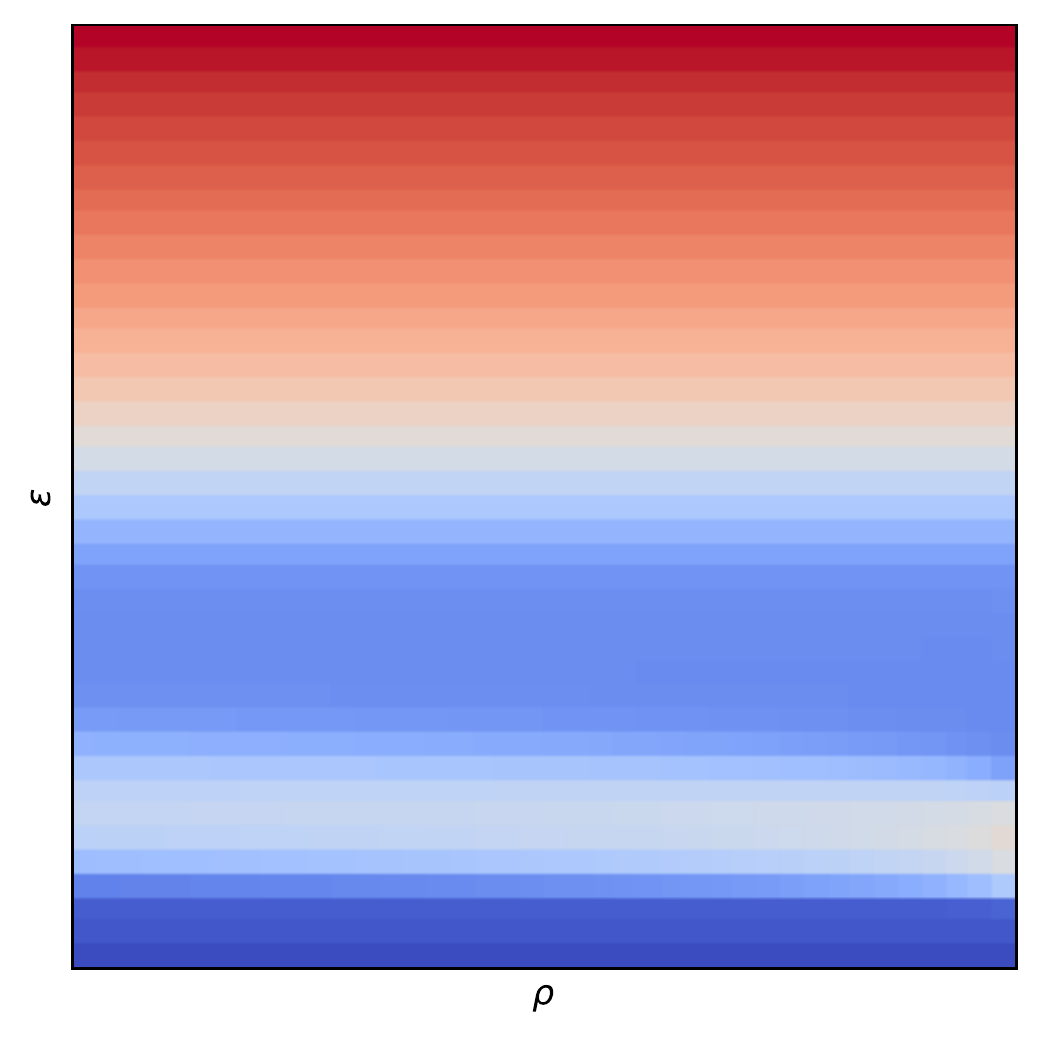}
      \includegraphics[trim={0.6cm 0.6cm 0 0}, clip,width=1\linewidth]{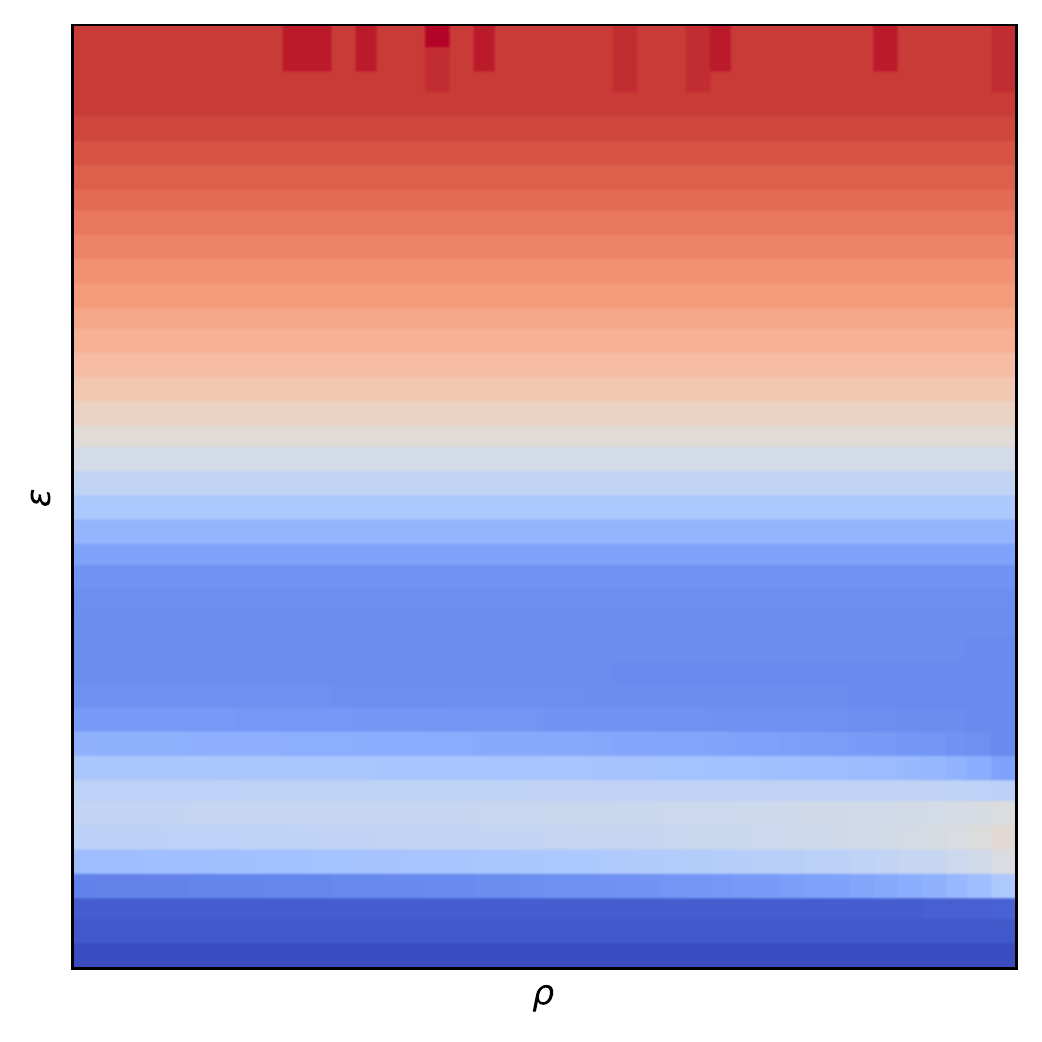}
    \end{center}
  \end{subfigure}
  \begin{subfigure}{0.22\linewidth}
    \begin{center}
      \includegraphics[trim={0.6cm 0.6cm 0 0}, clip,width=1\linewidth]{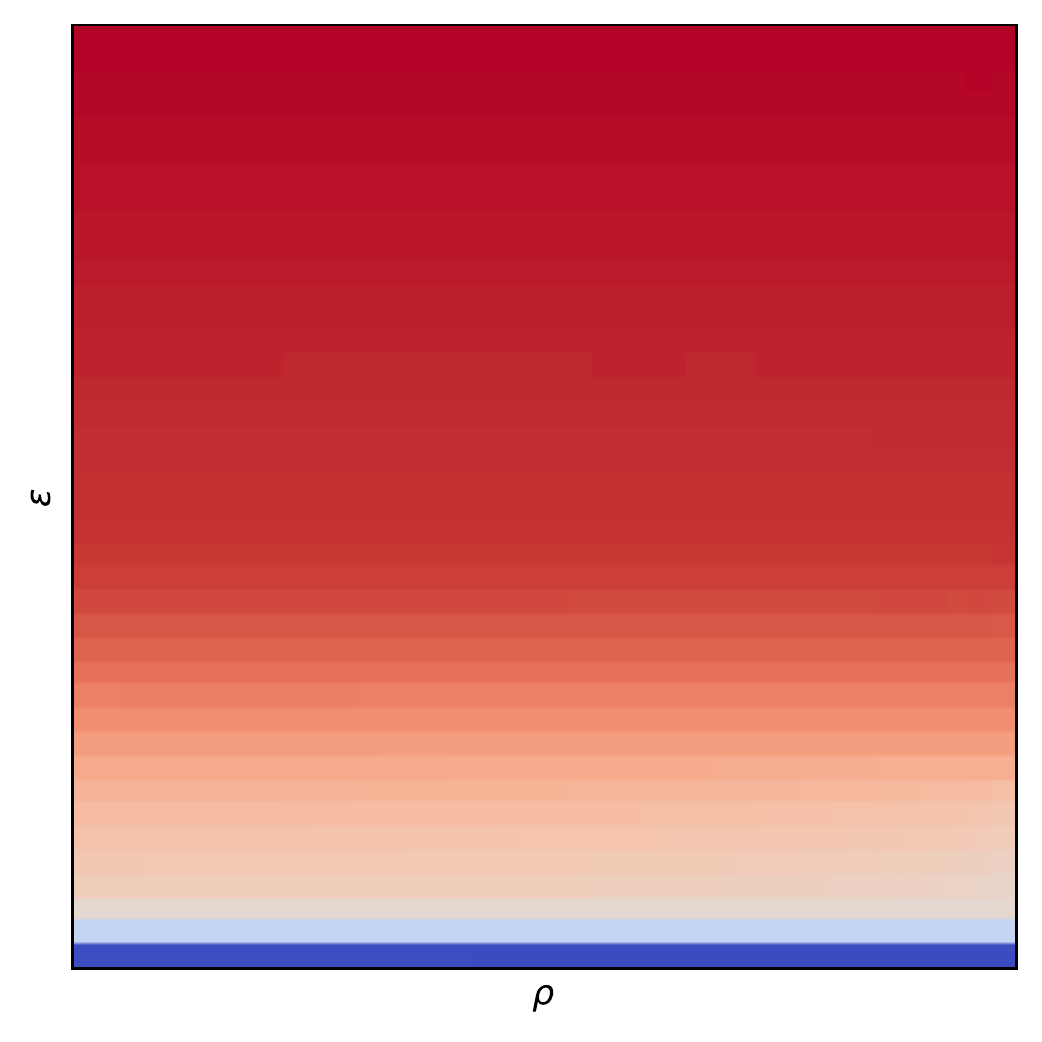}
      \includegraphics[trim={0.6cm 0.6cm 0 0}, clip,width=1\linewidth]{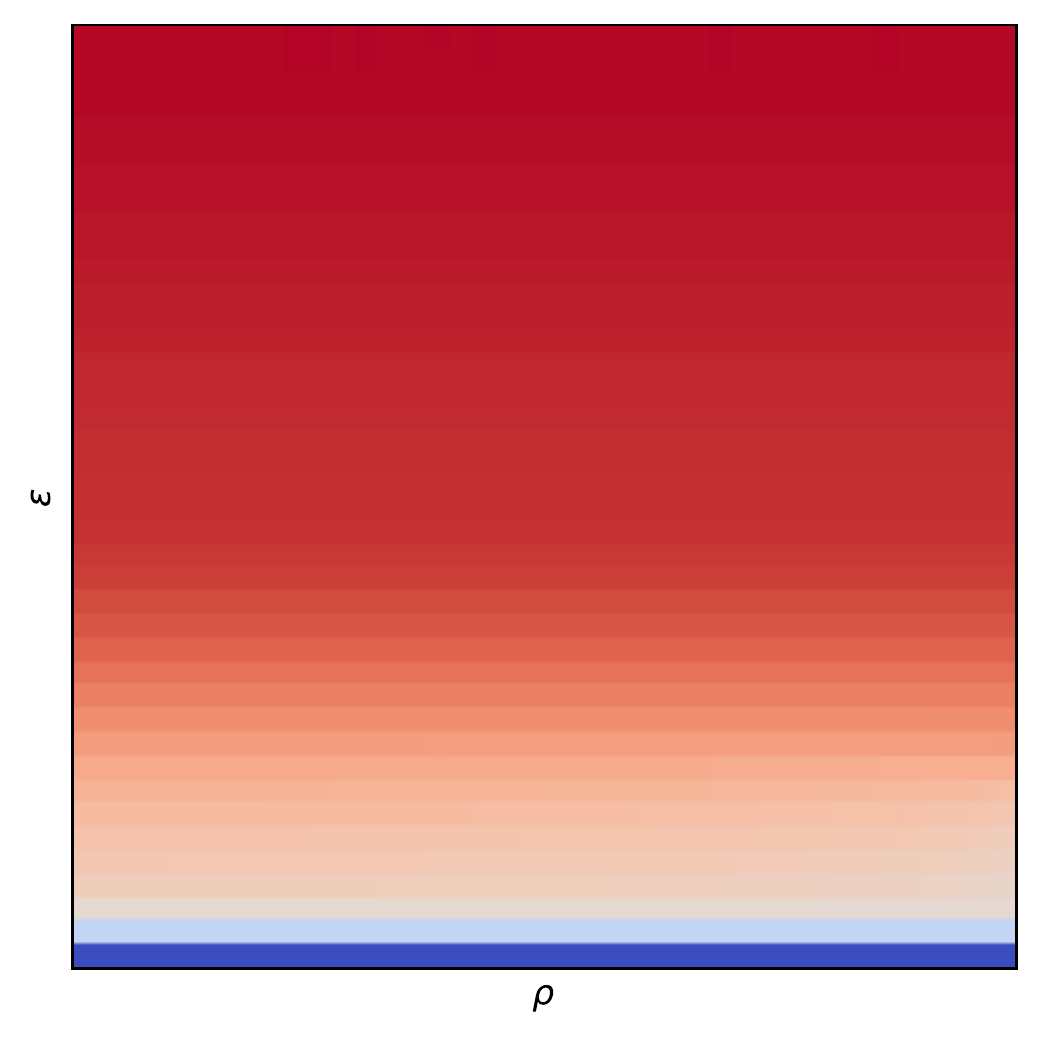}
    \end{center}
  \end{subfigure}\\
  \raggedleft
  \begin{subfigure}{0.08\linewidth}
  \end{subfigure}
  \begin{subfigure}{0.22\linewidth}
    \begin{center}
    $\rho$
    \end{center}
  \vspace{-0.2cm}
  \caption{$T$}
  \end{subfigure}
  \begin{subfigure}{0.22\linewidth}
    \begin{center}
      $\rho$
    \end{center}
    \vspace{-0.2cm}
    \caption{$C_p$}
  \end{subfigure}
  \begin{subfigure}{0.22\linewidth}
    \begin{center}
      $\rho$
    \end{center}
    \vspace{-0.2cm}
    \caption{$C_v$}
  \end{subfigure}
  \begin{subfigure}{0.22\linewidth}
    \begin{center}
      $\rho$
    \end{center}
    \vspace{-0.2cm}
    \caption{$c$}
  \end{subfigure}\\
   \caption{Predictions of the neural network (top) and predictions of $\mut$ (bottom) on the input domain. Axis values are omitted for clarity but we recall that $\rho$ and $\epsilon$ are defined on $[0.1, 3.8] and [2.07503 \times 10^7, 3 \times 10^8]$ respectively.}\label{fig:nnpred}
\end{figure}

For each output physical observable of figure \ref{fig:nnpred}, the predictions are compared to the reference values computed by $\mut$. The results are in very good agreement (remember we have a $L_2$ error
close to $10^{-8}$ with this neural network). 
Still, in figure \ref{fig:field} (f) and (g), for $C_p$ and $C_v$, there are numerical artifacts in the predictions of $\mut$ at the top of the domain (for high values of
$\epsilon$) whereas these are not observable with the neural network. 
These numerical instabilities $\mut$ may disturb the reentry computation in practice. The neural network does not seem to be subject to such instabilities. We will see that this point is of
importance in the following numerical results.\\

From now on, this section is articulated as follows. In section \ref{sec:toy_example_mpp}, we present the results and accelerations obtained with the hybrid reentry code. However, in critical decision-making, the use of machine learning is often controversial because it lacks natural accuracy guarantees. Sections \ref{sec:zeroerr} and \ref{non0error} are dedicated to alleviating this problem. Notably, in section
\ref{sec:zeroerr}, we explain how to make sure that the hybrid code recovers {\em exactly} the same results as the native one (together with ensuring a
$\times 10$ factor of acceleration). Finally, in section \ref{non0error}, we suggest a way to validate the reliability of the hybrid simulation code as such based on 
proper analysis and study of the different sources of errors and uncertainties of the original simulation code.  
%

\subsection{Acceleration of the reentry code}
\label{sec:toy_example_mpp}

\newcolumntype{P}[1]{>{\centering\arraybackslash}p{#1}}
In this section, we come back to the configuration of section \ref{intro}.
The reference simulations are provided by the native reentry simulation code described 
in section \ref{cfd} (and algorithm \ref{reentry_code}).

In the configuration of interest, see table \ref{tab:toy_case}, we consider a sphere of radius $r_\text{sphere}=10^{-2}$
entering a simplified Earth's atmosphere (species $N,O,NO$). 
\begin{table}[!h]
    \centering
    \begin{tabular}{p{6cm}P{4cm}}
      Input  &  value\\
        \hline
      Elements, $n_e=2$ (\texttt{elem:fraction}) & O:0.2, N:0.8 \\
      Upstream pressure & $35737.40 Pa$\\
      Upstream temperature & $216.57 K$\\
      Upstream velocity (Mach $16$) & $4930.83 \; m.s^{-1}$\\
      Chemical species ($n_s=3$) & N, O, NO \\
    \end{tabular}
    \caption{\label{tab:toy_case} Simulation parameters and boundary conditions for the toy example.}
  \end{table}
The boundary conditions are:
\begin{itemize}
  \item no-slip boundary conditions on the sphere incoming in Earth's atmosphere,
  \item incoming flux upstream of the sphere, see table \ref{tab:toy_case} for the considered nominal values of the upstream velocity, pressure, and temperature. 
\end{itemize}
%
%
%
Let us first perform some comparisons on the pressure fields in the same conditions (same mesh) as in section \ref{intro} with the three different reentry codes:
\begin{itemize}
  \item PG (for perfect gas) denotes the results obtained with the reentry code without simulating any chemical reactions.
  \item MPP (for Mutation++) denotes the results obtained with the reentry code with the simulation of chemical reactions using Mutation++.
  \item NN (for neural network) denotes the results obtained with the hybrid reentry code with the simulation of chemical reactions using the neural network approximating Mutation++
    obtained with the methodology described in section \ref{ml}.
\end{itemize}
\begin{figure}[!h]
  \centering
  \begin{subfigure}{0.31\textwidth}
    \begin{center}
      \includegraphics[width=1\textwidth]{figures_mutation/field_MPP_gpscale.pdf}
    \end{center}
  \caption{MPP, run-time 4090 s.}
  \end{subfigure}
  \begin{subfigure}{0.31\textwidth}
    \begin{center}
      \includegraphics[width=1\textwidth]{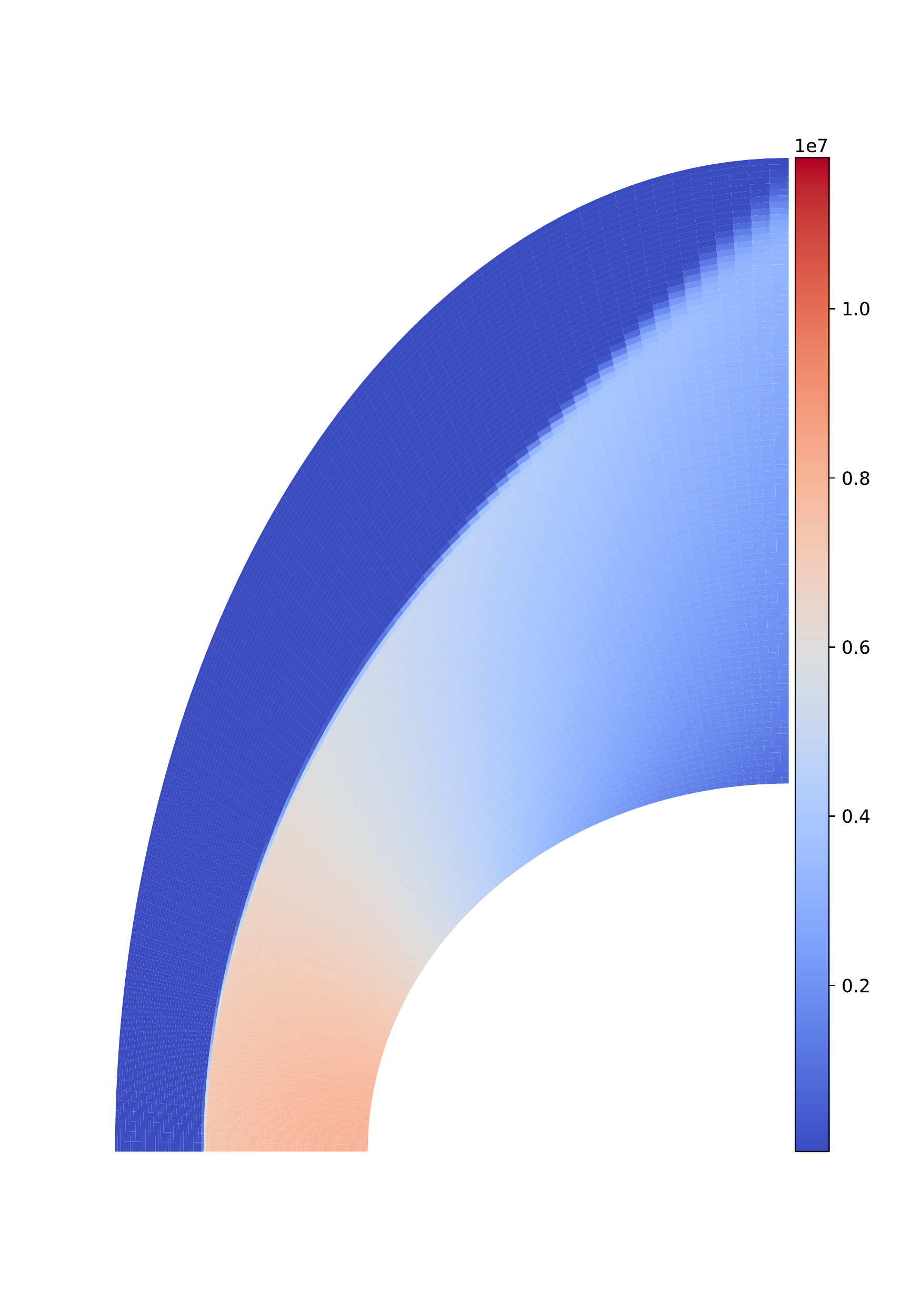}
    \end{center}
    \caption{NN, run-time 220 s.}
  \end{subfigure}
  \begin{subfigure}{0.31\textwidth}
    \begin{center}
      \includegraphics[width=1\textwidth]{figures_mutation/field_GP_gpscale.pdf}
    \end{center}
    \caption{PG, run-time 81 s.}
  \end{subfigure}
   \caption{Pressure field for MPP, NN and PG, with the same scales.}\label{fig:field}
\end{figure}
Figure \ref{fig:field} echoes figure \ref{fig:field1} and presents the results obtained with the three above reentry codes. 
As in section \ref{intro}, we can observe, by comparing figure \ref{fig:field} (a) and (c) that the perfect gas closure (PG) is a coarse model for our reentry problem as its results in terms of pressure field considerably differ
from the one obtained with chemical reactions (MPP), i.e. which takes into account finer physics. On the other hand, the pressure field obtained with the hybrid reentry code (NN) is
not visually distinguishable from the results obtained with Mutation++. Now, in terms of run-time, with about the same accuracy, the hybrid reentry code ensures a gain of a
factor $\times \frac{4090}{220}\approx 18.6$.

\begin{figure}[!h]
  \centering
  \begin{subfigure}{0.49\textwidth}
    \begin{center}
      \includegraphics[width=1\textwidth]{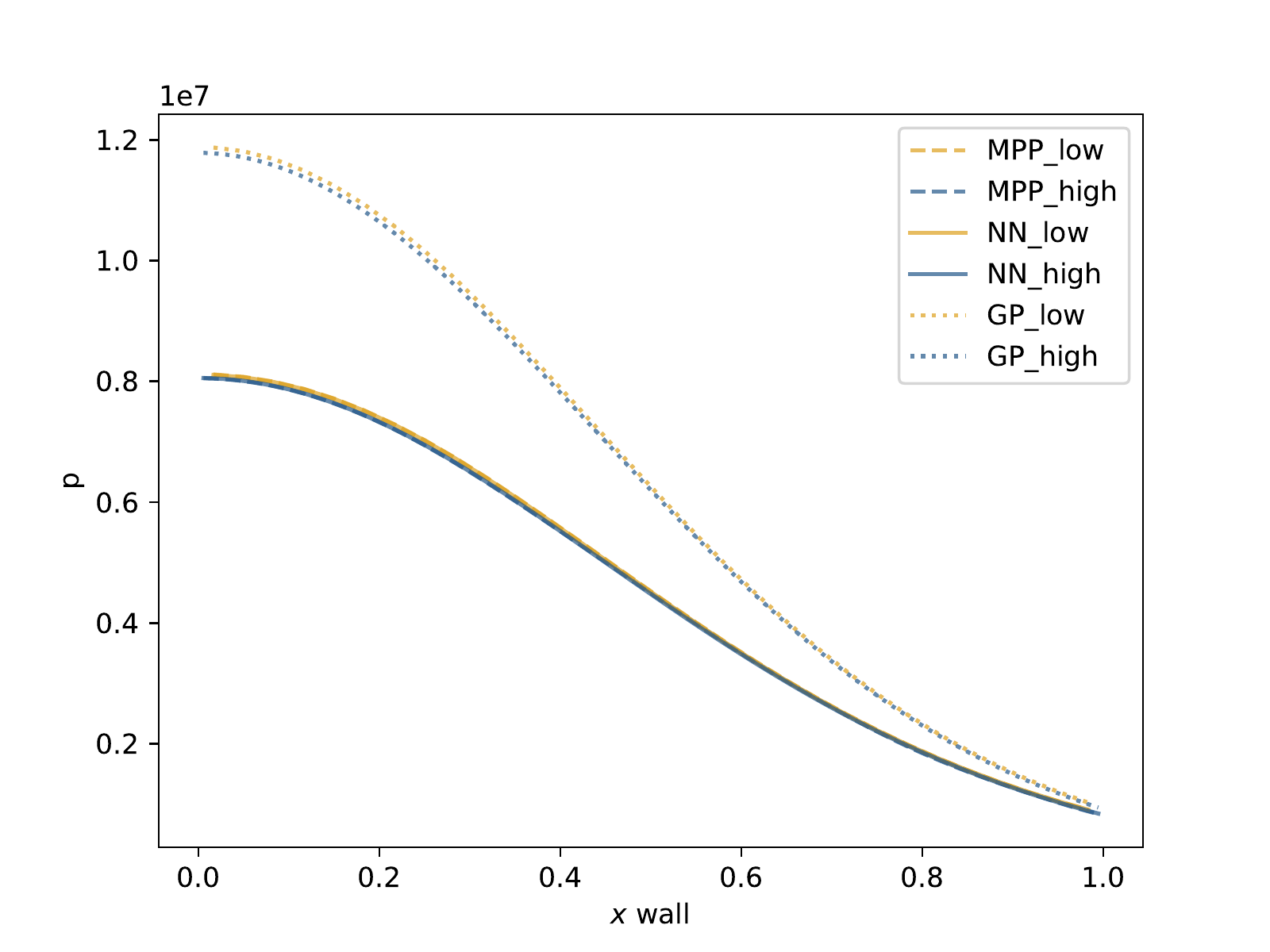}
      \includegraphics[width=1\textwidth]{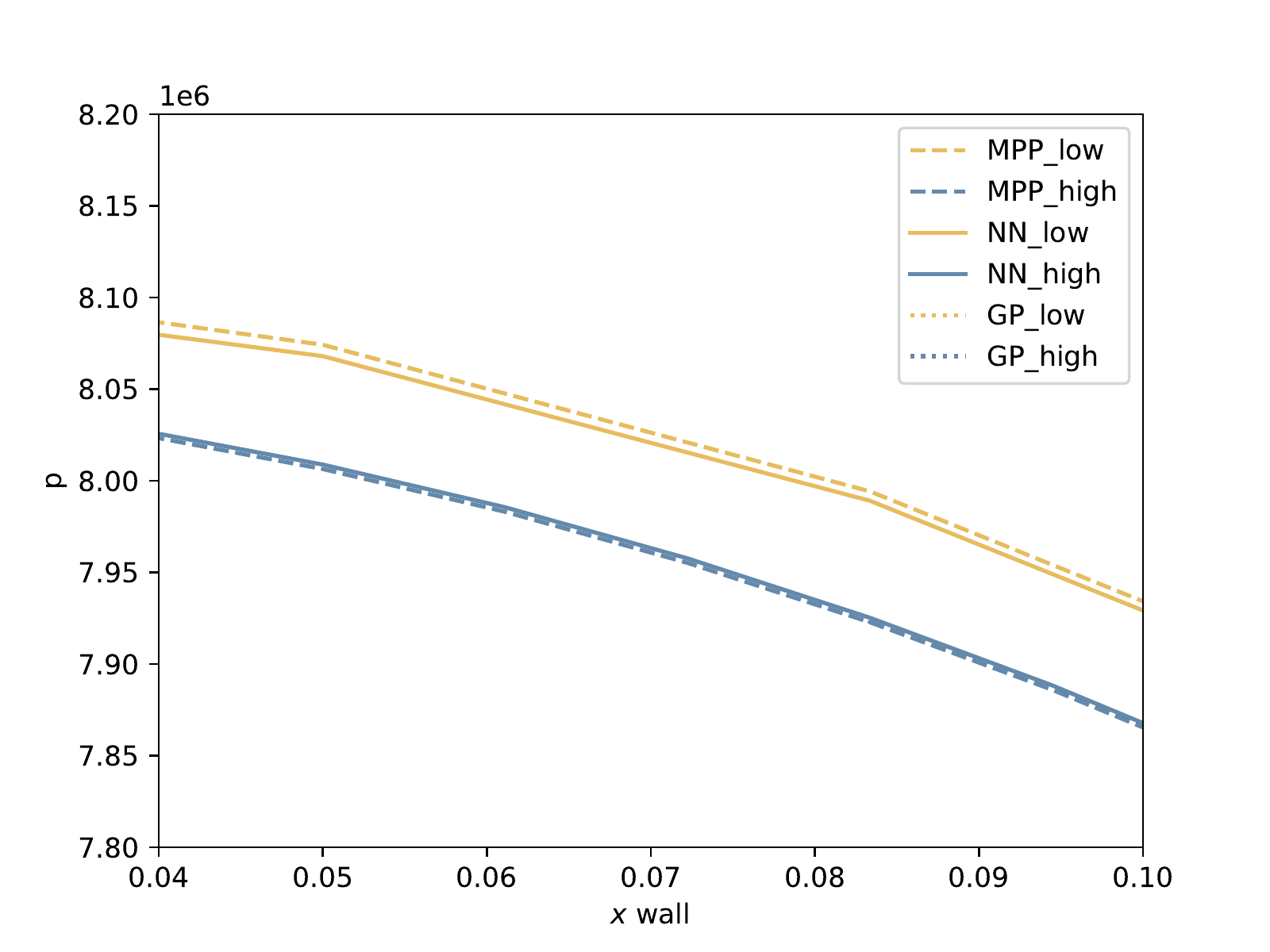}
    \end{center}
  \caption{Pressure profile}
  \end{subfigure}
  \begin{subfigure}{0.49\textwidth}
    \begin{center}
      \includegraphics[width=1\textwidth]{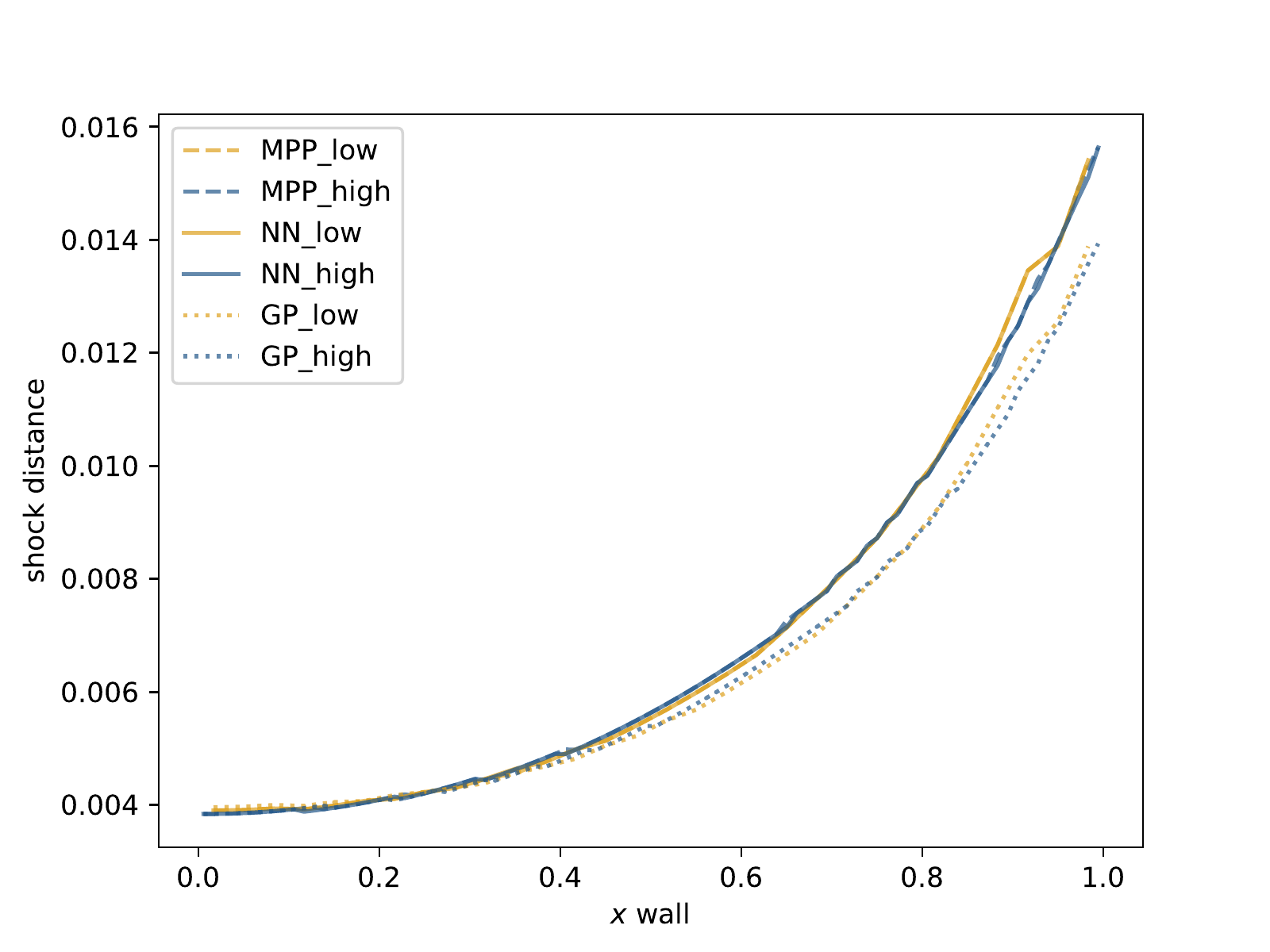}
      \includegraphics[width=1\textwidth]{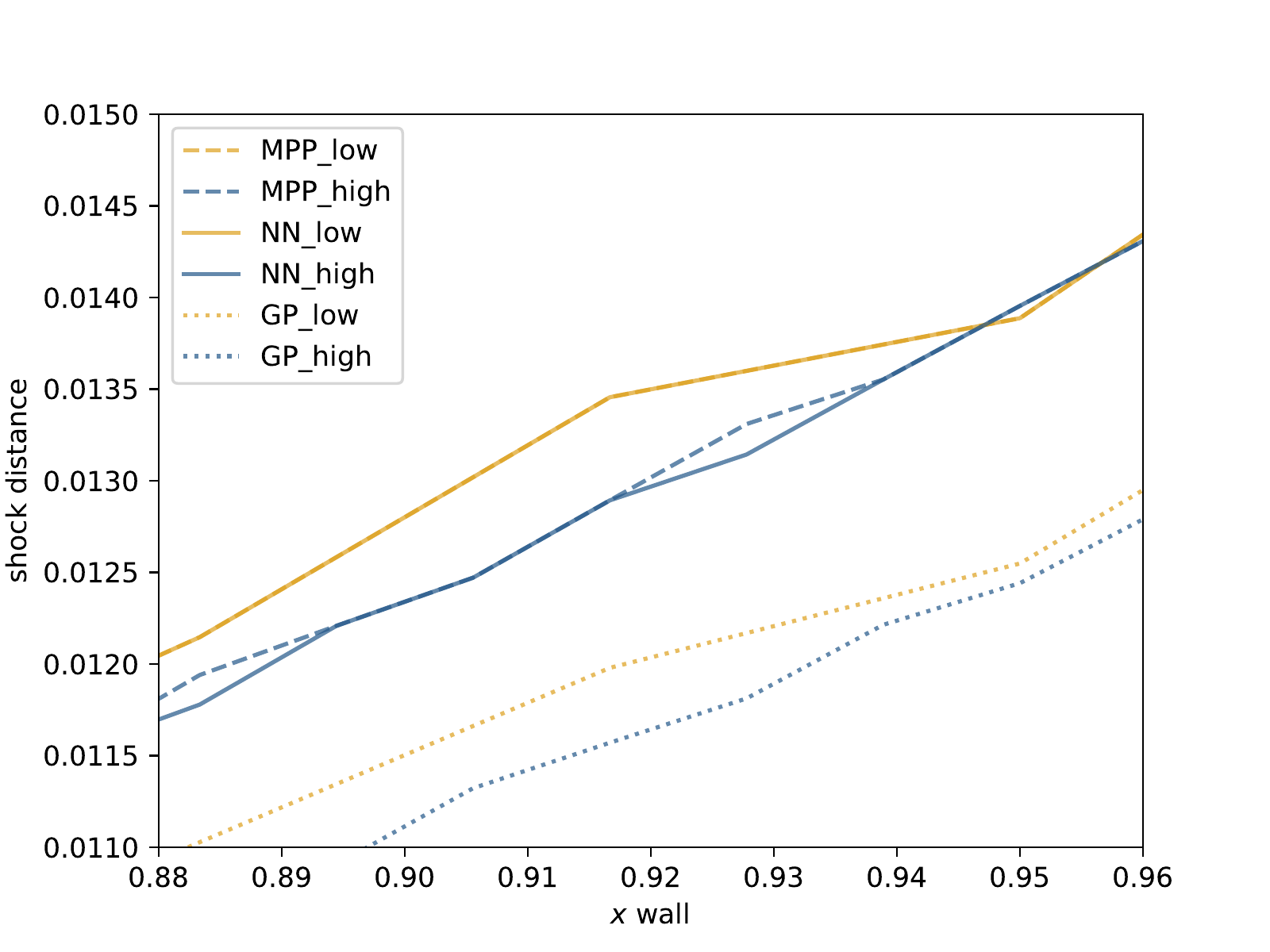}
    \end{center}
    \caption{Shock distance}
  \end{subfigure}
   \caption{Top: pressure profile and shock distance projected on the surface of the object. Bottom: same quantities with a zoom on the highest error area for the hybrid code. }\label{fig:profileshock}
\end{figure}

The results of figure \ref{fig:field} are rather qualitative. Let us progressively switch to more quantitative ones. For this, on figure \ref{fig:profileshock}, we extract the
pressure on the surface of the object (figure \ref{fig:profileshock} left) and the distance of the shock (figure \ref{fig:profileshock} right) for the three different codes. Note
that the suffixes \_high and \_low corresponds to two different meshes: 
\begin{itemize}
  \item [\_low] is for low-resolution mesh ($30 \times 100$),
  \item [\_high] is for high-resolution mesh ($90 \times 100$) and is $3$ times more refined in the direction of the shock. 
\end{itemize}
First, with figure \ref{fig:profileshock} (right column), we can see that the
perfect gas closure is also coarse for the shock distance: the model error (i.e. between MPP and PG independently of the observable of interest) is way more important than the
discretization error (as the error between the two meshes is small in comparison to the differences between PG and MPP).
This justifies taking the chemistry of the phenomenon into account for the simulation. 
Now, for those two observables at the surface of the object, we can see that the NN error is even smaller than the discretization error as the dotted lines are the closest to the
full lines. 
Finally, without zooming, no visual differences can be detected between MPP and NN. 
To assess the prediction error of NN, we have to look at more quantitative results:
table \ref{tab:profileerrortab} 
displays the $L_2$ and $L_{\infty}$ normalized errors for these curves with respect to MPP\_low and MPP\_high together with the run-times of the different codes:
Several comments can be made regarding these results:
\begin{itemize}
  \item The errors between PG (\_low and \_high) and MPP\_high are at least three decades higher than the errors between the results of NN and MPP, regardless of the resolution, which illustrates the need for simulating chemical reactions. 
  \item The errors of NN\_high and NN\_low are comparable to that of MPP\_low when MPP\_high is taken as the reference. Th error between NN\_low and MPP\_low is even lower.
  \item However, the run-times of NN are closer to those of PG with $220 s.$ and $529 s.$ for NN and $81 s.$ and $211s.$ for PG, against $4090s.$ and $9478s.$ for MPP. Note also that NN\_high is still faster than the MPP\_low while being more accurate.
\end{itemize}
%
%
%
%
%
\begin{table}[!h]
  \centering
  \begin{tabular}{llll}
    &  MPP\_low (ref) &  NN\_low & PG\_low\\
    \hline
  Time (s)  & $4090$ & $220$ & $81$ \\
    
  Impr. ($\times$) & - & $18.7$ & $58.5$ \\
    \hline
  Pressure &&& \\
  $L^2$        & - & $6.06 \times 10^{-7}$ & $7.74 \times 10^{-2}$ \\
  $L^{\infty}$ & - & $1.13 \times 10^{-3}$ & $4.64 \times 10^{-1}$ \\
    \hline
  Shock dist. &&& \\
  $L^2$        & - & $7.85 \times 10^{-7}$ & $ 1.70 \times 10^{-3}$ \\
  $L^{\infty}$ & - & $4.85 \times 10^{-3}$ & $ 9.89 \times 10^{-2}$ \\
\vspace{0.2cm}
\end{tabular}
  \resizebox{\textwidth}{!}{\begin{tabular}{lllllll}
      &  MPP\_high (ref) &  MPP\_low & NN\_high & NN\_low & PG\_high & PG\_low\\
      \hline
    Time (s)  & $9478$ & $4090$ & $529$ & $220$ & $211$ & $81$ \\
      
    Impr. ($\times$) & - & $2.3$ & $17.9$ & $43.1$ & $44.9$ & $117$ \\
      \hline
    Pressure &&&&&& \\
    $L^2$        & - & $3.19 \times 10^{-5}$ & $\bf1.13 \times 10^{-6}$ & $4.00 \times 10^{-5}$ & $7.65 \times 10^{-2}$ & $8.15 \times 10^{-2}$\\
    $L^{\infty}$ & - & $9.07 \times 10^{-3}$ & $\bf2.22 \times 10^{-3}$ & $9.82 \times 10^{-3}$ & $4.62 \times 10^{-1}$ & $4.75 \times 10^{-1}$ \\
      \hline
    Shock dist. &&&&&& \\
    $L^2$        & - & $9.14 \times 10^{-5}$ & $\bf 6.55 \times 10^{-6}$ & $9.19 \times 10^{-5}$ & $1.65 \times 10^{-3}$ & $1.29 \times 10^{-3}$\\
    $L^{\infty}$ & - & $3.60 \times 10^{-2}$ & $\bf 1.08 \times 10^{-2}$ & $3.60 \times 10^{-2}$ & $1.08 \times 10^{-1}$ & $9.00 \times 10^{-2}$ \\
  
  \end{tabular}}
  \caption{Execution times and normalized errors for the different codes. }\label{tab:profileerrortab}
  \end{table}

To sum up, on the one hand, MPP and NN are comparable in terms of error, and on the other hand, NN is almost one decade faster than MPP. These results are comforting and motivate us to study the method further and discuss the possibility of having guarantees with the hybrid code. 

The error of NN was comparable to that of MPP\_low on this prediction, but it was on one single prediction. This evaluation process is insufficient to state whether the error is acceptable. In order to ensure prediction guarantees, which are mandatory for using codes in production, we have to go deeper into the analysis.

In the next sections, we introduce two ways to obtain guarantees on the predictions of the hybrid code. The first ensures to have exactly the same prediction accuracy as
the fine reentry code (i.e. MPP) but brings additional computations. The second compares the error made with the hybridization of the reentry code with other sources of errors that are ubiquitous in numerical simulation to assess the acceptability of the hybrid code. If the hybridization error is lower than other errors, it is then possible to use the hybrid code (NN) at full speed.

\subsection{Zero-error guarantees of the hybrid code}
\label{sec:zeroerr}

As we mentioned earlier, the reentry simulation code is an iterative solver, see algorithm \ref{reentry_code}. It is initialized with a guess solution, which is uninformative - usually, the same value over the entire mesh - and iterations are made until a certain convergence criterion is reached. 
In this section, we suggest first executing the hybrid code (NN), using its prediction as initialization for the classical code (MPP). Then, the classical solver may hopefully converge in fewer iterations since the initialization is supposed to be close to the convergence point.
The new structure of the code is presented in algorithm \ref{reentry_code_mpp_nn}.\\ \ \\ 
    \begin{algorithm}[H]
    \caption{Core of the code with a call to a neural network surrogate model of Mutation++ in order to initialize the guess of a classical computation.}
        \label{reentry_code_mpp_nn}

      \textit{\# general initialization (mesh, quantities on mesh etc.)}

      initialise\_guess\_vector\_of\_unknowns\_on\_mesh($U^0$,$\x^0$)

      \textit{\# $1^{st}$ step with the neural network surrogate model}

      \While {\text{convergence\_criterion\_not\_satisfied}}{

      $U^{n+\frac12}=$solve\_Euler\_equations($U^n$,$\x^{n}$)

      $\x^{n+1}, U^{n+1}=$call\_NN\_surrogate\_model($U^{n+\frac12}$,$\x^{n}$)

      $U^n \leftarrow U^{n+1}$

      $\x^n \leftarrow \x^{n+1}$

                 }

      \textit{\# $2^{nd}$ step with Mutation++ using the NN results as the initial guess}

      \While {\text{convergence\_criterion\_not\_satisfied}}{

      $U^{n+\frac12}=$solve\_Euler\_equations($U^n$,$\x^{n}$)

      \For{$i\in\{1,...,N_{\mathcal{D}}\}$}{
      
        $\x_i^{n+1}, U^{n+1}=$minimize\_Gibbs\_free\_energy\_with\_mutation++($U^{n+\frac12}_i$,$\x^{n}_i$)
    }
      $U^n \leftarrow U^{n+1}$

      $\x^n \leftarrow \x^{n+1}$

                 }

    \end{algorithm} \ \\
The strategy described in algorithm \ref{reentry_code_mpp_nn} gives the following results. Once the prediction of NN\_low is given as a guess of the original
code, MPP\_low, the latter reaches convergence in $163 s.$. If we sum the run times of both codes, in that case, the simulation takes $383s.$ which is $\times 10.6$ faster than MPP\_low alone, for a prediction whose accuracy is {\em exactly} the same as the ones of the original code.  This strategy is denoted by NN+MPP in the following paragraphs.
In that case, there is no need to compare the pressure profiles or the shock distances between the different options of the code since \textbf{the prediction are indistinguishable
from one another}.
In other words, we obtained an acceleration of a factor $10.6$ for the exact same accuracy, i.e. {\em with the same guarantees as the original code}.

\begin{remark}\label{rmk:general}
This approach is not specific to our reentry problem: it can be applied to any stationary simulations or in a general manner to any computation code involving an iterative solver whose iterations are not of interest as simulation outputs. 
\end{remark}

\subsection{Guarantees of acceptable error for the hybrid code}
\label{non0error}


As we have seen in the previous section, for our reentry simulations, it is possible to have the same guarantees with the hybrid code as with the original one at the price of a
smaller acceleration ($\times 10$ with guarantees instead of $18.6$).
Now, one could be interested in this $18.6$ factor of acceleration instead of $\times 10.6$ (or one could not have a stationary problem).
In order to make sure the error coming from neural network approximation remains acceptable, it is mandatory to be able to quantify it and compare it to the other sources of errors that are classically found in numerical simulations. 
In this section, we compare the error due to the use of a neural network within the hybrid code with these other errors, namely the model error, the discretization error, and the fluctuations relative to the sources of
uncertainty. 

Let us formalize this: suppose we are interested in observable $y$ (it can be the pressure on the surface of the object, the distance to the shock etc.). 
Let us denote by $y_r$ the ground-truth value of $y$ that we want to predict with a model. We denote the original code MPP by $\mathcal{M}$ and the hybrid code NN by $\widehat{\mathcal{M}}$, with $\mathcal{M}_{high}$ denoting the model of MPP\_high, and so on for the other MPP and NN codes. The models take parameters $\vx$ as input and output a prediction $\mathcal{M}(\vx)$ and $\widehat{\mathcal{M}}(\vx)$. In our case, the input vector $\vx$ contains the upstream pressure, temperature, and speed. The predictions can be expressed as

\begin{equation}\label{eq:uncertainty}
  \begin{dcases}
    \mathcal{M}(\vx) = y_r +  e_{\mathcal{M}}(\vx) = y_r +  \delta_{\Delta, \mathcal{M}}(\vx) + \delta_{\vx, \mathcal{M}}(\vx) , \\
    \widehat{\mathcal{M}}(\vx) = y_r + e_{\widehat{\mathcal{M}}}(\vx) = y_r + \delta_{\Delta, \widehat{\mathcal{M}}} (\vx) + \delta_{\vx,\widehat{\mathcal{M}}}(\vx) + \delta_{\vtheta,\widehat{\mathcal{M}}} (\vx) .\\
  \end{dcases}
\end{equation}

\Eqref{eq:uncertainty} emphasizes three different types of errors :
\begin{itemize}
  \item The discretization errors $\delta_{\Delta, \mathcal{M}}$ and $\delta_{\Delta, \widehat{\mathcal{M}}}$. The choice of the mesh used to run the simulation has an impact on the prediction error. A low mesh resolution may degrade the error, as we saw with MPP\_low, but the geometry of the mesh also has its impact.
  \item The parameters' errors $ \delta_{\vx, \mathcal{M}}$ and $\delta_{\vx,\widehat{\mathcal{M}}}$. In practice, we conduct numerical simulations because we are interested in the output of a phenomenon under specific conditions of interest. Nonetheless, we may have imperfect knowledge of these conditions of interest. This translates into uncertainties on the input vector $\vx$, which contains the values of the parameters that define the conditions of the simulation. These uncertainties have an impact on the model prediction and, therefore, on their prediction error.
  \item The neural network approximation error $\delta_{\vtheta,\widehat{\mathcal{M}}}$. In the hybrid code, NN, the neural network approximates $\mut$ with a certain error. This error propagates through the hybrid code, thereby affecting its prediction error. 
\end{itemize}

In order to ensure that NN yields reliable predictions, the neural network approximation error must be compared with parameters and discretization errors. If the former is at most similar to the two others, NN could be used safely.

\begin{remark}
  Another type of error is often specified when decomposing the error of numerical codes. This error is called modeling error and refers to the error that stems from modeling choices. In our case, the model error would be the error between PG and MPP, coming from the choice not to simulate the chemistry in PG. Another model error could come from the choice not to simulate the chemistry, to simulate it with more or fewer species, or to use Naver-Stokes equations rather than Euler equations. In addition, round-off errors could also be taken into account. In this work, we suppose they are negligible with respect to the other sources of errors. The reader interested in ways to quantify them can refer to \cite{demeurePhd} for example. 
\end{remark}

\begin{figure}[!t]
  \centering
  \begin{subfigure}{0.49\textwidth}
    \begin{center}
      \includegraphics[width=1\textwidth]{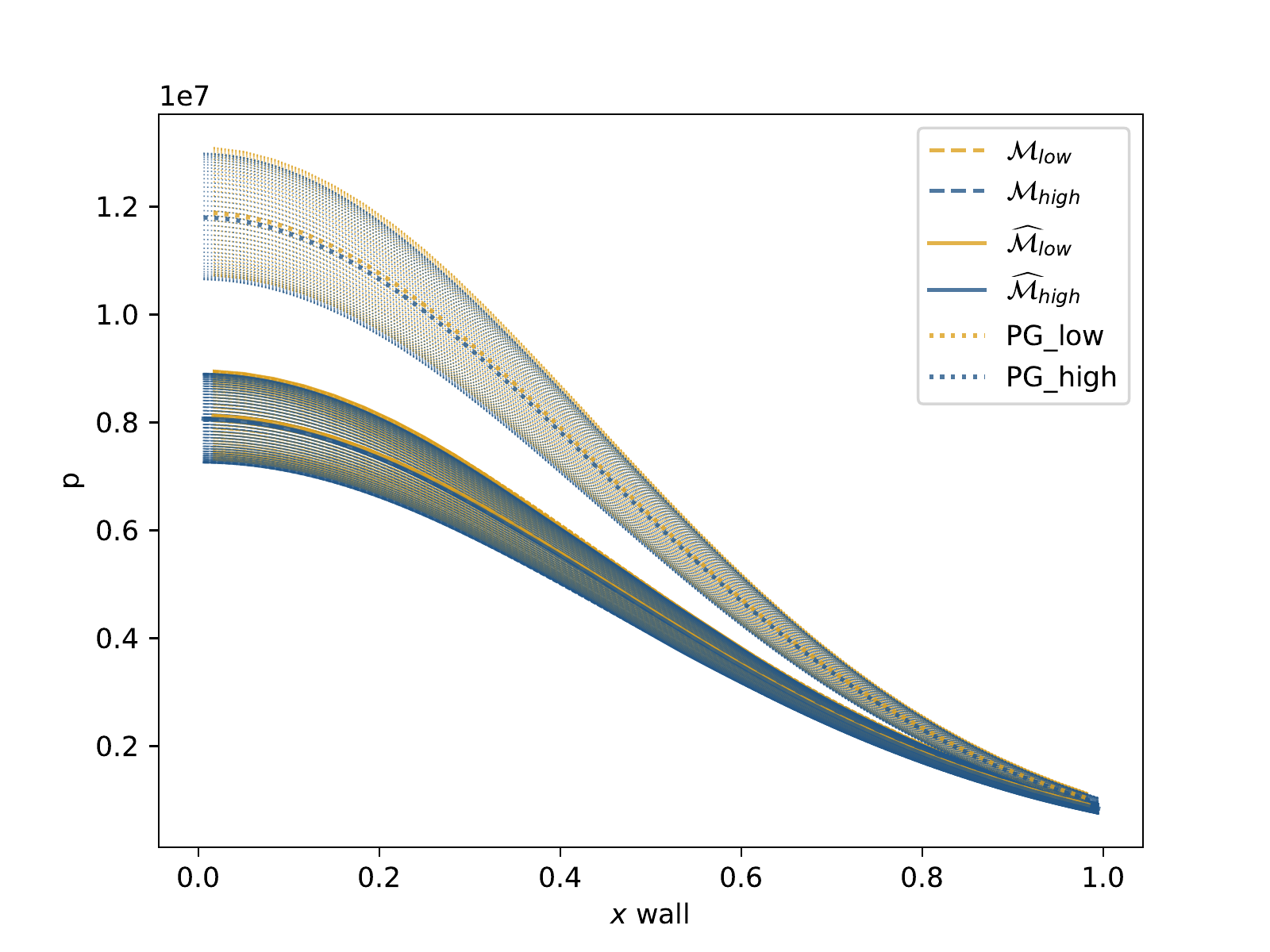}
      \includegraphics[width=1\textwidth]{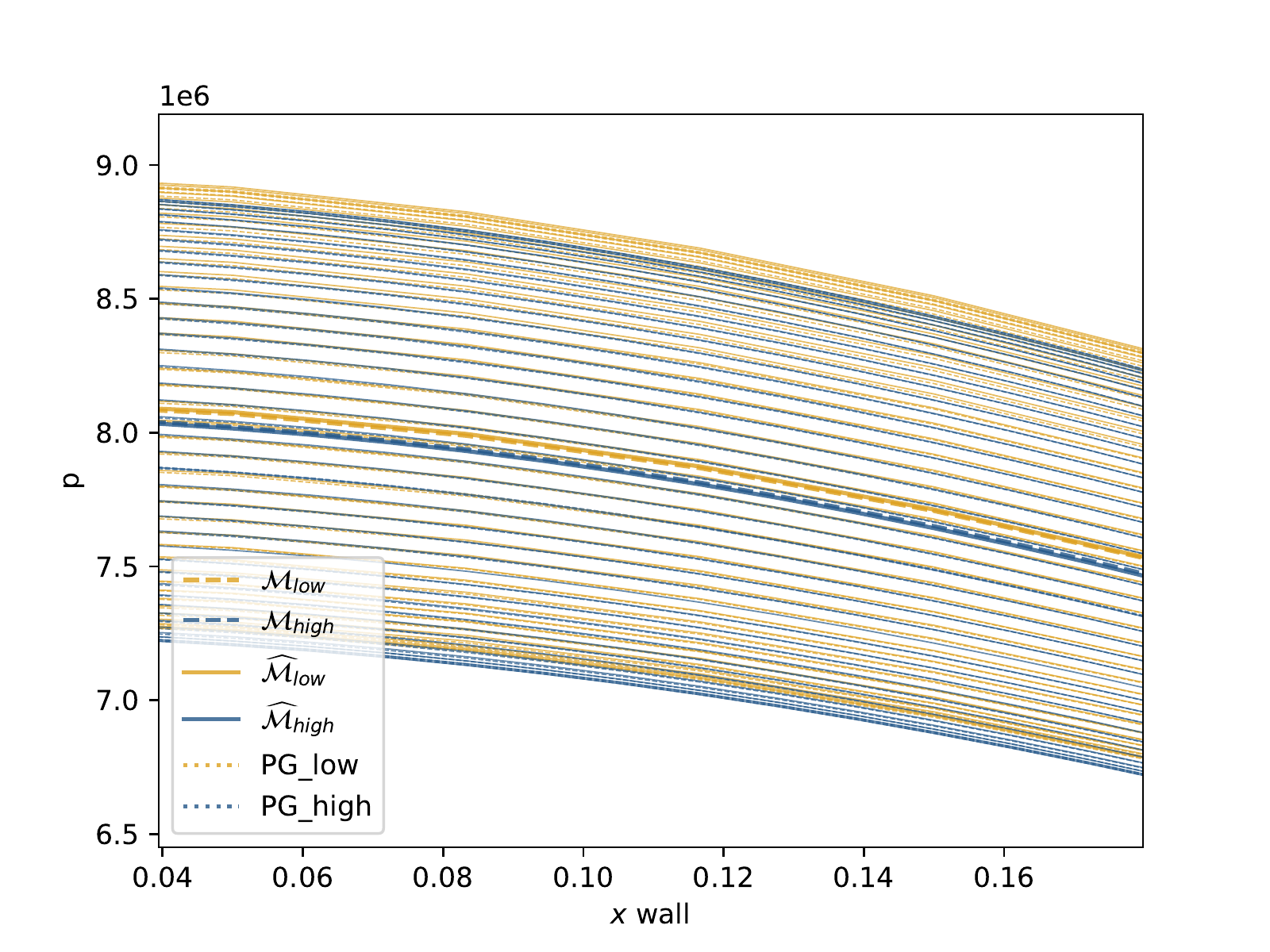}
    \end{center}
\caption{Pressure profile}
  \end{subfigure}
  \begin{subfigure}{0.49\textwidth}
    \begin{center}
      \includegraphics[width=1\textwidth]{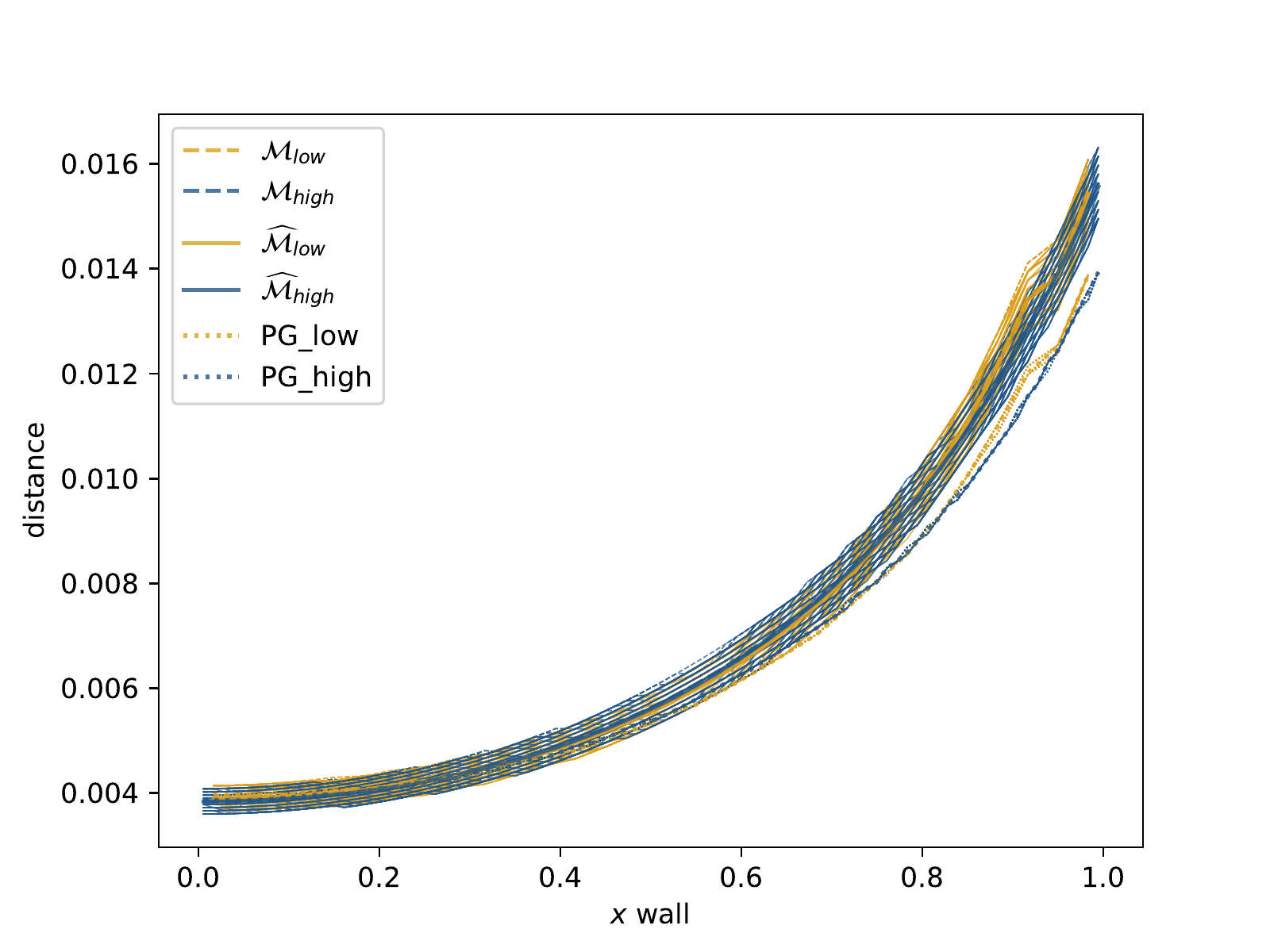}
      \includegraphics[width=1\textwidth]{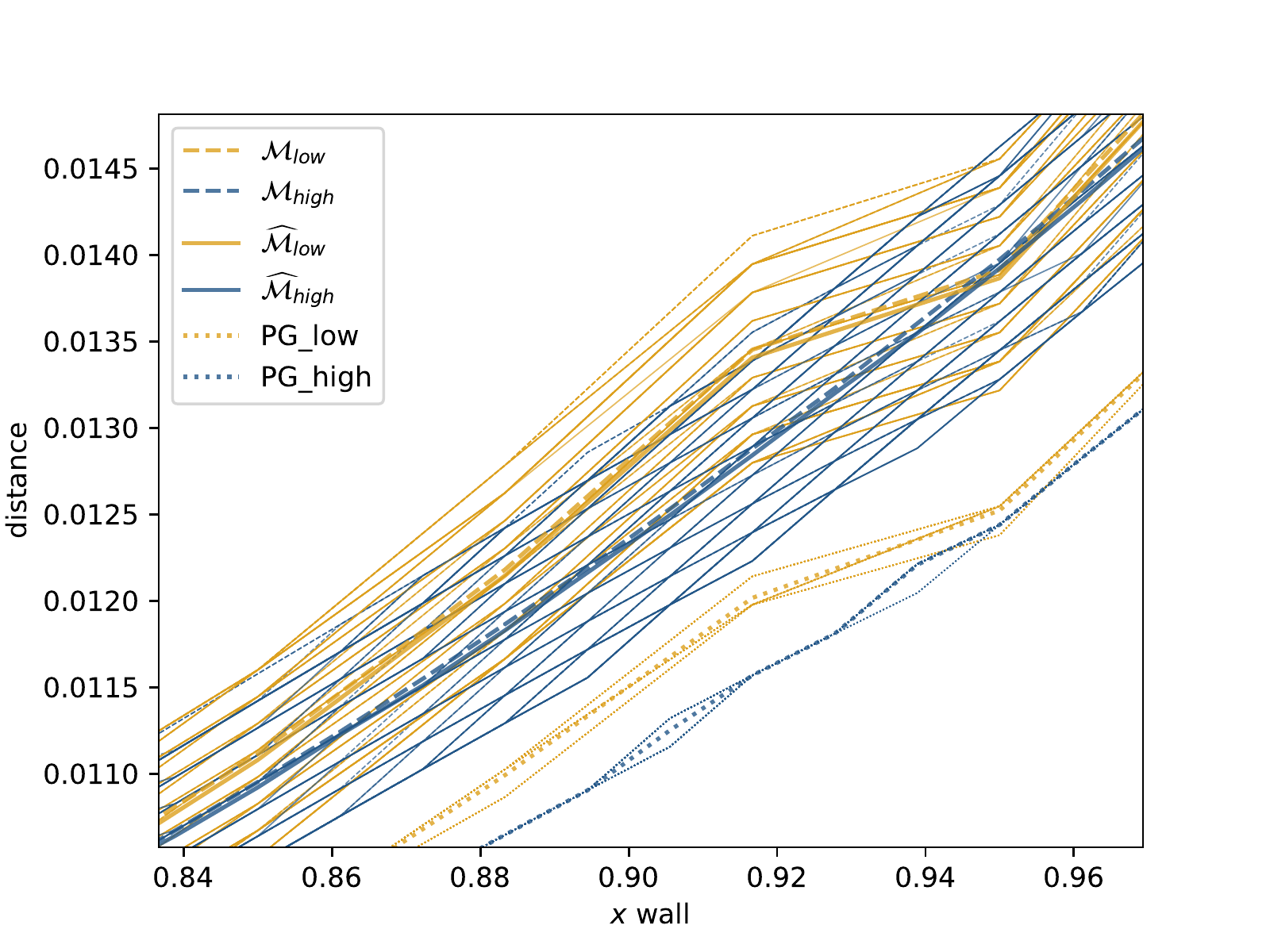}
    \end{center}
    \caption{Shock distance}
  \end{subfigure}\\
   \caption{Pressure profile and shock distance projected on the wall of the object, for $40$ different values of the upstream speed. In the bottom line, zoom of the curves in the highest error area.}\label{fig:errstudy}
\end{figure}

\subsubsection{Uncertainty propagation for reliable error comparison}

To compare the different errors, we introduce a perturbation in $\vx$, modeling the uncertainty on the upstream speed. Let $\rvx = (\rx, y, z)$, with $\rx \sim \mathcal{U}(0.95x, 1.05x)$, $x = 4930.83$ (the nominal value of the initial test case). The random variable $\rx$ traduces the uncertainty on the speed of the upstream field, and the values of $y$ and $z$ are the upstream pressure and temperature. We simulate the random variable $\rx$ on $N=40$ Gauss quadrature points $\{x_1,...,x_N\}$, with $x_i \in [0.95x, 1.05x]$, that are used to evaluate $\mathbb{E}[\mathcal{M}(\vx)]$ with $\mathcal{M}\in \{\mathcal{M}_{low}, \mathcal{M}_{high}, \widehat{\mathcal{M}}_{low}, \widehat{\mathcal{M}}_{high} \}$. Such an analysis allows us to study the effect of parameter uncertainty on the error, as well as statistically compare the different sources of error.

The mean $\mathbb{E}[\mathcal{M}(\vx)]$ and each of the $40$ curves are plotted in Figure \ref{fig:errstudy}. These graphs are quite loaded, but they highlight that the variability induced by parameter uncertainty is much higher than that coming from approximation and even discretization errors.

\begin{figure}[!h]
  \centering
  \begin{subfigure}{0.49\textwidth}
    \begin{center}
      \includegraphics[width=1\textwidth]{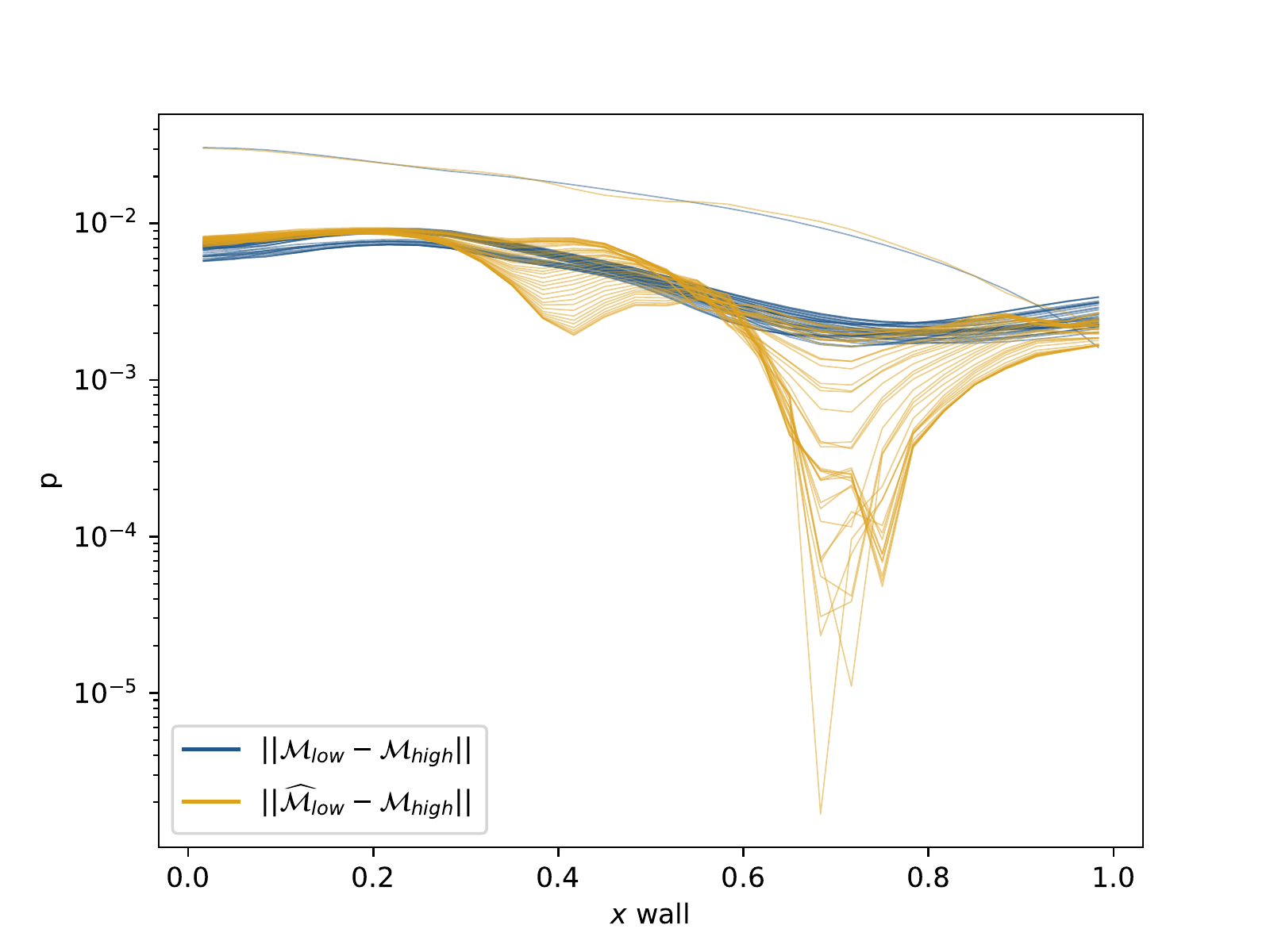}
    \end{center}
\caption{Pressure profile}
  \end{subfigure}
  \begin{subfigure}{0.49\textwidth}
    \begin{center}
      \includegraphics[width=1\textwidth]{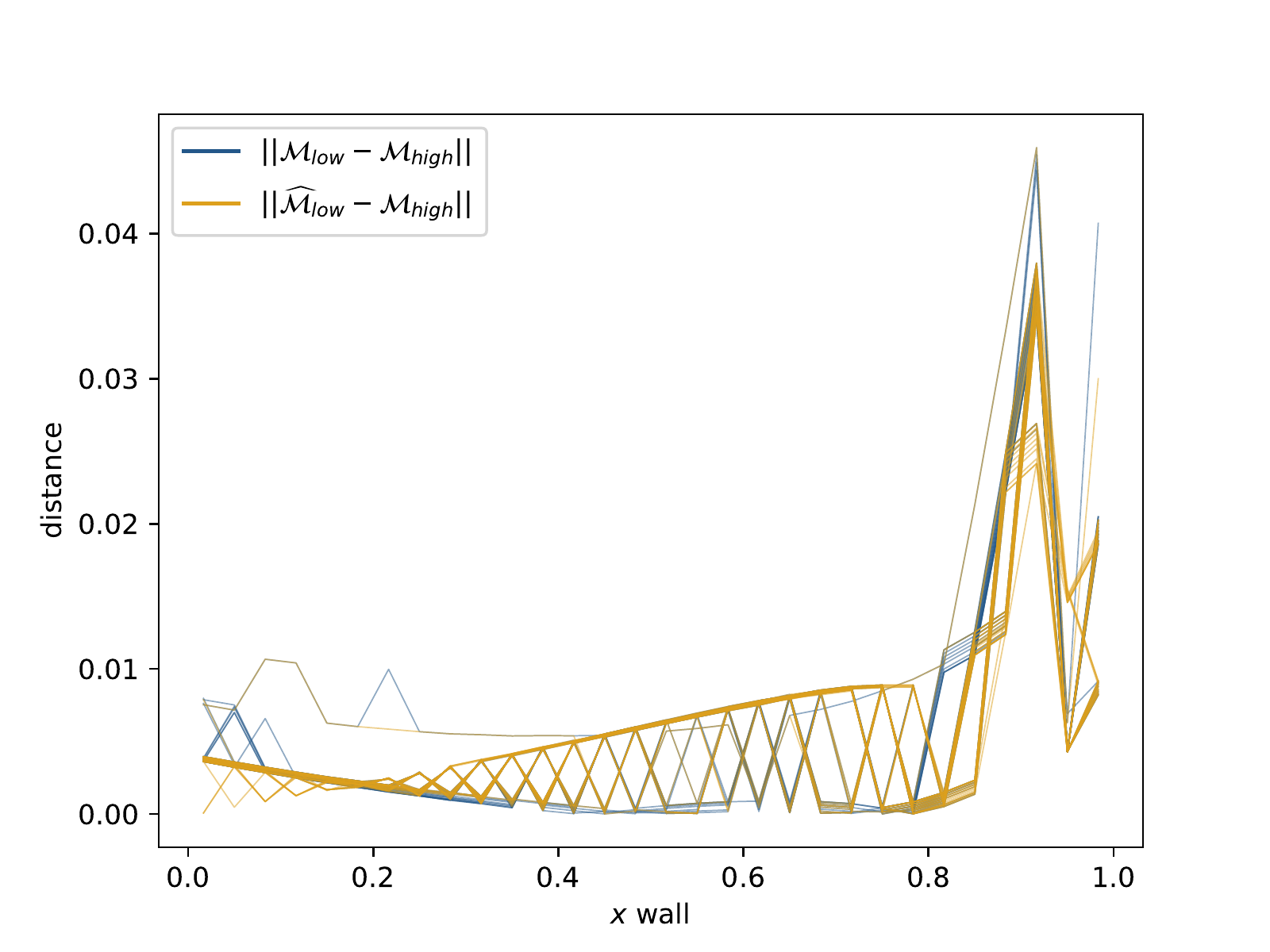}
    \end{center}
    \caption{Shock distance}
  \end{subfigure}\\
   \caption{Discretization errors $||\mathcal{M}_{low} - \mathcal{M}_{high}||$ and $||\widehat{\mathcal{M}}_{low} - \mathcal{M}_{high}||$ for the pressure profile and the shock distance for each of the $40$ different values of the upstream speed.}\label{fig:errdisc}
\end{figure}

Comparing discretization and neural network approximation errors is more subtle because it is not clear-cut in Figure \ref{fig:errstudy}. To do so, we directly compare the discretization error of $\mathcal{M}_{low} $ with the approximation error of $\widehat{\mathcal{M}}_{low}$ and $\widehat{\mathcal{M}}_{high}$ under parameters uncertainty. First, we plot $||\mathcal{M}_{low} - \mathcal{M}_{high}||$ and $||\widehat{\mathcal{M}}_{low} - \mathcal{M}_{high}||$ for the $40$ values of $x_i$ in Figure \ref{fig:errdisc}. Here, $||.||$ is the normalized absolute difference evaluated point-wise in the pressure profile and the shock distance. At first sight, the approximation error seems to be lower (for the pressure profile) or equivalent (for the shock distance) to the discretization error. 

\begin{figure}[!h]
  \centering
  \begin{subfigure}{0.49\textwidth}
    \begin{center}
      \includegraphics[width=1\textwidth]{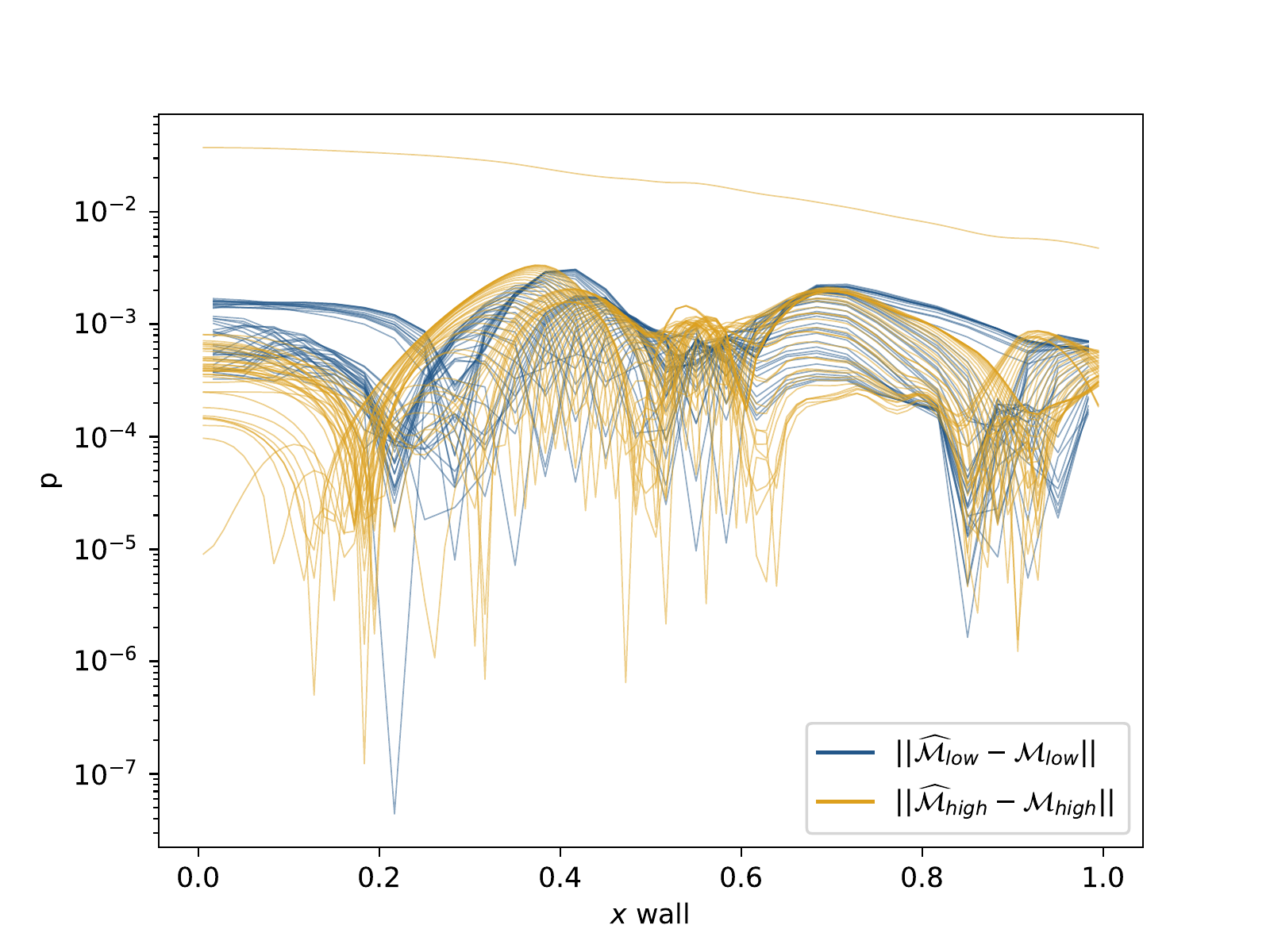}
      \includegraphics[width=1\textwidth]{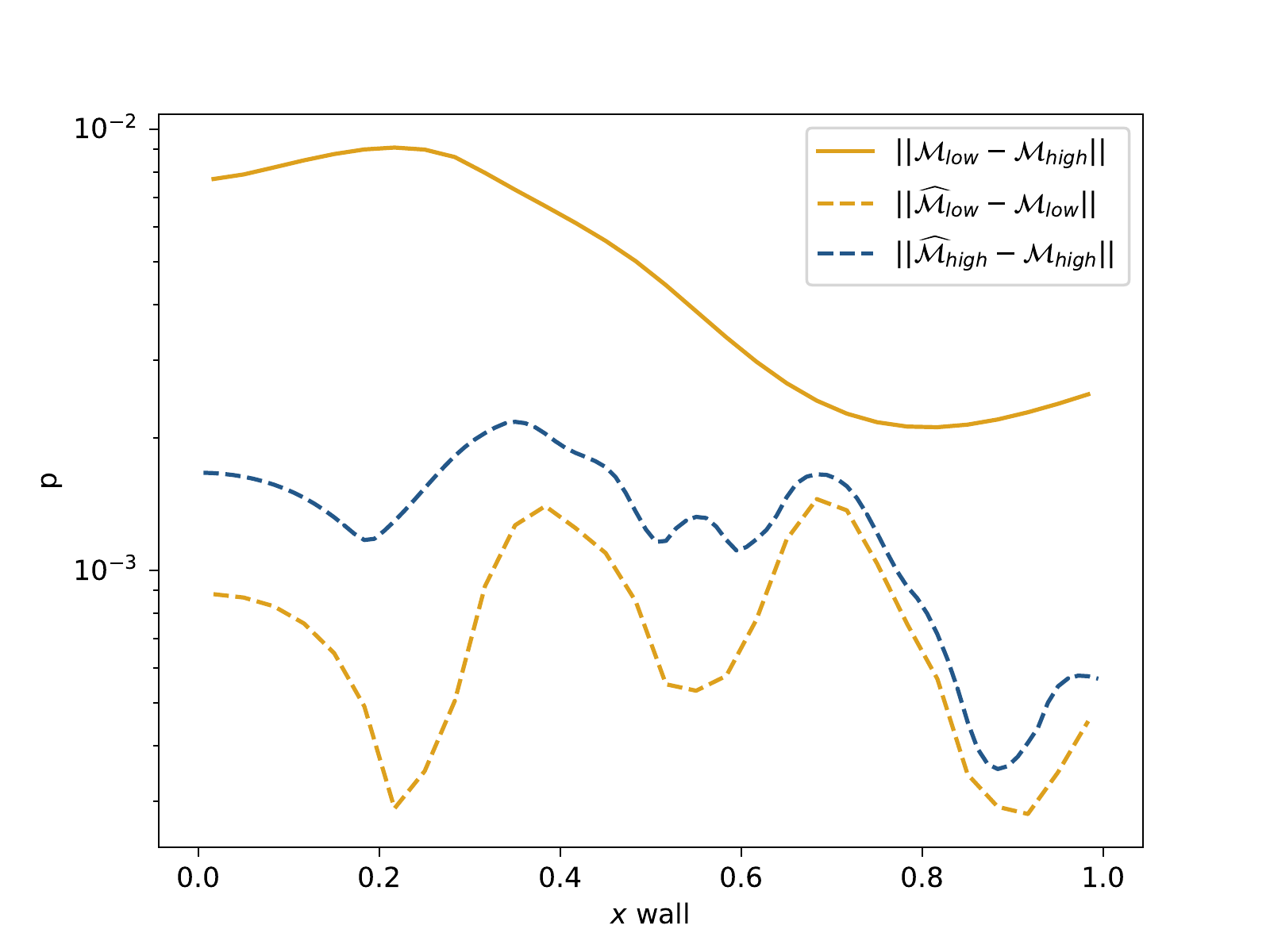}
    \end{center}
\caption{Pressure profile}
  \end{subfigure}
  \begin{subfigure}{0.49\textwidth}
    \begin{center}
      \includegraphics[width=1\textwidth]{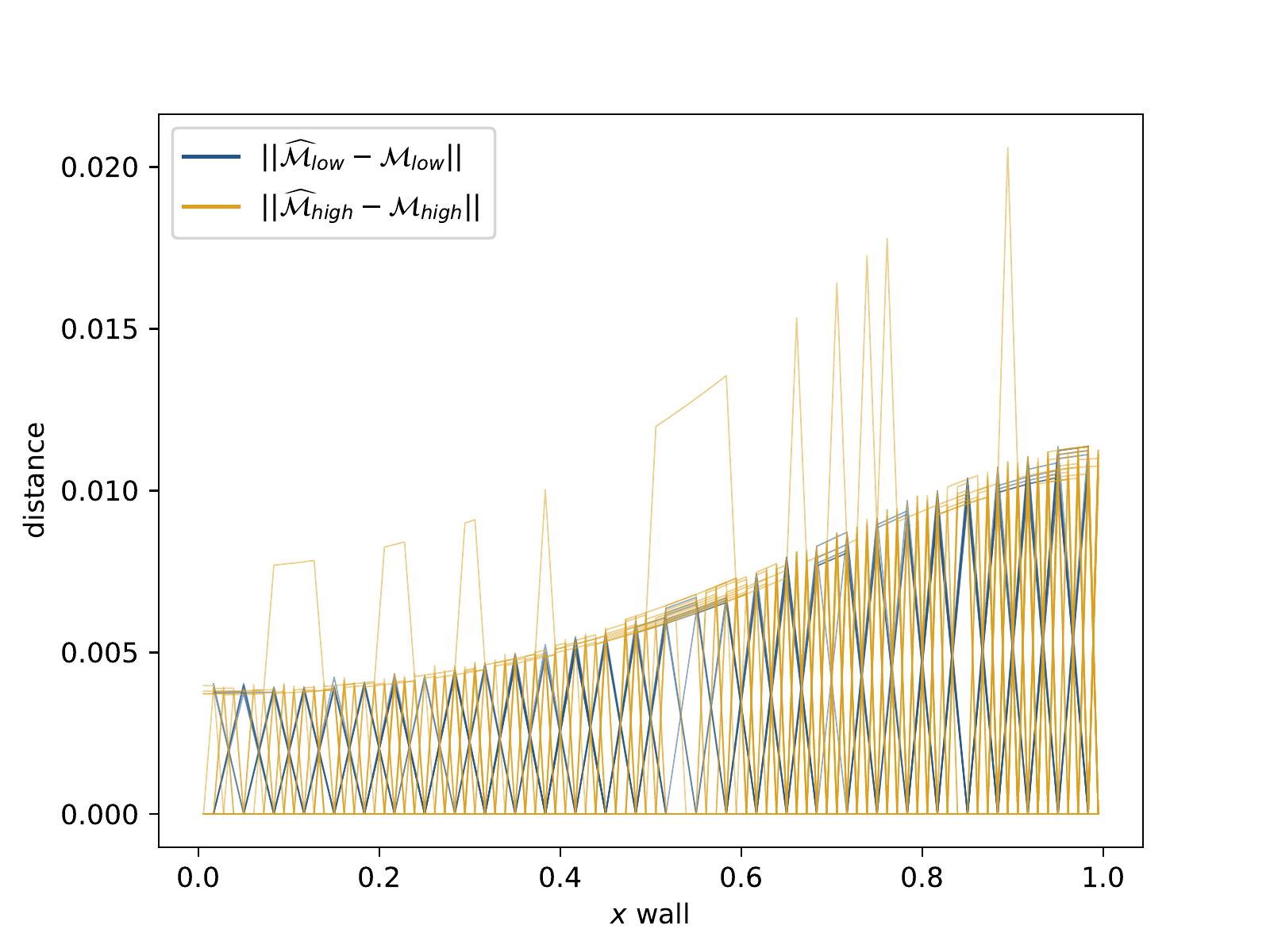}
      \includegraphics[width=1\textwidth]{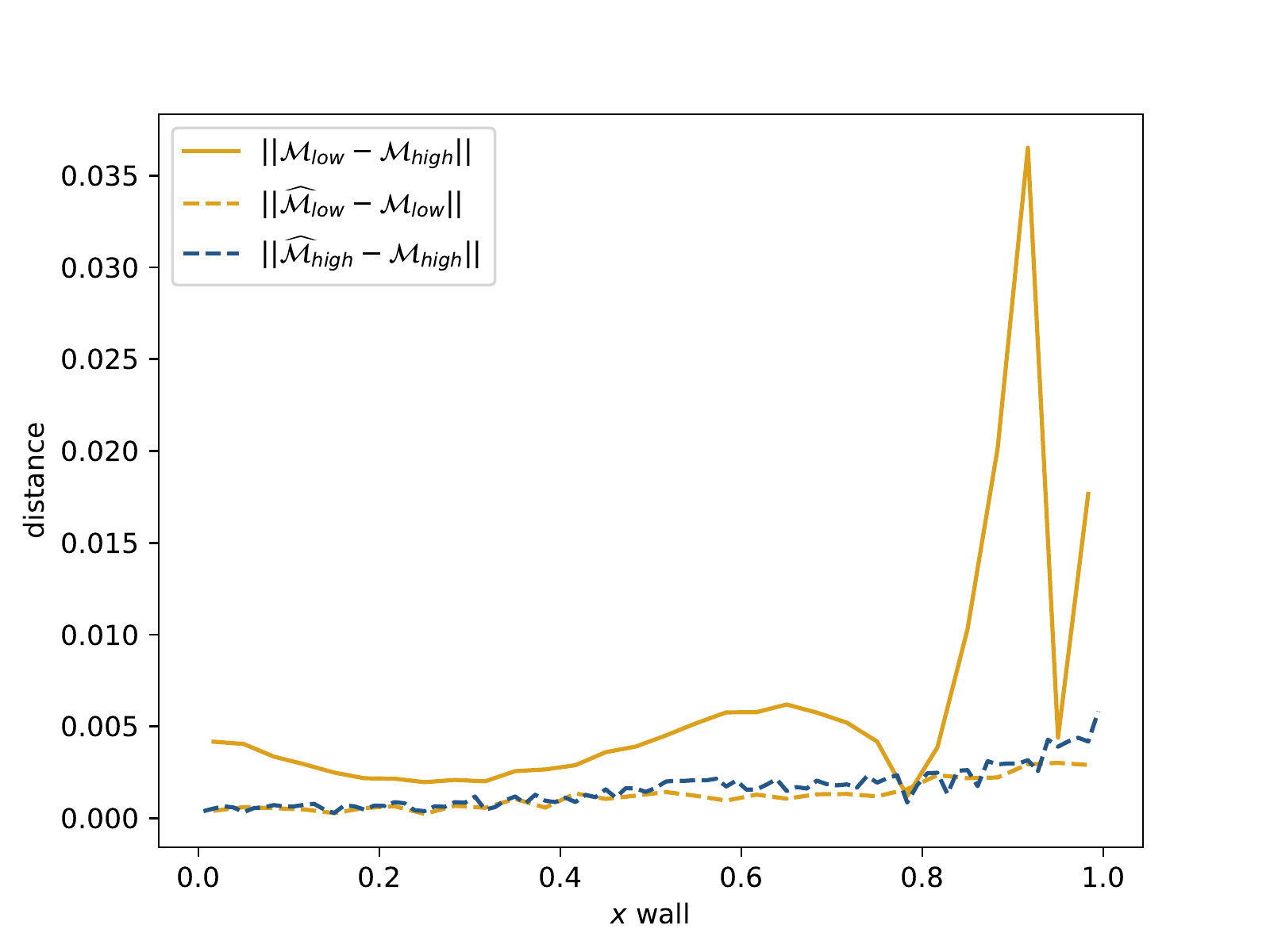}
    \end{center}
    \caption{Shock distance}
  \end{subfigure}\\
   \caption{Top: approximation errors $||\widehat{\mathcal{M}}_{low} - \mathcal{M}_{low}||$ and $||\widehat{\mathcal{M}}_{high} - \mathcal{M}_{high}||$ for the pressure profile and the shock distance for each of the $40$ different values of the upstream speed. Bottom: comparison of the mean of the discretization and approximation errors.}\label{fig:errapp}
\end{figure}

We confirm this observation by plotting $||\widehat{\mathcal{M}}_{low} - \mathcal{M}_{low}||$ and $||\widehat{\mathcal{M}}_{high} - \mathcal{M}_{high}||$ in Figure \ref{fig:errapp}. We plot the $40$ curves corresponding to each $x_i$ and the mean estimated using the Gauss quadrature. It clarifies the comparison and strengthens the conclusion that approximation error is lower than both discretization and parameters error. 

This error study shows that the neural network's approximation error can be small compared to other types of errors. In this case, the hybrid code is reliable, which is a strong argument in favor of the use of hybrid simulation codes.
%

\subsubsection{Benefits of NN+MPP to obtain reference predictions}

To conduct this experiment, we never executed $\mathcal{M}_{low}$ and $ \mathcal{M}_{high}$ entirely, but always used an initialization from the prediction of $\widehat{\mathcal{M}}_{low}$ and $ \widehat{\mathcal{M}}_{high}$ (NN+MPP, as described in Section \ref{sec:zeroerr}). The advantages were twofold. First, the study was much faster (approximately by a factor of $10$). Second, for some $\{x_1,...,x_N\}$, MPP did not converge, perhaps because of numerical instabilities. Initializing MPP using the hybrid code solved this problem. This echoes remark \ref{rmk:general} on the artefacts in predictions of $C_p$ and $C_v$ by $\mut$.

\section{Discussion and Perspectives}
\label{ccl}

In this work, we studied the acceleration of a simulation code involving the coupling between hypersonic fluid dynamic and chemical equilibrium. The simulation code is computationally expensive because of the chemical equilibrium simulator (Mutation++) that has to be called at each cell of the simulation mesh. This motivated using neural networks as surrogate models approximating chemical reactions because (1) their execution can be easily vectorized, so they can be called efficiently on a whole mesh, and (2) they can be trained on an extensive database constructed out of Mutation++.

By taking care of constructing neural networks not only accurate but also cost-effective thanks to the methodology of \cite{hsic:novello}, we achieve an acceleration factor of $18.7$ for the hybrid simulation code (NN).  The obtained prediction is qualitatively indistinguishable and quantitatively very close to that of the original simulation code (MPP).

Nonetheless, though promising, these results are insufficient to ensure NN guarantees for reliable use in production. We describe two simple methodologies for obtaining guarantees to tackle that problem. The first methodology ensures the same guarantees as MPP by initializing it with the prediction of NN. However, the acceleration factor decreases to $10$. The second methodology relies on the consideration that many sources of errors affect the prediction of numerical simulation code, even without hybridization. We conduct a statistical study of the effect of some of these errors, namely the parameter's uncertainty and discretization errors, and compare them to the neural network's approximation error. It turns out that in this test case, the neural network's error is negligible with respect to the others, so we tend to recommend using the hybrid code with its full acceleration factor of $18.7$ safely. We would like to conclude the paper with a discussion on the perspectives of this work.

\subsection{Towards a general approximation of Mutation++}

In this section, the methodology for approximating Mutation++ consists of constructing a training database and fitting a neural network. This is a strong advantage of this method since the neural network can be used in any simulation code involving the same chemical equilibrium setting (as we saw in section \ref{non0error}). 

However, the neural network is trained for a fixed output dimension corresponding to the number of species. It cannot be used for test cases that involve different species because it requires constructing a new training database for each different chemical setting. 

It would be interesting to investigate the use of transfer learning to make the approach easily applicable to other chemical reactions. One could pre-train a neural network once for a high number of different species and with a large database constructed out of Mutation++. Then, one could find a simple way to adapt this neural network for each different test case, for instance, with a least-squares linear regression on the feature space of the pre-trained network, using a smaller data set. 

\subsection{A general pattern for hybridization}

Our approach is not specific to hypersonic reentry coupled with chemical equilibrium. The idea of constructing a hybrid code based on both numerical simulation and machine learning has already been explored in previous works. In molecular simulations, \cite{hybrid_pes_0} use a neural network approximation of potential energy surface, and \cite{hybrid_fes_0} use Gaussian processes to sample Gibbs free energy surface, opening the avenue for applications based on such methodologies \cite{hybrid_2,hybrid_3,hybrid_4,hybrid_5}. \cite{hybrid_1} use neural networks to approximate physical components of multi-physics problems for electro-thermal simulation when conducting electrosurgery. In \cite{gilles1, gilles2}, the authors approximate non-local thermodynamic equilibrium in the simulation of inertial confinement fusion.

More formally, a multi-physics simulation code often solves a coupled system of several components that model different physics. To simplify the framework, we only consider a code with two coupled systems of equations. The system can be written :

\begin{subequations}
  \begin{empheq}[left={}\empheqlbrace]{align}
    F_1(U, \x, \valpha) &= 0\label{eq:a},\\
    F_2(U, \x, \valpha) &= 0\label{eq:b},
  \end{empheq}
\end{subequations}

where
\begin{itemize}
    \item $U$ and $\x$ are vectors of unknowns, 
    \item $\valpha$ is a vector of physical parameters, that are not computer during the simulation (e.g. physical or chemical constants), 
    \item $F_1$ and $F_2$ are mathematical (possibly differential) operators. In our test case, \eqref{eq:a} is Euler equations and \eqref{eq:b} is the Gibbs free energy minimization equation, which is behind Mutation++.
\end{itemize}

In such cases, most of the time, the solver needs to solve \eqref{eq:b} repeatedly in order to solve \eqref{eq:a}. As a result, \eqref{eq:a} is costly to solve because it regularly calls for the resolution of \eqref{eq:b}. The approach of approximating $F_2$ with a neural network and leveraging its implementation to process an entire mesh in a batch input fashion is, therefore, more general than the present hypersonic reentry test case. Numerical simulations, in general, could benefit from such an approach, strengthened by the described methodology for guaranteeing the results of the obtained hybrid code.

\subsection{Hybrid simulation codes as an additional lever for acceleration}
    
We would like to point out that even if the obtained hybrid code no longer uses the code part that is approximated by a neural network, this part is still crucial for constructing the hybrid code. Indeed, a good training set is mandatory to ensure neural network accuracy, and the original code part is key to achieving that accuracy. That is why we do not claim deep learning to replace simulation codes. Instead, we argue that it should be seen as an additional step in constructing simulation codes, allowing for significant accelerations.

\bibliography{refs}

\newpage
\section*{Appendix A: Additional plots}\label{appA}

In this appendix, we gather additional plots of the effect of the neural network input dimension on its execution time for different widths. These plots are complementary with those of Section \ref{capabilities}.

\begin{figure}[!h]
  \centering
  \begin{subfigure}{0.49\textwidth}
    \begin{center}
      \includegraphics[width=1\textwidth]{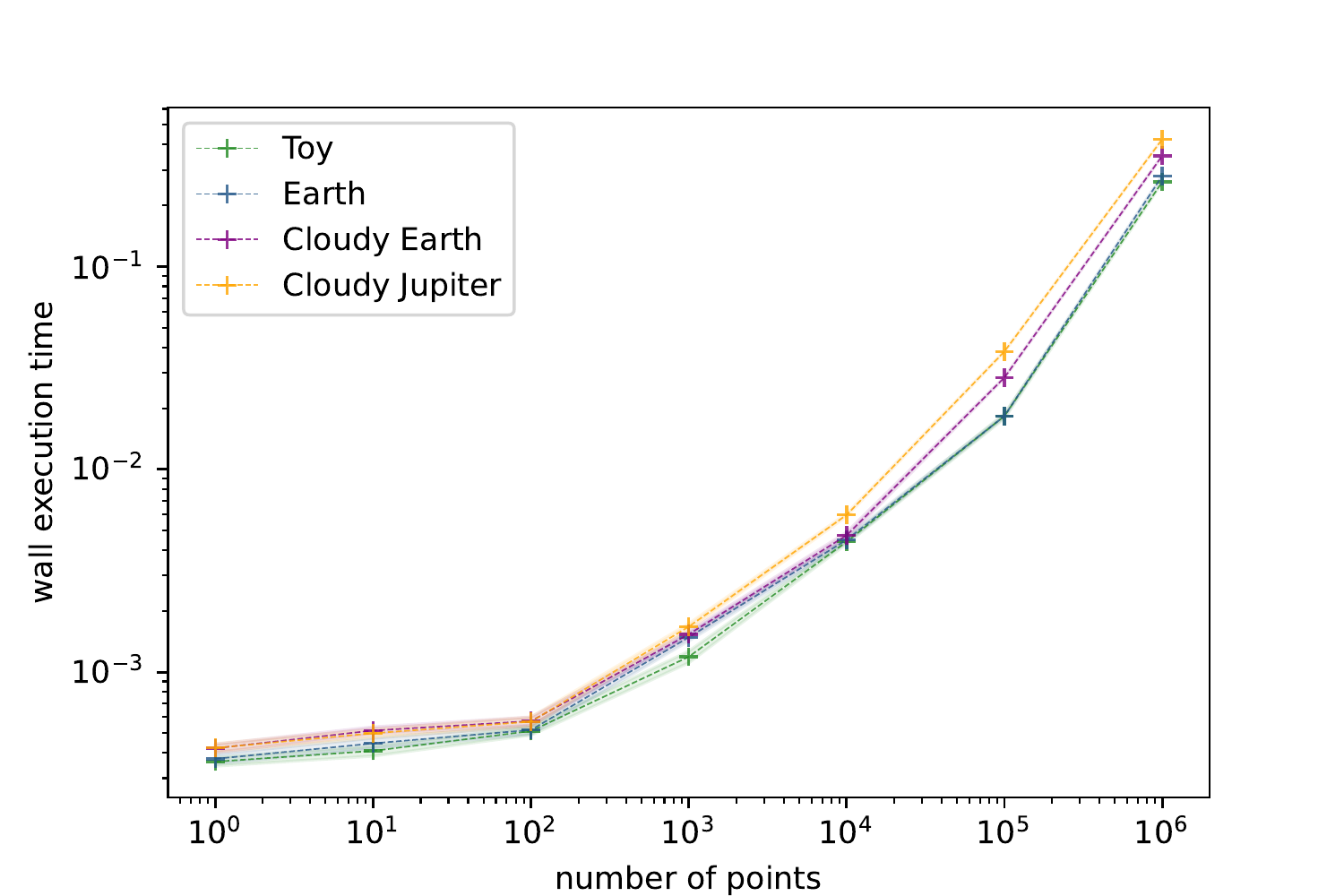}
    \end{center}  
\caption{Test cases with a network of width 40}
  \end{subfigure}
  \centering
  \begin{subfigure}{0.49\textwidth}
    \begin{center}
      \includegraphics[width=1\textwidth]{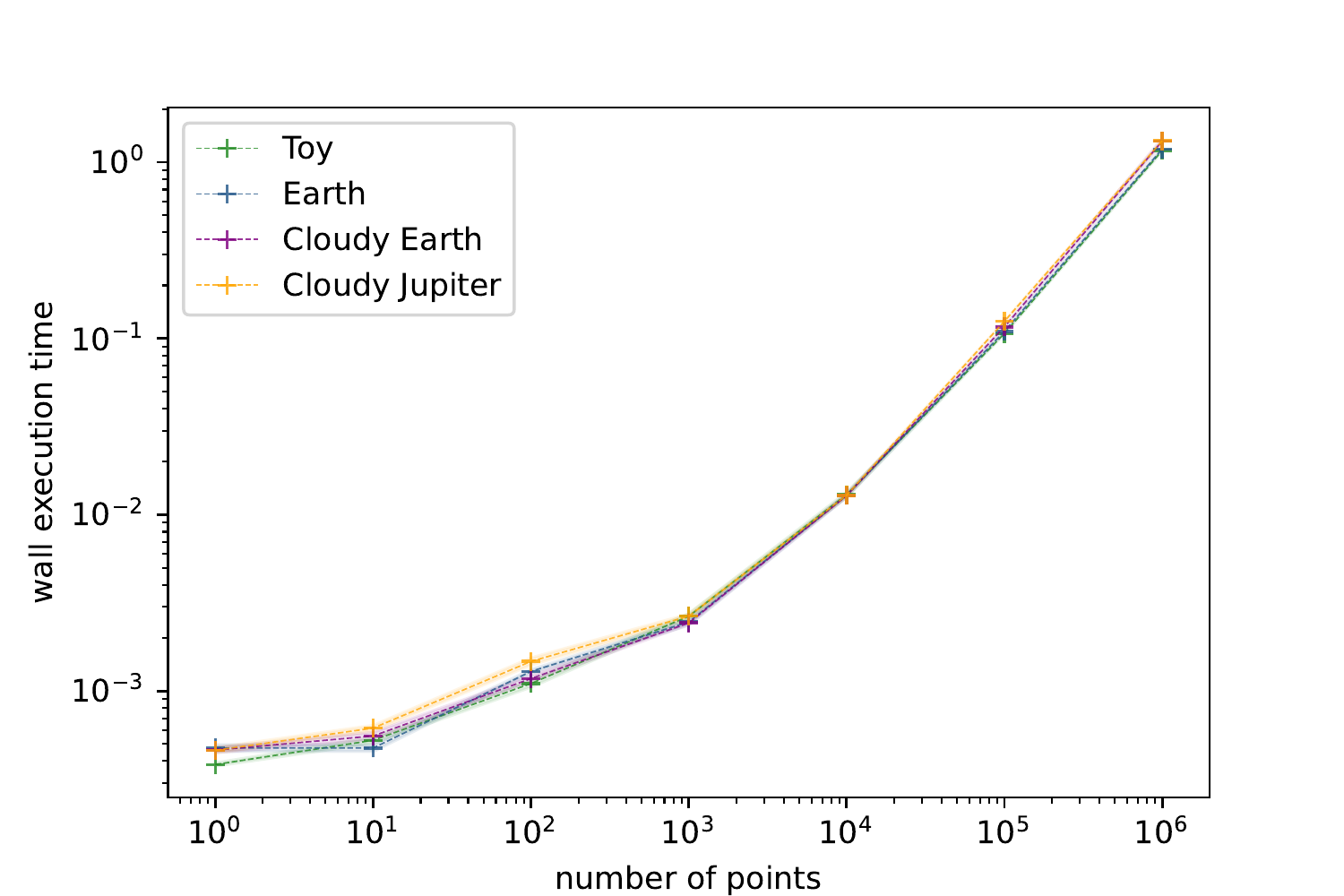}
    \end{center}  
\caption{Test cases with a network of width 160}
  \end{subfigure}
  \begin{subfigure}{0.49\textwidth}
    \begin{center}
      \includegraphics[width=1\textwidth]{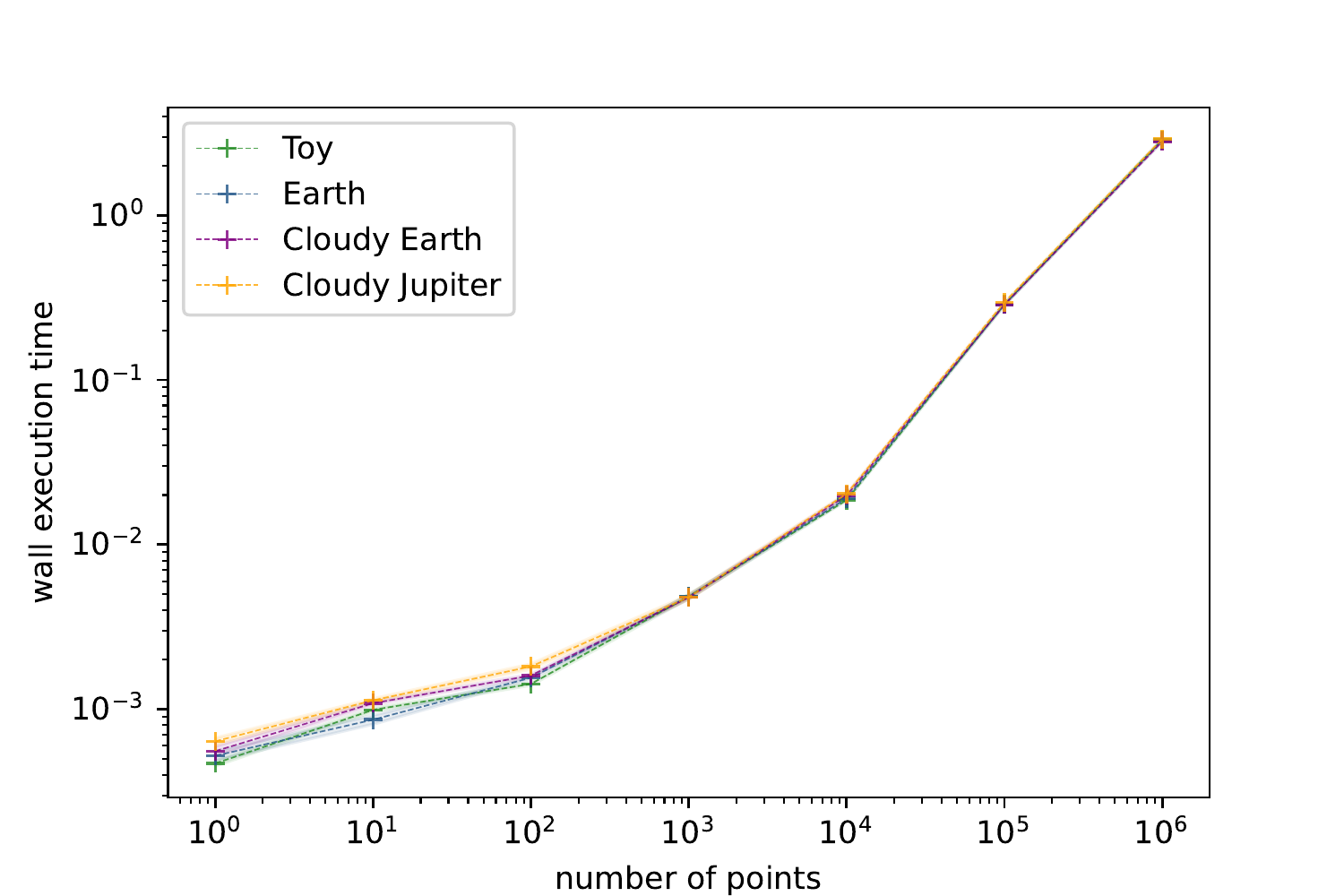}
    \end{center}  
\caption{Test cases with a network of width 320}
  \end{subfigure}
  \caption{Execution time of a neural network of a given width for the different test cases with respect to the number of input points (with log axes).}
\end{figure}

\newpage
\section*{Appendix B: Hyperparameter search space}\label{appB}

This table shows the hyperparameters considered in the hyperparameter search of Section \ref{sec:hsic}, as well es their possible values.

\begin{table}[h]
  \centering
  \begin{tabular}{lll}
    \toprule
    hyperparameter & type & values   \\
    \bottomrule
    \texttt{n\_layers} & integer & $\in \{1,...,10\}$ \\
    \texttt{n\_units} & integer & $\in \{7,...,512\}$\\
    \texttt{activation} & categorical & \texttt{elu}, \texttt{relu}, \texttt{tanh} or \texttt{sigmoid}\\
    \texttt{dropout} & boolean & \texttt{true} or \texttt{false} \\
    \texttt{dropout\_rate} & continuous & $\in [0,1]$ \\
    \texttt{batch\_norm} & boolean & \texttt{true} or \texttt{false} \\
    \texttt{learning\_rate} & continuous& $\in [1\times10^{-6},1\times10^{-2}]$ \\
    \texttt{weights\_reg\_l1} & continuous& $\in [1\times10^{-6},0.1]$ \\
    \texttt{weights\_reg\_l2} &continuous & $\in [1\times10^{-6},0.1]$ \\
    \texttt{bias\_reg\_l1} &continuous & $\in [1\times10^{-6},0.1]$ \\
    \texttt{bias\_reg\_l2} &continuous & $\in [1\times10^{-6},0.1]$ \\
    \texttt{batch\_size} & integer & $\in \{1,...,500\}$ \\
    \texttt{loss\_function} & categorical & $L_2$ error or $L_1$ error  \\
    \texttt{optimizer} & categorical  & \texttt{adam}, \texttt{sgd}, \texttt{rmsprop}, \texttt{adagrad} or \texttt{nadam} \\
    \texttt{amsgrad} & boolean & \texttt{true} or \texttt{false}  \\
    \texttt{1st\_moment\_decay} & continuous & $\in [0.8,1]$ \\
    \texttt{2nd\_moment\_decay} & continuous & $\in [0.8,1]$ \\
    \texttt{centered} & boolean & \texttt{true} or \texttt{false}\\
    \texttt{nesterov} & boolean & \texttt{true} or \texttt{false} \\
    \texttt{momentum} & continuous & $\in [0.5,0.99]$ \\
    \texttt{n\_seeds} & integer & $\in \{1,...,10\}$\\
    \texttt{sampling \cite{vbsw}} & boolean & \texttt{true} or \texttt{false} \\
    \texttt{weighting \cite{vbsw}} & boolean & \texttt{true} or \texttt{false} \\
  \end{tabular}
  \caption{ Hyperparameters values for Mutation++ approximation. We also include the weighting and the sampling scheme of \cite{vbsw} as hyperparameters. \label{tab:hyp}}
\end{table}

\end{document}

%% file: math_commands.tex
\def\eqref#1{equation~(\ref{#1})}
\def\Eqref#1{Equation~(\ref{#1})}

\def\1{\bm{1}}

\def\rx{{\textnormal{x}}}

\def\rvx{{\mathbf{x}}}

\def\vtheta{{\bm{\theta}}}
\def\valpha{{\bm{\alpha}}}

\def\valpha{{\bm{\alpha}}}

\def\vx{{\bm{x}}}

\DeclareMathAlphabet{\mathsfit}{\encodingdefault}{\sfdefault}{m}{sl}
\SetMathAlphabet{\mathsfit}{bold}{\encodingdefault}{\sfdefault}{bx}{n}

